\documentclass[runningheads]{llncs}

\usepackage{eccv}

\usepackage{eccvabbrv}

\usepackage{graphicx}
\usepackage{booktabs}
\graphicspath{{./figures/}}
\usepackage{hyperref}
\usepackage{xcolor}

\hypersetup{
    colorlinks=true,
    linkcolor=blue,
    filecolor=magenta,      
    urlcolor=red,  
    pdftitle={Overleaf Example},
    pdfpagemode=FullScreen,
}

\usepackage[accsupp]{axessibility}  

\usepackage{hyperref}

\usepackage{orcidlink}

\begin{document}
\title{Which Model Generated This Image? A Model-Agnostic Approach for Origin Attribution}
\titlerunning{Which Model Generated This Image?}

\author{Fengyuan Liu\inst{1} \and
Haochen Luo\inst{1} \and
Yiming Li\inst{2} \and
Philip Torr\inst{1} \and
Jindong Gu\inst{1}\thanks{Corresponding author}}

\authorrunning{F Liu et al.}

\institute{University of Oxford, Oxford OX1 3PJ, United Kingdom \and
Nanyang Technological University, Singapore 639798, Singapore \\
\email{oxfengyuan@gmail.com, jindong.gu@outlook.com}}

\maketitle

\begin{abstract}
Recent progress in visual generative models enables the generation of high-quality images. To prevent the misuse of generated images, it is important to identify the origin model that generates them. In this work, we study the origin attribution of generated images in a practical setting where only a few images generated by a source model are available and the source model cannot be accessed. The goal is to check if a given image is generated by the source model.
We first formulate this problem as a few-shot one-class classification task. To solve the task, we propose OCC-CLIP, a CLIP-based framework for few-shot one-class classification, enabling the identification of an image's source model, even among multiple candidates.
Extensive experiments corresponding to various generative models verify the effectiveness of our OCC-CLIP framework. Furthermore, an experiment based on the recently released DALL·E-3 API verifies the real-world applicability of our solution. 
Our source code is available at \url{https://github.com/uwFengyuan/OCC-CLIP}.
  
\end{abstract}

\section{Introduction}
\label{sec:intro}
Recent visual generative models are capable of producing images of exceptional quality, which have raised public concerns regarding Intellectual Property (IP) protection and accountability for misuse~\cite{gu2024responsible,li2024self,liu2023mm,liu2024latent}. In response to both the challenges and opportunities posed by Artificial Intelligence-Generated Content (AIGC), a recent U.S. executive order~\cite{BIDEN_2023} mandates that all AI-generated content must clearly label its source, such as Stable Diffusion~\cite{rombach2022high}. This makes the attribution of origins for generated images crucial in real-world applications, referring to the process of identifying whether a given image is generated by a particular model.

To address the origin attribution problem above, three main methods have been explored in the community. The first method involves watermarking~\cite{swanson1996transparent, luo2009reversible, pereira2000robust, tancik2020stegastamp, liu2022watermark}, which requires additional modifications to the generated results, affecting the quality of generation. The second method involves injecting fingerprints~\cite{yu2019attributing, ding2021does, yu2021artificial, yu2020responsible} into the model during training and employing a supervised classifier to identify these fingerprints. This process necessitates changes in training. Modification-free approaches have also been proposed, which do not require modifications to the generation or training processes. Specifically, the existing methods utilize inverse engineering~\cite{wang2023alteration, laszkiewicz2023single}, based on the idea that a synthetic sample can be most accurately reconstructed by the generator that created it. However, inverse engineering approaches necessitate access to the target model and require sampling many images as references.

\begin{figure}[t]
    \centering
    \scriptsize
    \includegraphics[width=0.95\linewidth]{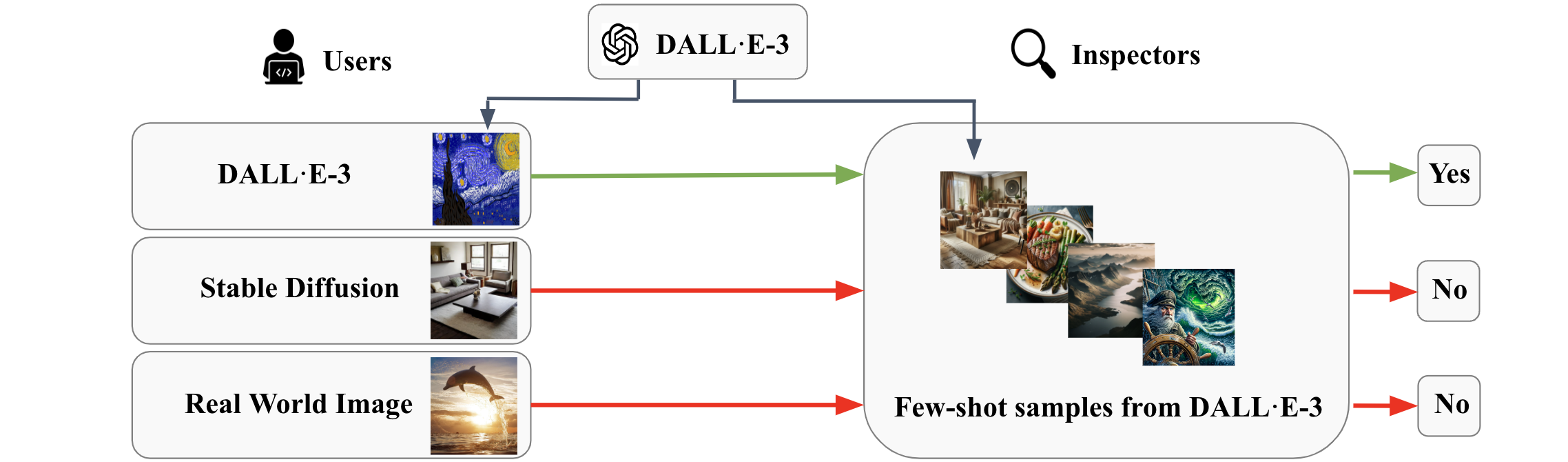}
    \vspace{-0.2cm}
    \caption{A simple demonstration of origin attribution in a practical, open-world setting. Inspectors receive a few samples from DALL·E-3. Users then submit an image, which could either be a real photograph or generated by DALL·E-3 or other models. If it's determined that the image and the provided samples were generated by the same model, we can then identify DALL·E-3 as the origin model of the query image.
    }
    \label{fig:teaser}
    \vspace{-0.6cm}
\end{figure}

In this work, we aim to conduct origin attribution in a practical open-world setting (Fig.~\ref{fig:teaser}), where model parameters cannot be accessed and only a few samples generated by the model are available. This setting is meaningful in real-world applications since current generative models, e.g., DALL·E-3~\cite{betker2023improving}, are not open-sourced, and sampling many images from them requires substantial costs.

To overcome the challenges in this setting, we first formulate the problem as a few-shot one-class classification task. Then, we propose a CLIP-based framework as an effective solution, dubbed OCC-CLIP. With our framework, we can determine if a given image and the few-shot available images were generated from the same model. If so, we can confidently identify the model that generated the few images as the origin model of the given image. Furthermore, we also demonstrate that our method can be extended to conduct origin attribution for multiple source models via One-vs-Rest.

Extensive experiments on various generative models have verified that our proposed framework can effectively determine the origin attribution of a given image. Additionally, our framework demonstrates superiority when different numbers of shots are available, and when image preprocessing is applied to the given images. It also proves effective in multi-source origin attribution scenarios. Furthermore, our experiments, based on the recently released DALL·E-3~\cite{betker2023improving} API, confirm the effectiveness of our solution in real-world commercial systems.

Our main contributions can be summarized as follows.
\begin{itemize}
    \item We propose a new task within a practical setting where generated images are attributed to the origin model only with few-shot available images generated by the model.
    \item We formulate the problem as a few-shot one-class classification task and then propose a CLIP-based framework, named OOC-CLIP, to address it.
    \item Extensive experiments are conducted on 8 generative models, including diffusion models and GANs. Further verification of our solution is carried out on a real-world image generative system, namely, DALL·E-3~\cite{betker2023improving}.
\end{itemize}

\section{Related Work}
\vspace{-0.2cm}

\textbf{Deep Visual Generative Models:} The advent of deep learning has brought significant advancements in deep visual generative models, leading to the development of sophisticated methods for creating synthetic media. Variational Autoencoders (VAEs) \cite{rezende2014stochastic, kingma2019introduction, oussidi2018deep} are notable for their dual-structure framework. In VAEs, an encoder condenses complex data into simpler latent representations, which a decoder then uses to reconstruct the original data. Generative Adversarial Networks (GANs) \cite{goodfellow2020generative, arjovsky2017wasserstein, tao2023galip, kang2023scaling, sauer2023stylegan} operate on a principle of competition between two components: a generator that creates data samples and a discriminator that judges their authenticity. Through iterative training, both components enhance their capabilities, improving the overall quality of the generated data. Diffusion Models \cite{ho2020denoising, song2020denoising, rombach2022high, gu2022vector, nichol2021glide, saharia2022photorealistic} employ a two-phase process. Initially, they transform data into noise, and then methodically remove this noise to reverse the process. This approach can generate highly realistic images.

\vspace{0.3em}
\noindent\textbf{Origin Attribution of Generated images:}
Origin attribution, distinct from generated image detection, seeks to determine whether specific images were produced by a particular model. Various strategies have been proposed to address this challenge. One method involves embedding watermarks~\cite{swanson1996transparent, luo2009reversible, pereira2000robust, tancik2020stegastamp} in images to trace their origins. However, this approach faces limitations as different models might use identical watermarks, which can also be manipulated or removed~\cite{liu2022watermark}. Alternatively, injecting unique fingerprints~\cite{yu2019attributing, ding2021does, yu2021artificial, yu2020responsible} into a model during its training phase enables the identification of these markers using a supervised classifier. Although effective, this technique necessitates modifications to the training process and the model's architecture. A different, modification-free method is inverse engineering~\cite{wang2023alteration, laszkiewicz2023single}, which leverages the principle that a synthetic sample can be most accurately reconstructed by its original generator. However, this approach requires access to the target model and extensive sampling of images for comparison. Furthermore, three studies have explored this area under different settings. One develops a multi-class classifier known as an attributor for fake image attribution. Yet, it is only applicable to text-to-image models and operates within a limited, closed-world scenario\cite{sha2023fake}. One~\cite{girish2021towards} utilizes an iterative multi-step pipeline to detect the origins of images generated by different GANs. However, it requires many samples. The other study\cite{kim2020decentralized} demonstrates a theoretical lower bound on attribution accuracy using smaller datasets (e.g. MNIST~\cite{lecun2010mnist}). This method is only suitable for unconditional GANs and reveals limited scalability to more complex systems, such as DALL·E-3~\cite{betker2023improving}.

\vspace{0.3em}
\noindent\textbf{Few-shot One-class classification:} 
One-class classification has long been recognized as a challenging problem, with numerous studies~\cite{scholkopf2001estimating,chen2017outlier,schlegl2017unsupervised,sabokrou2018adversarially} addressing it. However, few works have explored the few-shot one-class classification (FS-OCC) problem in the image domain. Previous FS-OCC research in the image domain, specifically based on meta-learning~\cite{frikha2021few}, suffers from memory inefficiency, making it unsuitable for processing high-resolution images. In contrast, the CLIP-based classifier, CoOp~\cite{zhou2022learning}, has demonstrated excellent performance in few-shot learning settings, where a few images from each class are available. Our setting, however, differs as it involves only a few images from a single class, generated by a generative model. Therefore, instead of pursuing multi-class classification, we utilize CLIP for one-class classification.

\section{Approach}
In this section, we first introduce the standard CLIP-based classification framework and then present our OCC-CLIP framework for few-shot one-class classification. At the end, we show how to extend our framework to multiple classes.

\subsection{Background of CLIP-based Classification}
CLIP is a multimodal model pre-trained to predict whether an image matches a text prompt. It includes an image encoder \(E_v(\cdot)\) and text encoder \(E_t(\cdot)\). The pre-trained CLIP can perform zero-shot multi-class classification by comparing the image with a list of prompts~\cite{gu2023systematic}, each representing a class. Formally, assume we have \(K\) classes. Let \(X^v\)$ \in\mathcal{X}$ denote an image and \(X^t_i\) represent the \(i_{\text{th}}\) prompt representing the \(i_{\text{th}}\) class. The predicted \(i_{\text{th}}\) class probability of the image \(X^v\) is computed as follows:
\begin{equation}
p(\text{class} = i|v) = \frac{\exp(\mathrm{sim}(E_t(X^t_i), E_v(X^v)))}{\sum_{j=1}^{K} \exp(\mathrm{sim}(E_t(X^t_j), E_v(X^v)))},
\end{equation}
where $\mathrm{sim}(\cdot)$ measures the distance between two embeddings, e.g., dot product.

In the classification above, hand-crafted text prompts are applied to represent classes~\cite{gu2023systematic}. The prompt designs require specialized knowledge and are time-consuming to create. To alleviate this, Zhou et al.~\cite{zhou2022learning} introduced the concept of Context Optimization (CoOp), which employs learnable vectors to refine prompt-related words. Instead of manually designing a prompt, CoOp enables the model to automatically optimize for a suitable prompt. Formally, the prompt for the \(i_{\text{th}}\) class can be represented as \(X^p_i = [t]\otimes[CLASS_{i}]\), where \(t\) is the learnable context vectors, $\otimes$ is a concatenation operation and $CLASS_i$ corresponds to the name of \(i_{\text{th}}\) class. The objective of optimization is to minimize the error of predicting ground truth $Y_{i}$ for each $X^v_{i}$. This is achieved by using a cross-entropy loss function $\mathcal{L}$ with respect to the learnable prompts. Let \(f\) represent the pre-trained CLIP model. The optimization can be described as:
\begin{equation}
    \underset{X^p}{\text{min}} \sum_{j=1}^{N}\sum_{i=1}^{K} \mathcal{L}(f(E_v(X^v_j), E_t(X^p_{i})), Y_{i}),
\end{equation}
where $N$ is the number of images.
In practice, CoOp~\cite{zhou2022learning} shows that only a small number of labeled examples are required to learn effective prompts. In summary, prompt learning enables CLIP-based Classification effective in the few-shot learning setting.

\begin{figure*}[t] 
    \centering
    \includegraphics[width=\linewidth]{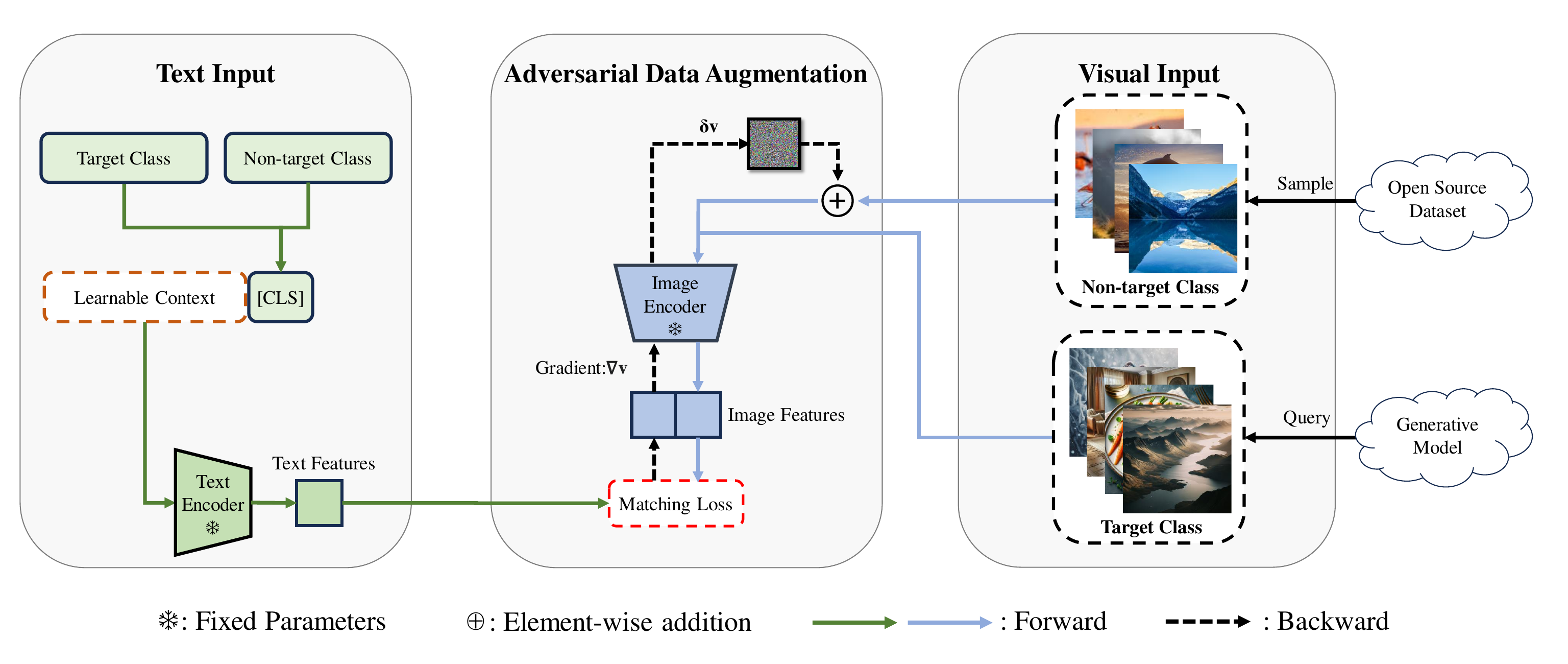}
    \vspace{-0.8cm}
    \caption{Overview of OCC-CLIP. The input text is represented by learnable context vectors, followed by two discrete classes: the \textit{target} class corresponds to an image set queried from a generative model, and the \textit{non-target} class corresponds to randomly sourced open-domain images. These classes can be labeled as contrasting pairs, such as non-target vs. target or negative vs. positive. The parameters for the text and image encoders, derived from the CLIP model, are fixed. Adversarial Data Augmentation (ADA) calculates the gradient of each pixel across non-target images. In the training phase, these gradients \(\delta^v\) are applied to the non-target images.}
    \label{fig:my_framework}
    \vspace{-0.6cm}
\end{figure*}

\subsection{CLIP-based Few-shot One-Class Classification}
The CLIP-based classifier, CoOp, can achieve excellent performance in the few-shot learning setting where a few images from each class are available.
Nevertheless, in our setting, only a few images from one class, generated by a generative model, are available. Hence, a standard CLIP-based classifier cannot be applied directly to solve the few-shot one-class classification task.

We now present our CLIP-based framework for One-Class Classification, called \textbf{OCC-CLIP}. In our framework (Fig.~\ref{fig:my_framework}), the few images collected from a generative model are treated as the target class, while the images randomly sampled from a clean dataset are labeled as the non-target class. The two classes can be labeled as any contrasting pairs, such as non-target vs. target or negative vs. positive. Two learnable prompts corresponding to the target and non-target classes are optimized on these images, respectively. 

The few images selected for the non-target class cannot represent the whole distribution of the non-target class well since they are randomly sampled from an open-source dataset (e.g. ImageNet~\cite{deng2009imagenet}). To overcome this challenge, we propose an adversarial data augmentation (\textbf{ADA}) technique that, during training, extends coverage of the non-target class space and more closely approximates the boundary to the target space, thereby improving the model's ability to learn the attribution of the target model. ADA aims to maximize the loss by adding small perturbations \(\delta^v\) to non-target images, while the learnable prompts aim to minimize the loss by learning the boundary between the target and non-target classes. In summary, the optimization in OCC-CLIP can be formulated as:
\begin{equation}
    \min_{X^p} \max_{\delta^v} \sum_{j=1}^{N}\sum_{i=1}^{K} \mathcal{L}(f(E_v(X^v_j + \delta^v_j), E_t(X^p_i)), Y_i),
\end{equation}
where \(\delta^v\) is the adversarial image perturbation computed by our ADA technique. 

\begin{algorithm}[t] 
\SetKwComment{Comment}{$\triangleright$\ }{}
\SetKwData{Left}{left}\SetKwData{This}{this}\SetKwData{Up}{up} 
\SetKwInOut{Input}{input}\SetKwInOut{Output}{output}
    \Input{\(X^v\): images, \(X^p\): learnable prompts, \(\mathcal{G}\): ground truth, \(\epsilon\): updating step size} 
    \BlankLine 
    \For{Iterations}{
    $EV = E_v(X^v)$, $ET = E_t(X^p)$ \Comment*{Extract embeddings}
    \(G = \nabla_{\mathcal{X}}\mathcal{L}(f(EV, ET), \mathcal{G})\) \Comment*{Gradient calculation} 
    \(\delta^v = \epsilon \cdot \text{sign}(G)\) \Comment*{Perturbation generation}
    $EV \leftarrow E_v(X^v + \delta^v)$ \Comment*{Update image embeddings}
    \(X^p \leftarrow \min_{X^p} \mathcal{L}(f(EV, ET), \mathcal{G})\) \Comment*{Prompt optimization}
     }
    \caption{OCC-CLIP Framework}
    \label{alg: adversarial_data_augmentation} 
\end{algorithm}

The implementation of our OCC-CLIP is shown in Algorithm~\ref{alg: adversarial_data_augmentation}. As shown in the algorithm, a forward pass and a backward pass are conducted to obtain gradient information for the images of the non-target class. The gradient information is used to compute adversarial perturbation. The learnable prompt will be updated in another forward and backward passes on the perturbed images. The sensitivity of training hyperparameters is discussed in experiments section.  

In the optimization process, both visual and textual encoders of CLIP are frozen. In the verification process, if it is classified into the target class, an image will be determined to be generated by the same generative model as the source model of the target images.

\subsection{CLIP-based Few-shot Multi-Class Classification}
We also explore the multi-source origin attribution scenarios. For example, to determine if the origin of an image can be attributed to ProGAN~\cite{karras2017progressive}, Stable Diffusion~\cite{rombach2022high}, or Vector Quantized Diffusion~\cite{gu2022vector}, we can employ three one-class classifiers corresponding to these models for classification. Given a set of trained $K$ one-class classifiers $\{OCC_1, OCC_2, \ldots, OCC_K\}$ for $K$ classes and a threshold $\theta$ (e.g. 0.5), for an input sample $X^v$, let $s_i(X^v)$ denote the score of $X^v$ given by $i$-th classifier $OCC_i$. The predicted class $C(X^v)$ for the sample $X^v$ is determined as follows:
\begin{equation}
C(X^v) = 
\begin{cases} 
\arg\max_{i \in \{1, \ldots, K\}} s_i(X^v) & \text{if } \max_{i \in \{1, \ldots, N\}} s_i(X^v) > \theta, \\
\text{others} & \text{otherwise}.
\end{cases}
\end{equation}
Given an image, if the maximum score of \(X^v\), provided by the \(i\)-th classifier, exceeds the threshold, then \(X^v\) is classified into the \(i\)-th class. Otherwise, it is considered to belong to a category outside those defined by the \(K\) classifiers.

\section{Experiments}
In this section, we first describe experimental settings and present our comparison with baseline methods. We also study the sensitivity of our method to various factors, such as target class corresponding to source models, non-target class datasets, the number of available images, and image preprocessing. Furthermore, we show the effectiveness of our framework in multi-source origin attribution scenarios and real-world commercial generation API.

\subsection{Experimental Setting}

\noindent\textbf{Dataset.} There are total 202,520 images generated by five different generative models, i.e. Stable Diffusion Model~\cite{rombach2022high}, Latent Diffusion Model~\cite{rombach2022high}, GLIDE~\cite{nichol2021glide}, Vector Quantized Diffusion~\cite{gu2022vector}, and GALIP~\cite{tao2023galip}, based on the validation set of  Microsoft Common Objects in Context (COCO) 2014 dataset~\cite{lin2014microsoft}. These models are pre-trained on four different datasets, i.e. LAION-5B~\cite{schuhmann2022laion}, COCO~\cite{lin2014microsoft}, LAION-400M~\cite{schuhmann2021laion}, and filtered CC12M~\cite{changpinyo2021conceptual}. In total, five image datasets from different source models are generated, namely SD, VQ-D, LDM, Glide, GALIP. To balance the number of image datasets generated by Diffusion models and the number of image datasets generated by GANs, we utilized pre-existing datasets (namely GauGAN~\cite{park2019semantic}, ProGAN~\cite{karras2017progressive}, and StyleGAN2~\cite{karras2020analyzing}) as provided by~\cite{wang2020cnn}. These datasets collectively serve as a robust benchmark covering two primary generative techniques: GANs and Diffusion Models.

\vspace{0.3em}
\noindent\textbf{Model.} OCC-CLIP utilizes 16 context vectors. This model is built upon the open-source CLIP framework. The image encoder uses the ViT-B/16 architecture. Except for the prompt learner, all pre-trained parameters are fixed. Initial context vectors are stochastically sampled from a Normal distribution characterized by a mean of 0 and a standard deviation of 0.02. 

\vspace{0.3em}
\noindent\textbf{Training Setting.} All the generated images are resized to \(224 \times 224\) and then normalized according to the pre-trained datasets of each model. Stochastic Gradient Descent is utilized as the optimization strategy with a learning rate of 0.0001, modulated through cosine annealing. The cross-entropy loss is utilized as the loss function. By default, the training process is capped at a maximum of 200 epochs for 50-shot scenarios. The test dataset consists of 1,000 images that are randomly selected from the test sets to ensure a reliable evaluation. To counteract the onset of explosive gradients in the nascent phases of training, the learning rate is steadfastly maintained at \(1 \times 10^{-5}\) during the first epoch. The eight generative models (i.e. SD, VQ-D, LDM, Glide, GALIP, ProGAN, StyleGAN2, GauGAN) are iteratively treated as the target class, while four open-source datasets (i.e. COCO~\cite{lin2014microsoft}, ImageNet~\cite{deng2009imagenet}, Flickr~\cite{young2014image}, and CC12M~\cite{changpinyo2021conceptual}) are iteratively treated as the non-target class. However, in the default settings, only half of the non-target images are augmented by ADA, SD is designated as the target image set, and COCO is chosen as the non-target image set.

\vspace{0.1cm}
\noindent\textbf{Evaluation.} Each model is evaluated by the Area Under the Receiver Operating Characteristic Curve (AUC). 
The Receiver Operating Characteristic Curve (ROC) is a graphical representation that plots the True Positive Rate (TPR) against the False Positive Rate (FPR) at various threshold levels. 

The AUC represents the probability that a classifier will rank a randomly chosen positive instance higher than a randomly chosen negative one. To reduce randomness, each model is trained 10 times with different training sets each time. Then, the mean AUC and the corresponding standard deviation are reported over the test set. A higher AUC score signifies better performance. For each table, the corresponding standard deviations are shown in the supplementary. More experimental details, such as different testing tasks and accuracy scores, can be found in the supplementary.

\subsection{Comparison with Baselines}

\noindent\textbf{Baselines.} Since there are currently no methods perfectly suited to our setting, we conduct a comprehensive evaluation of OCC-CLIP by assessing 12 benchmark methods from various usage areas (see Table~\ref{table: 14 baselines}). The supplementary can find a comparison with other baselines~\cite{wang2023alteration,girish2021towards}.

\begin{table*}[t]
    \centering
    \scriptsize
        \caption{Compare the performance of OCC-CLIP in orgin attribution against 12 basic methods. During the training phase, the target class is sourced from SD, and the non-target class is from COCO. In the test phase, the target class remains sourced from SD, but the non-target class is sourced from another generative image dataset. The optimal outcomes for individual datasets are emphasized using \textbf{bold} formatting.}
        \vspace{-0.3cm}
        \setlength{\tabcolsep}{3pt}
        \begin{tabular}{@{}l|cccccccccc@{}}
        \toprule[1pt]
        Methods & VQ-D & LDM & Glide & GALIP & ProGAN & StyleGAN2 & GauGAN & Overall \\
        \midrule 
        VGG16~\cite{simonyan2014very} & $0.6458$
            & $0.5438$
            & $0.5652$
            & $0.7017$
            & $0.6609$
            & $0.6170$
            & $0.7096$
            & $0.6349$ \\
        ResNet50~\cite{he2016deep} & $0.6693$
            & $0.5485$
            & $0.6103$
            & $0.7307$
            & $0.6646$
            & $0.6458$
            & $0.7475$
            & $0.6595$ \\ 
        Inception-v3~\cite{szegedy2016rethinking} & $0.6378$
            & $0.5349$
            & $0.5252$
            & $0.6759$
            & $0.6453$
            & $0.5947$
            & $0.6982$
            & $0.6160$ \\
        DenseNet-121~\cite{huang2017densely} & $0.7264$
            & $0.5337$
            & $0.5730$
            & $0.7478$
            & $0.7204$
            & $0.6645$
            & $0.8091$
            & $0.6821$ \\
        \midrule
        ViT-B-16~\cite{dosovitskiy2020image} & $0.7502$
            & $0.5387$
            & $0.6069$
            & $0.7837$
            & $0.6062$
            & $0.6593$
            & $0.7259$
            & $0.6673$ \\ 
        DeiT-B-16~\cite{touvron2021training} & $0.6825$
            & $0.5366$
            & $0.5698$
            & $0.6592$
            & $0.5721$
            & $0.5898$
            & $0.6507$
            & $0.6087$ \\ 
        CaiT-S-24~\cite{touvron2021going} & $0.6522$
            & $0.5301$
            & $0.5550$
            & $0.6466$
            & $0.5697$
            & $0.5804$
            & $0.6645$
            & $0.5998$ \\ 
        Swin-B-4~\cite{liu2021swin} & $0.8634$
            & $0.7180$
            & $0.7473$
            & $0.8478$
            & $0.5879$
            & $0.7054$
            & $0.7822$
            & $0.7503$ \\ 
        \midrule
        Image-Patch~\cite{mandelli2022detecting} & $0.6638$
            & $0.5269$
            & $0.5534$
            & $0.6847$
            & $0.6624$
            & $0.6706$
            & $0.7674$
            & $0.6470$ \\ 
        Feature-Patch~\cite{mandelli2022detecting} & $0.6999$
            & $0.6116$
            & $0.7385$
            & $0.6285$
            & $0.7136$
            & $0.8546$
            & $0.8152$
            & $0.7231$ \\ 
        \midrule
        CLIP~\cite{radford2021learning} & $0.7272$
            & $0.7243$
            & $0.6078$
            & $0.7304$
            & $0.5229$
            & $0.5393$
            & $0.6463$
            & $0.6426$ \\ 
        CoOp~\cite{zhou2022learning} & $0.9503$
            & $0.8373$
            & $0.9266$
            & $0.9660$
            & $0.8861$
            & $0.9533$
            & $0.9643$
            & $0.9263$ \\ 
        \midrule
        OCC-CLIP & $\mathbf{0.9703}$
            & $\mathbf{0.8801}$
            & $\mathbf{0.9519}$%
            & $\mathbf{0.9798}$
            & $\mathbf{0.9452}$
            & $\mathbf{0.9651}$
            & $\mathbf{0.9910}$
            & $\mathbf{0.9548}$ \\ 
        \bottomrule[1pt]
        \end{tabular}
    
    \label{table: 14 baselines}
    \vspace{-0.5cm}
\end{table*}

\textit{Traditional Binary Classification Models:} Four of these methods are notable CNN architectures that are often applied to Computer Vision problems, especially for image classification tasks, including ResNet \cite{he2016deep}, Inception \cite{szegedy2016rethinking}, DenseNet \cite{huang2017densely}, and VGG~\cite{simonyan2014very}. The other four models are adaptations of the transformer paradigm, including ViT \cite{dosovitskiy2020image}, Deit \cite{touvron2021training}, Cait \cite{touvron2021going}, and Swin \cite{liu2021swin}. For both CNN and Transformer models, all layers are retained in a frozen state except for the last classification layer. 

\textit{Patch-Driven Methods~\cite{mandelli2022detecting}:} Two of the methods, Feature-Patch and Image-Patch, are patch-based techniques that are primarily used for deepfake and rely on the ResNet-50 architecture for feature extraction. For the Image-Patch model, each image is divided into 2x2 patches, with each patch serving as a separate input to the ResNet-50 architecture.
In the case of the Feature-Patch model, which shares the same backbone, the last five layers of the standard ResNet-50 architecture are discarded and the remaining structures are employed to extract image features. Subsequently, these features are segmented into 4x4 patches, with each patch being passed through the concluding linear classification layer.

\textit{Vision-Language Models:} Considering the power of the vision-language model in tackling downstream classification tasks, zero-shot CLIP~\cite{radford2021learning} and CoOp\cite{zhou2022learning} are also evaluated. The zero-shot CLIP~\cite{radford2021learning} approach utilizes custom-crafted prompts (hard prompt)~\cite{gu2023systematic}, adopting the format "a photo of a [CLASS]" to tackle the tasks of origin attribution. Conversely, CoOp employs prompt tuning (soft prompt) and distinguishes itself by using \(\{v_1, v_2, \ldots, v_{16}, \text{[CLASS]}\}\) where \(v_i\) is an adjustable context vector and [CLASS] is the class token which is deliberately located at the end of the sequence. 

\begin{table*}[t]
    \centering
    \tiny
        \setlength{\tabcolsep}{3pt}
        \caption{Evaluation sensitivity of OCC-CLIP to Source Models. The leftmost column represents target datasets. In the training phase, the non-target images are from COCO. In the testing phase, the non-target images are from a different generative image dataset shown in the first row. The optimal outcomes for individual datasets are emphasized using \textbf{bold} formatting.}
        \vspace{-0.3cm}
        \begin{tabular}{l|c|cccccccccc}
        \toprule[1pt]
        Target $\downarrow$ & Method 
        & SD
        & VQ-D
        & LDM
        & Glide
        & GALIP
        & ProGAN
        & StyleGAN2
        & GauGAN
        & Overall \\
        \midrule
        \multirow{2}{*}{SD} 
            & \tiny CoOp
            & -- 
            & $0.9503$
            & $0.8373$
            & $0.9266$
            & $0.9660$
            & $0.8861$
            & $0.9533$
            & $0.9643$
            & $0.9263$ \\ 
            & \tiny OCC-CLIP 
            & -- 
            & $\mathbf{0.9703}$
            & $\mathbf{0.8801}$
            & $\mathbf{0.9519}$
            & $\mathbf{0.9798}$
            & $\mathbf{0.9452}$
            & $\mathbf{0.9651}$
            & $\mathbf{0.9910}$
            & $\mathbf{0.9548}$ \\ 
        \midrule 
        \multirow{2}{*}{VQ-D} 
            & \tiny CoOp
            & $0.9988$
            & --
            & $\mathbf{0.7337}$
            & $\mathbf{0.7435}$
            & $\mathbf{0.7558}$
            & $0.9290$
            & $0.9924$
            & $0.9352$
            & $0.8698$ \\ 
            & \tiny OCC-CLIP 
            & $\mathbf{0.9992}$
            & --
            & $0.6924$
            & $0.7264$
            & $0.7327$
            & $\mathbf{0.9931}$
            & $\mathbf{0.9936}$
            & $\mathbf{0.9936}$
            & $\mathbf{0.8758}$ \\ 
        \midrule
        \multirow{2}{*}{LDM} 
            & \tiny CoOp 
            & $0.9925$
            & $0.6793$
            & --
            & $0.6468$
            & $0.6263$
            & $0.9565$
            & $0.9896$
            & $0.9758$
            & $0.8381$ \\ 
            & \tiny OCC-CLIP 
            & $\mathbf{0.9957}$
            & $\mathbf{0.7507}$
            & --
            & $\mathbf{0.6847}$
            & $\mathbf{0.6530}$
            & $\mathbf{0.9956}$
            & $\mathbf{0.9940}$
            & $\mathbf{0.9992}$
            & $\mathbf{0.8676}$ \\ 
        \midrule
        \multirow{2}{*}{Glide}
            & \tiny CoOp 
            & $0.9985$
            & $0.8573$
            & $0.8314$
            & --
            & $0.6687$
            & $0.9629$
            & $0.9916$
            & $0.9814$
            & $0.8988$ \\ 
            & \tiny OCC-CLIP 
            & $\mathbf{0.9998}$
            & $\mathbf{0.8958}$
            & $\mathbf{0.8585}$
            & --
            & $\mathbf{0.6834}$
            & $\mathbf{0.9974}$
            & $\mathbf{0.9949}$
            & $\mathbf{0.9997}$
            & $\mathbf{0.9185}$ \\ 
        \midrule
        \multirow{2}{*}{GALIP} 
            & \tiny CoOp 
            & $0.9999$
            & $0.9036$
            & $0.8441$
            & $\mathbf{0.7802}$
            & --
            & $0.9982$
            & $0.9992$
            & $0.9996$
            & $0.9321$ \\ 
            & \tiny OCC-CLIP 
            & $\mathbf{0.9999}$
            & $\mathbf{0.9345}$
            & $\mathbf{0.8626}$
            & $0.7779$
            & --
            & $\mathbf{0.9999}$
            & $\mathbf{0.9993}$
            & $\mathbf{1.0000}$
            & $\mathbf{0.9392}$ \\ 
        \midrule
        \multirow{2}{*}{ProGAN} 
            & \tiny CoOp 
            & $\mathbf{0.9972}$
            & $0.9475$
            & $0.9544$
            & $\mathbf{0.9453}$
            & $0.9818$
            & --
            & $\mathbf{0.9278}$
            & $0.7993$
            & $0.9362$ \\ 
            & \tiny OCC-CLIP 
            & $0.9961$
            & $\mathbf{0.9477}$
            & $\mathbf{0.9585}$
            & $0.9320$
            & $\mathbf{0.9885}$
            & --
            & $0.8471$
            & $\mathbf{0.8885}$
            & $\mathbf{0.9369}$ \\ 
        \midrule
        \multirow{2}{*}{StyleGAN2} 
            & \tiny CoOp 
            & $0.9992$
            & $0.9856$
            & $0.9850$
            & $0.9584$
            & $0.9849$
            & $0.8687$
            & --
            & $0.9543$
            & $0.9623$ \\ 
            & \tiny OCC-CLIP 
            & $\mathbf{0.9996}$
            & $\mathbf{0.9937}$
            & $\mathbf{0.9851}$
            & $\mathbf{0.9652}$
            & $\mathbf{0.9926}$
            & $\mathbf{0.9649}$
            & --
            & $\mathbf{0.9946}$
            & $\mathbf{0.9851}$ \\ 
        \midrule
        \multirow{2}{*}{GauGAN}
            & \tiny CoOp 
            & $\mathbf{0.9991}$
            & $\mathbf{0.9388}$
            & $\mathbf{0.9862}$
            & $\mathbf{0.9757}$
            & $0.9901$
            & $0.6812$
            & $\mathbf{0.9676}$
            & --
            & $0.9368$ \\ 
            & \tiny OCC-CLIP 
            & $0.9982$
            & $0.9132$
            & $0.9745$
            & $0.9593$
            & $\mathbf{0.9958}$
            & $\mathbf{0.7752}$
            & $0.9425$
            & --
            & $\mathbf{0.9370}$ \\ 
        \bottomrule[1pt]
        \end{tabular}
        
        \vspace{-0.5cm}
        \label{Table:Detect Different Datasets}
\end{table*}

\noindent\textbf{Results and Analysis.}
Table~\ref{table: 14 baselines} provides a comparative performance analysis of OCC-CLIP and 12 benchmark models. During the training phase, the target class is sourced from SD, and the non-target class is from COCO. In the test phase, the target class remains sourced from SD, but the non-target class is sourced from another generative image dataset (VQ-D, LDM, Glide, GALIP, ProGAN, StyleGAN2, and GauGAN). The goal of this section is to assess each model's ability to accurately determine whether the origin of a given image can be attributed to Stable Diffusion.

Among the CNN-based and Transformer-based baselines, Swin outshines its counterparts. However, the overall disparity in performance between CNN-based and Transformer-based models is not obvious. The following two methods, Image-Patch and Feature-Patch are both based on ResNet50. Image-Patch performs worse than ResNet50 while Feature-Patch has somewhat but limited improvements compared with ResNet50. This suggests that current methods used to do deepfake detection perform poorly in this few-shot origin attribution scenarios. In addition to the aforementioned classical methods, OCC-CLIP surpasses widely-spread vision-language models such as zero-shot CLIP and CoOp. In detail, OCC-CLIP outperforms zero-shot CLIP by an average of around 31\% and CoOp by approximately 2.9\%. This progression emphasizes the value of ADA. In a comprehensive assessment, OCC-CLIP emerges as a leading standard, underscoring its exceptional ability in origin attribution. However, it is harder to distinguish images generated by very similar algorithms trained on similar datasets, such as DeepFloyd, which is based on Stable Diffusion and also trained on the COCO dataset. In this case, our model's performance does decrease (dropping to $0.7650\scriptscriptstyle \pm \scriptstyle4.80e\text{-}2$). Nevertheless, our method still outperforms the best baseline (CoOp: $0.7301\scriptscriptstyle \pm \scriptstyle7.91e\text{-}2$).

\subsection{Ablation Study}
\textbf{Sensitivity to Source Models.} Table~\ref{Table:Detect Different Datasets} shows the performance of CoOp and OCC-CLIP in origin attribution with eight different source models. During training, the target dataset is from the dataset shown in the leftmost column, and the non-target dataset is COCO. During testing, the target remains the same, but the non-target dataset is replaced with one of the other generated datasets shown in the first row. For example, the source model for the second row in the table is Stable Diffusion~\cite{rombach2022high}. 

OCC-CLIP demonstrates superior performance in most cases and outperforms the baseline on average in the results from testing with seven other datasets. Overall, our proposed model has exhibited superior performance and generalization capabilities in determining if an image is from the same source model as a set of target images, compared to the baseline.

\begin{table}[t]
    \vspace{0.1cm}
    \centering
    \scriptsize
        \caption{Evaluation of OCC-CLIP on Different Non-target Image Datasets. During training phase, the non-target class is chosen from one of the four datasets (COCO, CC12M, Flickr, and ImageNet) or from combined datasets. The target dataset is from SD. The performance is averaged across seven different testing tasks.}
        \vspace{-0.3cm}
        \setlength{\tabcolsep}{14pt}
        \begin{tabular}{@{}lccccc@{}}
        \toprule[1pt]
        Methods & COCO & CC12M & Flickr & ImageNet & Combined \\
        \midrule 
        CoOp 
        & $0.9263$
        & $0.8689$
        & $0.8788$
        & $0.9377$
        & $0.9320$ \\
        OCC-CLIP 
        & $\mathbf{0.9548}$
        & $\mathbf{0.9320}$
        & $\mathbf{0.9355}$
        & $\mathbf{0.9710}$
        & $\mathbf{0.9651}$ \\
        \bottomrule[1pt]
        \end{tabular}
    
    \label{table: Different Real Datasets}
    \vspace{-0.5cm}
\end{table}

\vspace{0.05cm}\noindent\textbf{Sensitivity to Selection of Non-target Class.} As shown in Table~\ref{table: Different Real Datasets}, we select non-target images from combined datasets or from one of the four open-domain image datasets: COCO, ImageNet, Flickr, or CC12M. The target images are from SD. The average AUC score over 7 different testing tasks is computed. It can be noted that although the choice of the source of non-target images can somewhat affect the performance of origin attribution, our framework consistently outperforms the baseline.

\begin{table}[t]
    \centering
    \scriptsize
    \caption{Evaluation of OCC-CLIP with Various ADA Approaches. `Non-Target' applies ADA to the non-target images; `T' applies it to the target images; `Both' applies it to both the target and non-target images; and `T-NT' treats part of the target images as non-target ones post-ADA. The results are averaged across seven testing tasks.}
    \vspace{-0.3cm}
        \setlength{\tabcolsep}{12pt}
        \begin{tabular}{@{}lccccc@{}}
        \toprule[1pt]
        Methods & None & Non-Target & Both & Target & T-NT \\
        \midrule
        OCC-CLIP 
        & $0.9263$
        & $\mathbf{0.9548}$
        & $0.9080$
        & $0.8236$
        & $0.9514$ \\
        \bottomrule[1pt]
        \end{tabular}
    
    \label{table: Different Attacks}
     \vspace{-0.2cm}
\end{table}

\vspace{0.05cm}\noindent\textbf{Sensitivity to ADA Settings.} We employ four methods for applying ADA: 1) `Non-Target': on half of the non-target images; 2) `Target': on half of the target images; 3) `Both': on half of both the non-target images and target images; 4) `T-NT': on half of the target images which are then treated as non-target ones. The non-target images are selected from the COCO dataset, while the target images are from SD.

As shown in Table~\ref{table: Different Attacks}, it is evident that applying ADA on non-target images yields the best performance. This outcome is reasonable, as gradient ascent on the non-target image set enlarges the learned space of this set and further approximates the boundary to the target image set. Applying ADA on `T-NT' is the next most effective method, performing only slightly worse than `Non-Target'. However, the boundary learned by `T-NT' may be too tight for effective origin attribution tasks. Applying ADA on `Both' is less effective than doing nothing, as `Both' also enlarges the learned distribution of the target image set. Solely applying ADA on the target set performs significantly worse than doing nothing. This is because this approach not only enlarges the learned distribution space of the target images but also shrinks the learned distribution space of the non-target image set, leading to a higher likelihood of misclassifying many images as target images.

\begin{table}[t]
    \centering
    \scriptsize
    \caption{Evaluation of OCC-CLIP Relative to the Proportion of Augmented Non-Target Images. Five proportions are investigated: \(0\%\), \(25\%\), \(50\%\), \(75\%\), and \(100\%\). The results are averaged across seven testing tasks.}
    \vspace{-0.3cm}
        \setlength{\tabcolsep}{14pt}
        \begin{tabular}{@{}lccccc@{}}
        \toprule[1pt]
        Methods & 0\% & 25\% & 50\% & 75\% & 100\% \\
        \midrule
        OCC-CLIP & 0.9263 
            & $0.9405$ 
            & $0.9548$ 
            & $\mathbf{0.9558}$ 
            & $0.9172$ \\
        \bottomrule[1pt]
        \end{tabular}
    \label{table: Different Proportion}
     \vspace{-0.4cm}
\end{table}

\vspace{0.05cm}\noindent\textbf{Sensitivity to the Proportion of Augmented Non-target Images.} We further investigate the influence of the proportion of augmented non-target images on the performance of OCC-CLIP. As indicated in Table~\ref{table: Different Proportion}, the performance of origin attribution is optimal when 50\% to 75\% of non-target images are augmented. However, when all data are augmented, the performance deteriorates, even falling below that with 0\% non-target images augmented. This may be because augmenting all data could lead to a loss of the original distribution space of the non-target image set.

\begin{figure}[t]
    \centering
    \scriptsize
    \includegraphics[width=0.99\linewidth]{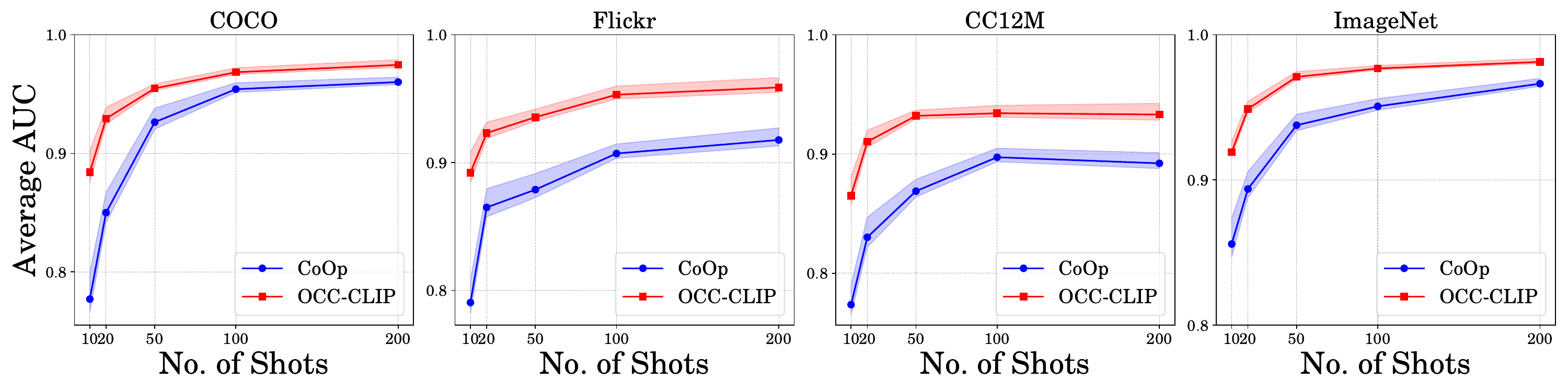}
    \vspace{-0.3cm}
    \caption{Evaluation of OCC-CLIP on Different Numbers of Shots. This figure shows the average origin attribution performance of CoOp and OCC-CLIP on 7 different testing tasks with a variable number of shots: 10, 20, 50, 100, and 200. The target dataset is from SD. The non-target datasets are from COCO, CC12M, Flickr, or ImageNet.}
    \label{fig: Num of Shots}
     \vspace{-0.2cm}
\end{figure}

\vspace{0.05cm}\noindent\textbf{Sensitivity to the Number of Shots.} We then analyze how the performance of OCC-CLIP relates to the number of shots. Our investigation covers the effects of using 10, 20, 50, 100, and 200 shots. In these experiments, SD is used for the target images, while one of COCO, ImageNet, Flickr, and CC12M is employed as the non-target dataset. Other generated image datasets are used for testing. As shown in Figure~\ref{fig: Num of Shots}, OCC-CLIP consistently outperforms CoOp in all scenarios, particularly when the number of shots is small. Additionally, the incremental improvements in AUC tend to diminish as more shots are added, eventually plateauing at an upper boundary when the number of shots reaches 200.

\begin{table*}[t]
    \centering
    \scriptsize

    \caption{Evaluation of OCC-CLIP under Image Processing. Six image processing methods are executed: Gaussian Blur, Gaussian Noise, Grayscale, Rotation, Flip, and a mixture of all these data augmentation methods. The results are averaged across seven testing tasks.}%
    \vspace{-0.3cm}
      \setlength{\tabcolsep}{3pt}
      \begin{tabular}{l|ccccccccccc}
        \toprule[1pt]
        Method
        & None
        & Gaussian Blur
        & Gaussian Noise
        & Grayscale
        & Rotation
        & Flip
        & Mixture \\
        \midrule
        CoOp 
        & $0.9263$
        & $0.8539$
        & $0.8859$
        & $0.8017$
        & $0.9040$
        & $0.9241$
        & $0.7457$ \\ 
        OCC-CLIP 
        & $\mathbf{0.9548}$
        & $\mathbf{0.9002}$
        & $\mathbf{0.9089}$
        & $\mathbf{0.8405}$
        & $\mathbf{0.9337}$
        & $\mathbf{0.9479}$
        & $\mathbf{0.7538}$ \\
        \bottomrule[1pt]
        \end{tabular}
        \label{Table: Defence Against Image Processing}
        \vspace{-0.4cm}
\end{table*}

\vspace{0.05cm}\noindent\textbf{Sensitivity to Image Processing in Verification Stage.} Table~\ref{Table: Defence Against Image Processing} presents a comparative analysis of the performance between OCC-CLIP and baseline methods in the face of potential image processing: Gaussian Blur, Gaussian Noise, Grayscale, Rotation, Flip, or a mixture of those attacks. The data indicate that OCC-CLIP exhibits superior robustness compared to the baseline across a range of potential processing of input images. This enhanced resilience of OCC-CLIP underscores its effectiveness in safeguarding against various forms of image manipulation, highlighting its utility in origin attribution.

\vspace{0.05cm}\noindent\textbf{Sensitivity to Choice of Prompts.} Since we focus on the one-class classification question, the classes in our setting can be labeled as any contrasting pairs, such as negative vs. positive. Therefore, we explore the effect of the choice of prompts on the performance of OCC-CLIP. As shown in the supplementary material, the choice of prompts can have some different effects. However, regardless of which pair of prompts is chosen, OCC-CLIP always outperforms the baseline.

\begin{table}[t]
        \centering
        \scriptsize
        \caption{ADA vs. other data augmentation methods. Different ways of data augmentation are used during training. The results are averaged across seven testing tasks.} 
        \vspace{-0.3cm}
        \begin{tabular}{@{}l|cccccccc@{}}
        \toprule
        Methods & Gaussian Blur & Gaussian Noise & Grayscale & Rotation & Flip & Mixture & None  & ADA \\
        \midrule
        OCC-CLIP & $0.8639$ & $0.8340$ & $0.8308$ & $0.8465$ & $0.8422$ & $0.7991$  & 0.9263 & \textbf{0.9548} \\
        \bottomrule
        \end{tabular}
    
    \label{Table: Different Data Aug}
    \vspace{-0.2cm}
\end{table}

\subsection{Comparison to Standard Data Augmentation} Data augmentation methods have been extensively developed. In real-fake detection tasks, Wang~\cite{wang2020cnn} and CR~\cite{chandrasegaran2022discovering} utilized a combination of various data augmentation methods to improve model performance. To conduct a more comprehensive evaluation of ADA, we compare its performance with other common data augmentation methods: Gaussian Blur, Gaussian Noise, Grayscale, Rotation, Flip, and a mixture of all these. According to Table~\ref{Table: Different Data Aug}, all traditional data augmentation methods perform poorly, and in some cases, even degrade the model's performance. Only ADA shows some improvements. Consequently, we can infer that normal data augmentation methods are not suitable for few-shot one-class classification scenarios.

\begin{figure}[t]
    \centering
    \footnotesize
    \begin{subfigure}[b]{0.49\linewidth}
        \centering
        \includegraphics[scale=0.25]{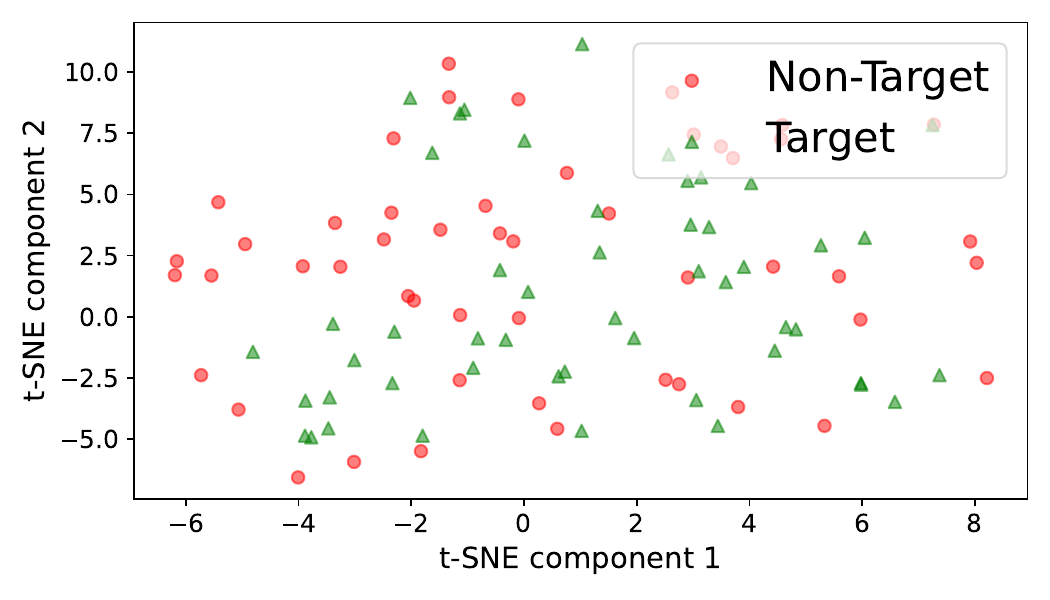}
        \vspace{-0.1cm}
        \caption{Without ADA}
        \label{fig: Vb}
    \end{subfigure}
    \begin{subfigure}[b]{0.49\linewidth}
        \centering
        \includegraphics[scale=0.25]{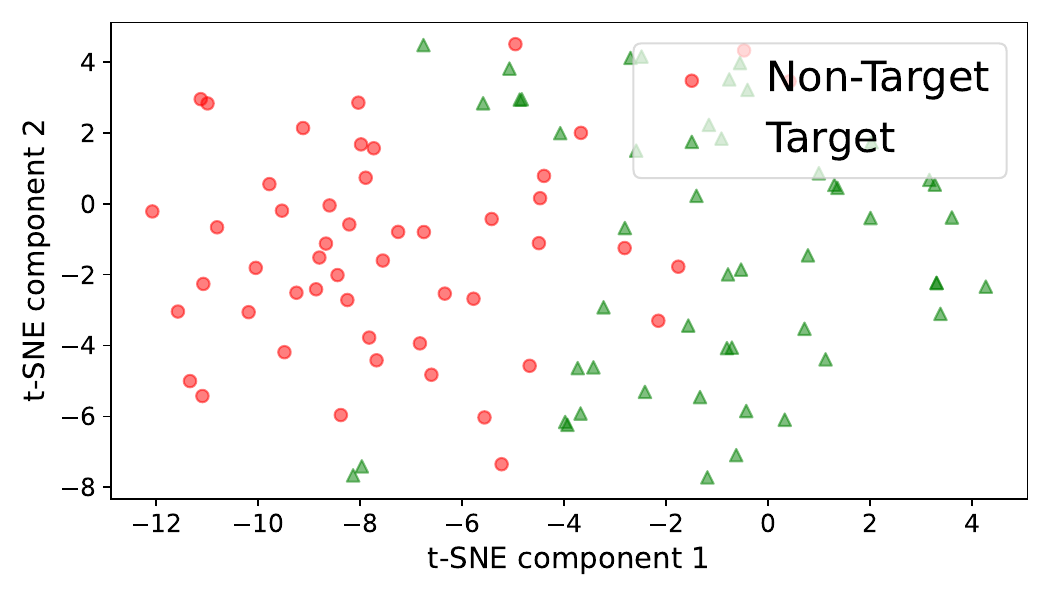}
        \vspace{-0.1cm}
        \caption{With ADA}
        \label{fig: Vc}
    \end{subfigure}
    \vspace{-0.3cm}
    \caption{Visualize the 2-dimensional mapping of image features with t-SNE~\cite{van2008visualizing}. The target images are from SD, and the non-target images are from COCO. Figure (a) shows the distribution of non-target images and target images without using ADA. Figure (b) shows the distribution of augmented non-target images and target images after using the ADA technique.}
    \vspace{-0.4cm}
\end{figure}

\subsection{Understanding Adversarial Data Augmentation}
We utilize the embeddings from CLIP's image encoder for visualization. According to Figure~\ref{fig: Vb}, it is evident that there is no clear boundary between the target image set and the non-target image set, i.e., images from SD and images from COCO. However, after applying ADA, as illustrated in Figure~\ref{fig: Vc}, a clearer boundary emerges between the augmented non-target image set and the target image set, demonstrating the effectiveness of ADA. During training, unchanged non-target images and augmented non-target images are mixed. This approach not only preserves the properties of the original non-target images but also ensures that the augmented non-target images approximate the boundary of the distribution of the target images.

\begin{table}[t]
    \centering
    \scriptsize
    \caption{Evaluation of OCC-CLIP with commercial generation API. The target images are generated by DALL·E-3. The non-target images are from COCO. The results are averaged across eight testing tasks.}
    \vspace{-0.3cm}
        \setlength{\tabcolsep}{16pt}
        \begin{tabular}{@{}lccccccccc@{}}
        \toprule
        Methods & 10 & 20 & 30 
        & 50 \\
        \midrule
        CoOp 
        & 0.9216 
        & 0.9687 
        & 0.9563 
        & 0.9788 \\
        OCC-CLIP 
        & $\mathbf{0.9754}$ 
        & $\mathbf{0.9848}$ 
        & $\mathbf{0.9936}$ 
        & $\mathbf{0.9959}$ \\
        \bottomrule
        \end{tabular}
    
    \label{table: Detect DALLE3}
    \vspace{-0.2cm}
\end{table}

\subsection{Source Model Attribution with Commercial Generation API}
To investigate the effectiveness of OCC-CLIP in the real world, we utilize the latest commercial digital image-generating model -- DALL·E-3~\cite{betker2023improving} -- to generate images. As shown in the supplementary material, 103 prompts are randomly selected from the annotations of the validation set of COCO\cite{lin2014microsoft}, resulting in a total of 200 images generated. The default setting of DALL·E-3 is to generate 2 images simultaneously. However, due to OpenAI's content policy, some prompts can only generate one image. In this scenario, the targets are images generated by DALL·E-3, while the non-targets are images from other datasets. We compare the performance of detection across different numbers of shots: 10, 20, 30, 50. According to Table~\ref{table: Detect DALLE3}, our method consistently outperforms the baseline and performs well even with as few as 10 shots.

\subsection{Model Attribution with Multiple Source Models}
In addition to verifying single models, we also investigate multi-source origin attribution conditions. We explore the process of verifying multiple sources using the eight trained one-class classifiers. Since this involves multi-source origin attribution, the evaluation metric used in this section is \textbf{accuracy} instead of AUC. We evaluate the model under the following conditions: 2 classes (SD, VQ-D), 4 classes (SD, VQ-D, LDM, Glide), and 6 classes (SD, VQ-D, LDM, Glide, GALIP, ProGan). As shown in Table~\ref{Table: OCC Multi Class}, the accuracy decreases with an increasing number of source models. In every multi-class classification scenario mentioned above, one-class classifiers trained with OCC-CLIP consistently outperform those trained with CoOp.

Furthermore, our methods can be adapted to directly train a multi-class classifier. As demonstrated in the table included in the supplementary material, the multi-class classifier trained using the OCC-CLIP approach is still shown to be more effective compared to the one trained using the CoOp approach.

\begin{table*}[t]
    \centering
    \scriptsize
        \caption{Employing an ensemble of one-class classifiers for multi-class classification tasks. The following scenarios are considered: 2 classes, 4 classes, 6 classes, and 8 classes.}
        \vspace{-0.3cm}
        \begin{tabular}{l|c|cccccccccc}
        \toprule[1pt]
        Num of Class & Method 
        & LDM
        & Glide
        & GALIP
        & ProGAN
        & StyleGAN2
        & GauGAN
        & Overall \\
        \midrule
        \multirow{2}{*}{2 classes} 
            & CoOp
            & $0.6366$
            & $0.6448$
            & $0.6378$
            & $0.6928$
            & $0.8706$
            & $0.7259$
            & $0.7014$ \\ 
            & OCC-CLIP 
            & $\mathbf{0.6449}$
            & $\mathbf{0.6499}$
            & $\mathbf{0.6457}$
            & $\mathbf{0.8742}$
            & $\mathbf{0.8938}$
            & $\mathbf{0.9255}$
            & $\mathbf{0.7723}$ \\ 
        \midrule 
        \multirow{2}{*}{4 classes} 
            & CoOp
            & -- 
            & -- 
            & $0.5863$
            & $0.6124$
            & $0.7004$
            & $0.6376$
            & $0.6342$ \\ 
            & OCC-CLIP 
            & -- 
            & -- 
            & $\mathbf{0.6130}$
            & $\mathbf{0.7461}$
            & $\mathbf{0.7319}$
            & $\mathbf{0.7908}$
            & $\mathbf{0.7204}$ \\ 
        \midrule
        \multirow{2}{*}{6 classes} 
            & CoOp 
            & -- 
            & -- 
            & -- 
            & -- 
            & $0.6459$
            & $0.6092$
            & $0.6276$ \\ 
            & OCC-CLIP 
            & -- 
            & -- 
            & -- 
            & -- 
            & $\mathbf{0.6587}$
            & $\mathbf{0.6727}$
            & $\mathbf{0.6657}$ \\ 
        \bottomrule[1pt]
        \end{tabular}
        \vspace{-0.4cm}
        \label{Table: OCC Multi Class}
\end{table*}

\section{Conclusions}
In this work, we study origin attribution in a practical setting where only a few images generated by a source model are available and the source model cannot be accessed. The introduced problem is first formulated as a few-shot one-classification task. A simple yet effective solution CLIP-based framework is proposed to solve the task. Our experiments on both open-source popular generative models and commercial generation API shows the effectiveness of our framework. Our OCC-CLIP framework can also be applied to solve the few-shot one-classification task in other domains, which we leave in future work. Another future work is to evaluate the natural and adversarial robustness of our framework~\cite{chen2024benchmarking,cheng2024unveiling,gu2022vision,gu2020improving} and build adversarially robust variants~\cite{wu2022towards,jia2024revisiting}.

\clearpage  

\noindent \textbf{Acknowledgement:} This work is supported by the UKRI grant: Turing AI Fellowship EP/W002981/1, EPSRC/MURI grant: EP/N019474/1. We thank the Royal Academy of Engineering

%
%
\bibliographystyle{splncs04}
\bibliography{main}

\appendix

\section{Generation of Datasets}
\begin{figure*}[t]
    \centering
    \footnotesize
    \begin{tabular}{l|ccccc|c} & SD & VQ-D & LDM & Glide & GALIP & DALL·E-3 \\

        \raisebox{0.1cm} {\rotatebox{90}{Caption 1}} &
        \includegraphics[width=0.13\textwidth]{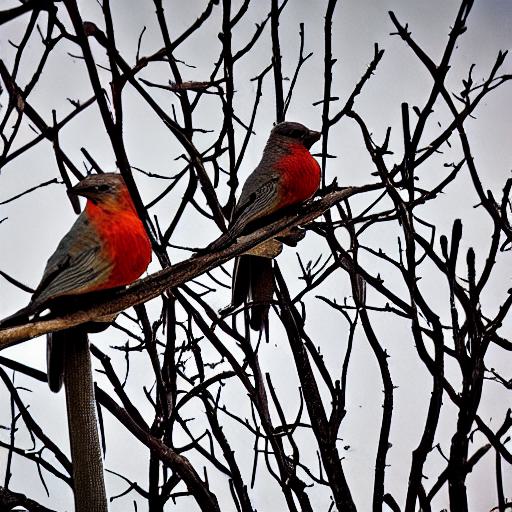} & 
        \includegraphics[width=0.13\textwidth]{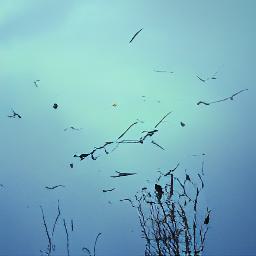} & 
        \includegraphics[width=0.13\textwidth]{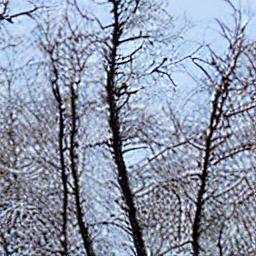} & 
        \includegraphics[width=0.13\textwidth]{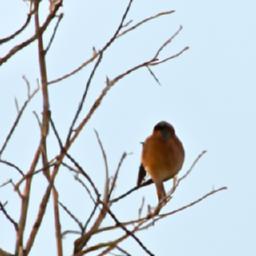} &
        \includegraphics[width=0.13\textwidth]{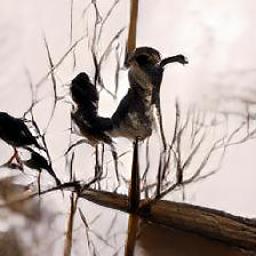} 
        & \includegraphics[width=0.13\textwidth]{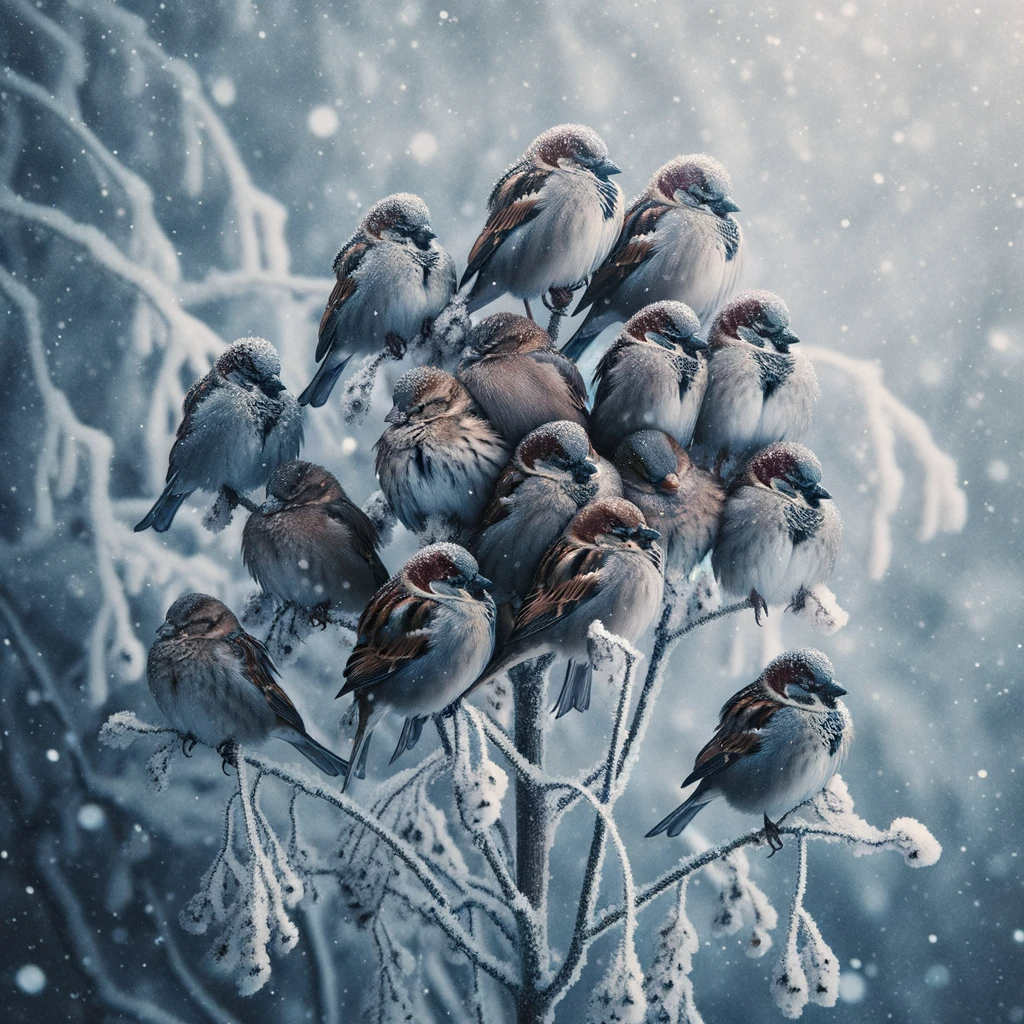}
        \\ 
    
        \raisebox{0.1cm} {\rotatebox{90}{Caption 2}} &
        \includegraphics[width=0.13\textwidth]{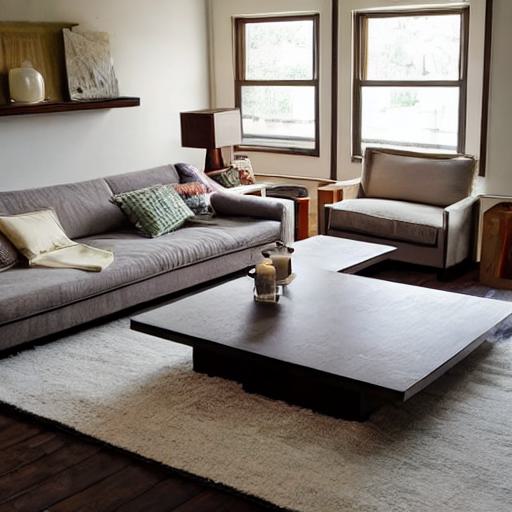} & 
        \includegraphics[width=0.13\textwidth]{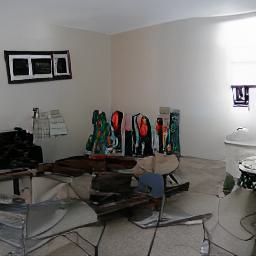} &
        \includegraphics[width=0.13\textwidth]{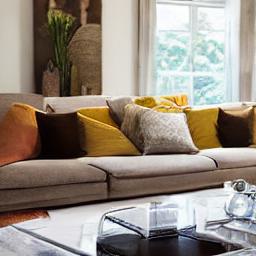} & 
        \includegraphics[width=0.13\textwidth]{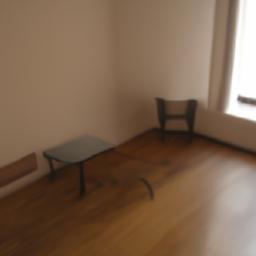} &
        \includegraphics[width=0.13\textwidth]{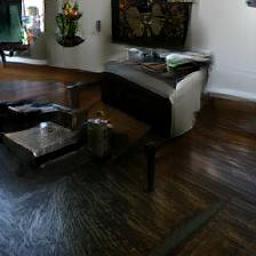} 
        & \includegraphics[width=0.13\textwidth]{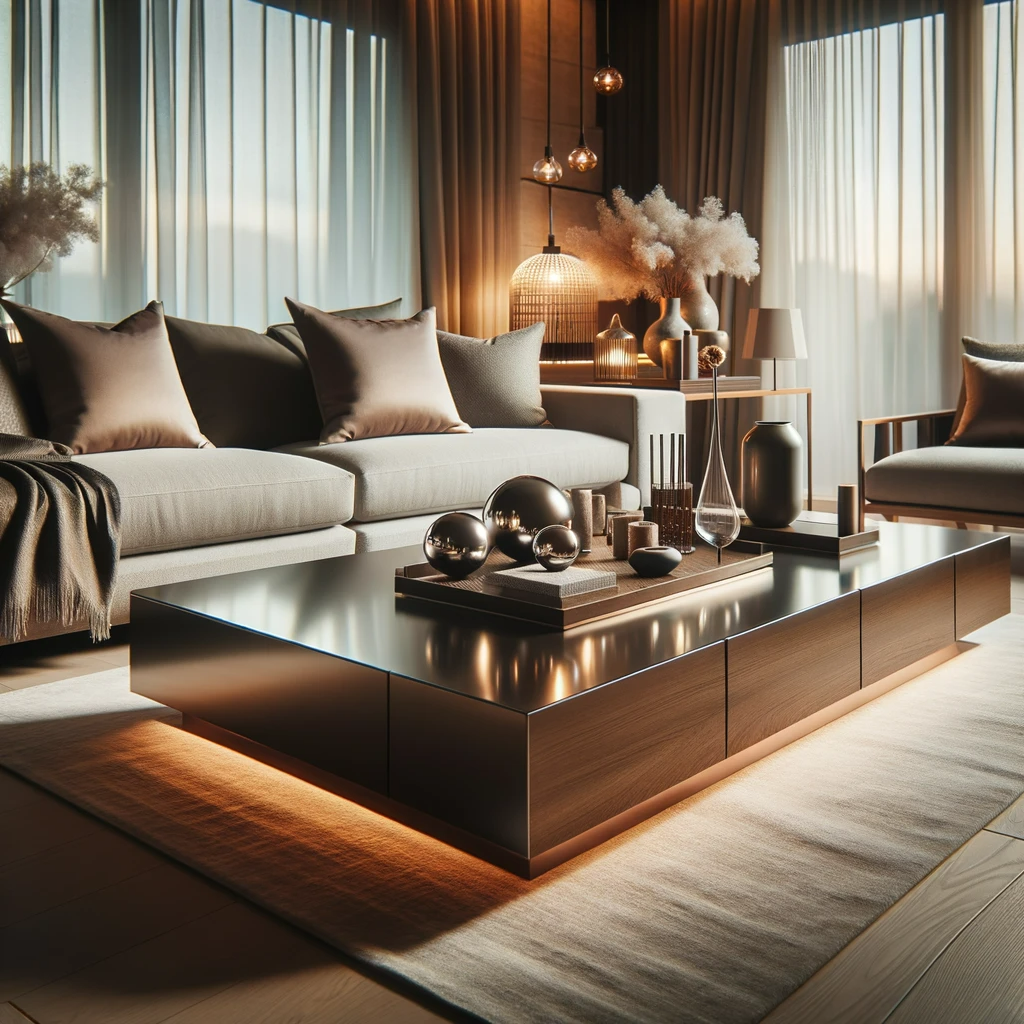} 
        \\ 
    
        \raisebox{0.1cm} {\rotatebox{90}{Caption 3}} &
        \includegraphics[width=0.13\textwidth]{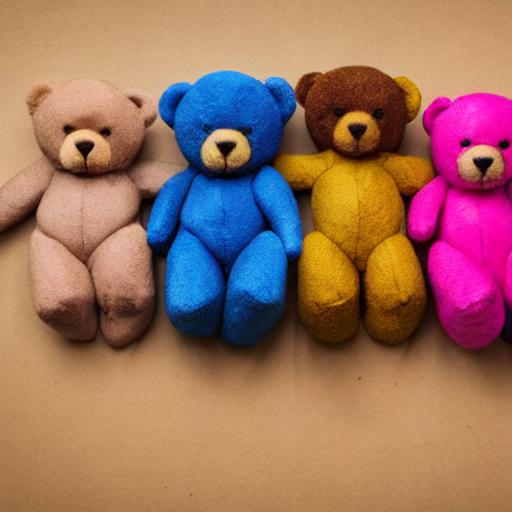} & 
        \includegraphics[width=0.13\textwidth]{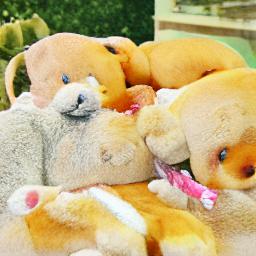} &
        \includegraphics[width=0.13\textwidth]{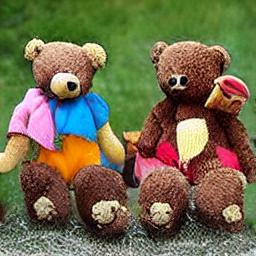} & 
        \includegraphics[width=0.13\textwidth]{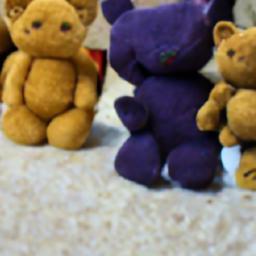} &
        \includegraphics[width=0.13\textwidth]{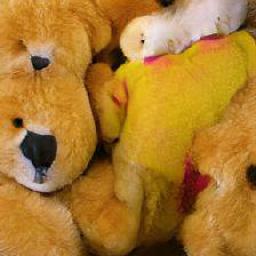} 
        & \includegraphics[width=0.13\textwidth]{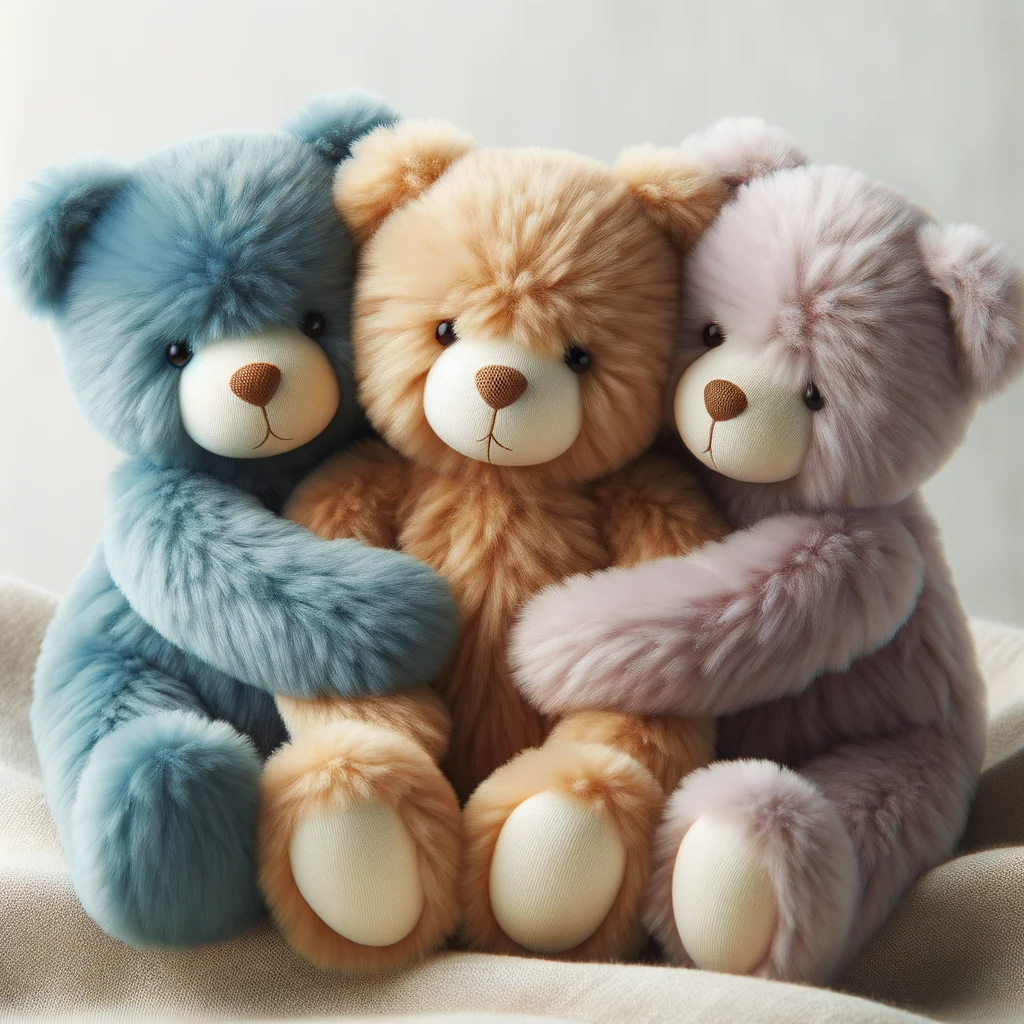} 
        \\ 
    
        \raisebox{0.1cm} {\rotatebox{90}{Caption 4}} &
        \includegraphics[width=0.13\textwidth]{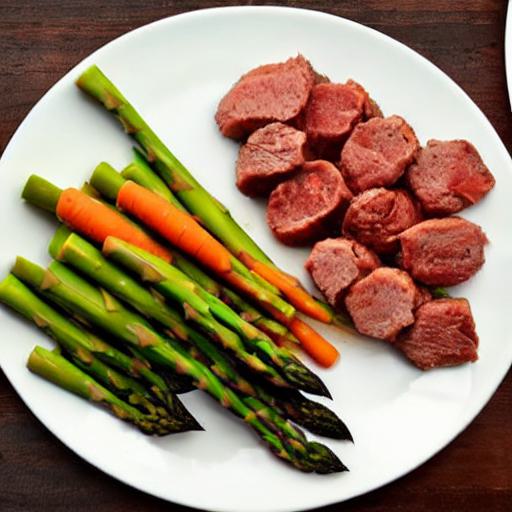} & 
        \includegraphics[width=0.13\textwidth]{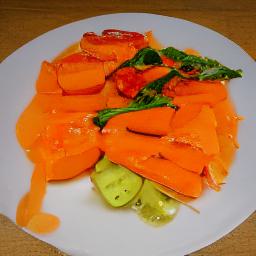} &
        \includegraphics[width=0.13\textwidth]{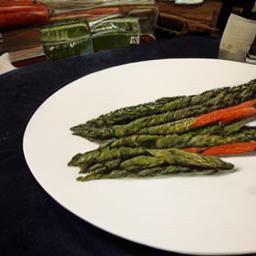} & 
        \includegraphics[width=0.13\textwidth]{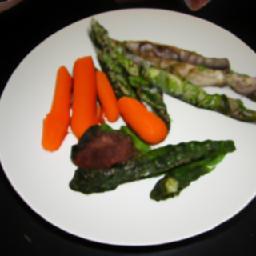} &
        \includegraphics[width=0.13\textwidth]{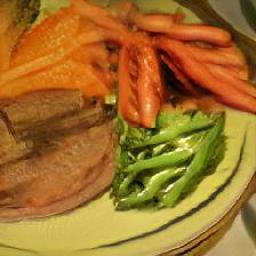} 
        & \includegraphics[width=0.13\textwidth]{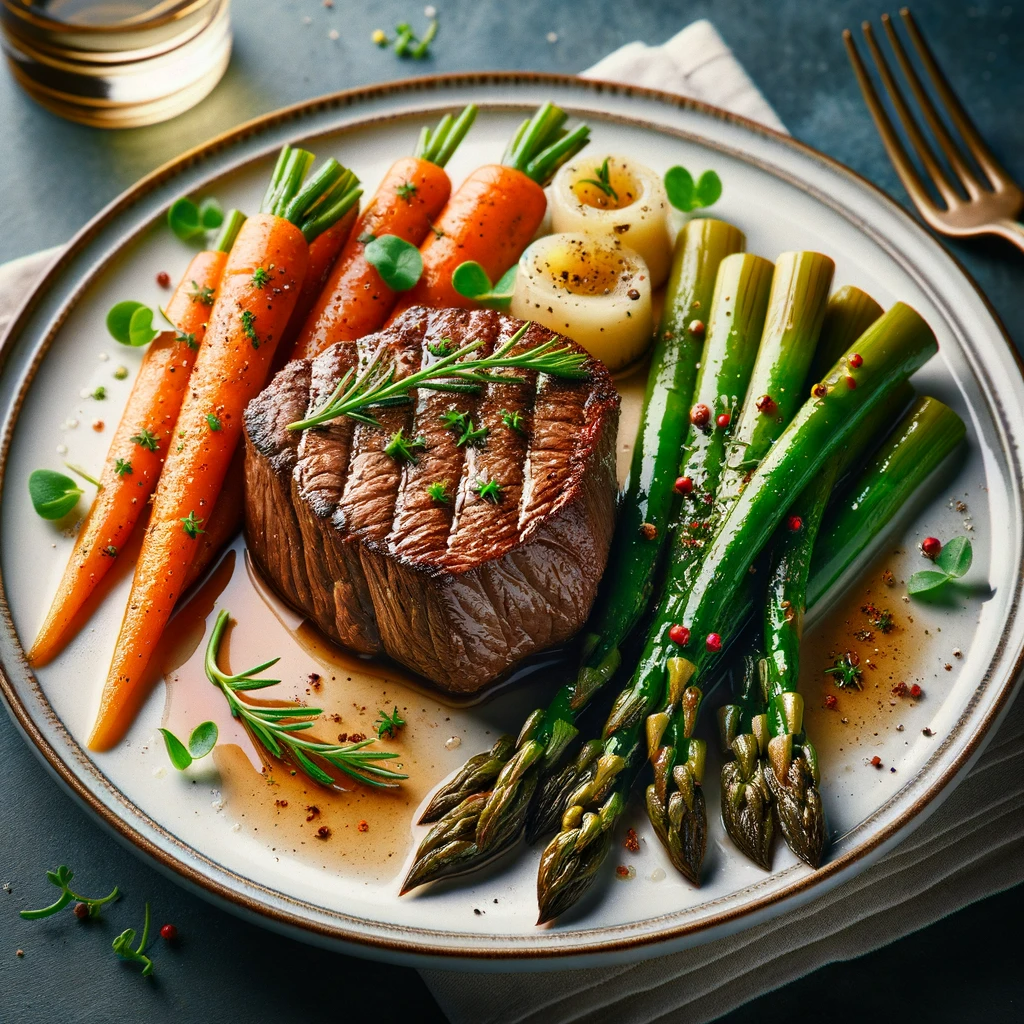} 
        \\

        \raisebox{0.1cm} {\rotatebox{90}{Caption 5}} &
        \includegraphics[width=0.13\textwidth]{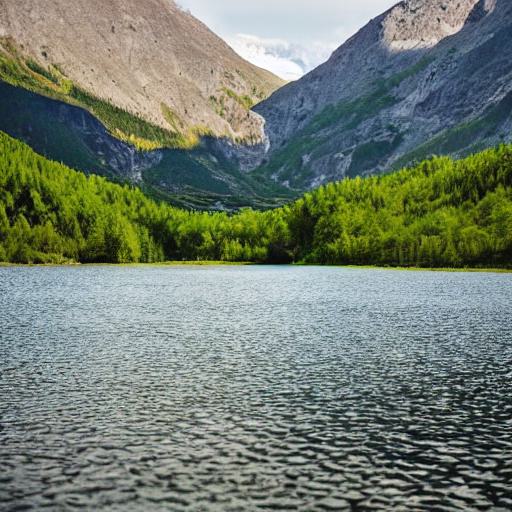} & 
        \includegraphics[width=0.13\textwidth]{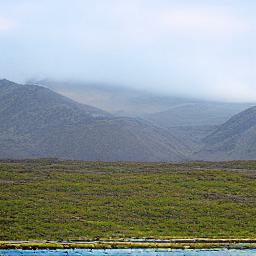} &
        \includegraphics[width=0.13\textwidth]{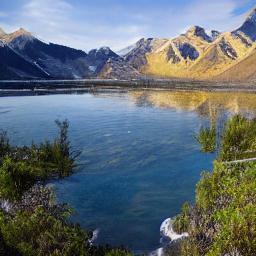} & 
        \includegraphics[width=0.13\textwidth]{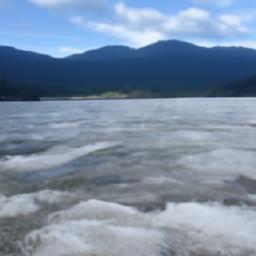} &
        \includegraphics[width=0.13\textwidth]{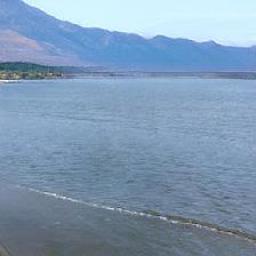} 
        & \includegraphics[width=0.13\textwidth]{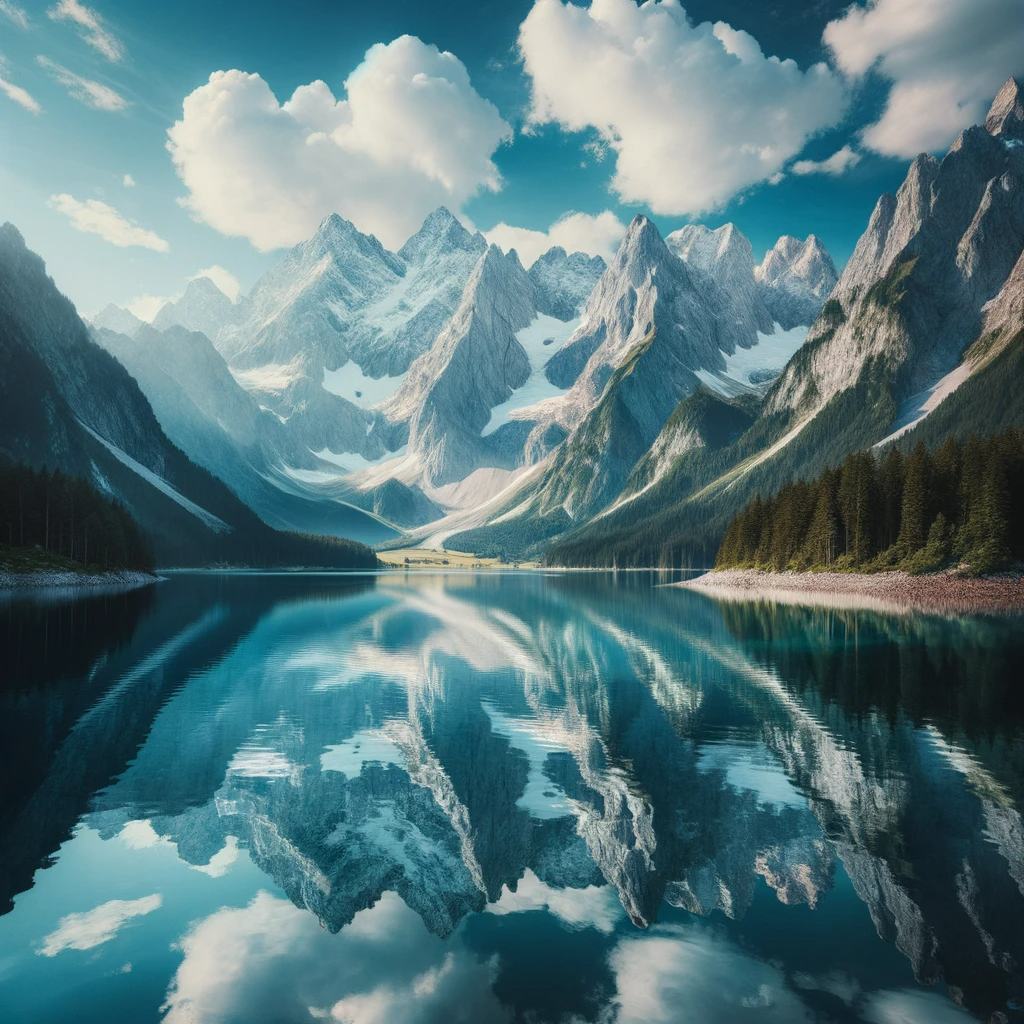} 
        \\

    \end{tabular} 
    \caption{
    This figure presents a demo of synthesized images produced by six distinct models, each predicated upon four specific captions. 
    Caption 1: \textit{Birds perch on a bunch of twigs in the winter}. 
    Caption 2: \textit{A coffee table sits in the middle of a living room}.
    Caption 3: \textit{Three teddy bears, each a different color, snuggling together}.
    Caption 4: \textit{The dinner plate has asparagus, carrots and some kind of meat}.
    Caption 5: \textit{A large body of water sitting below a mountain range}.
    }
\label{fig: visualisation}
\end{figure*}

There are a total of 202,520 fake images generated by five different generative models, namely, Stable Diffusion Model~\cite{rombach2022high}, Latent Diffusion Model~\cite{rombach2022high}, GLIDE~\cite{nichol2021glide}, Vector Quantized Diffusion~\cite{gu2022vector}, and GALIP~\cite{tao2023galip}. The prompts used are the first captions of each image in the validation set of the Microsoft Common Objects in Context (COCO) 2014 dataset \cite{lin2014microsoft}. These models were pre-trained on four different datasets: LAION-5B~\cite{schuhmann2022laion}, COCO~\cite{lin2014microsoft}, LAION-400M~\cite{schuhmann2021laion}, and filtered CC12M~\cite{changpinyo2021conceptual}. In total, as shown in Figure~\ref{fig: visualisation}, five synthetic image datasets are generated, namely SD, VQ-D, LDM, GLIDE, and GALIP. 

\textbf{SD:} This marks the data generated by the Stable Diffusion Model~\cite{rombach2022high}\footnote{\url{https://github.com/CompVis/stable-diffusion}}. It was pre-trained on the LAION-5B dataset. The size of the generated images is \(512 \times 512\). The pre-trained model used is `sd-v1-4'.

\textbf{VQ-D:} This marks the data generated by the Vector Quantized Diffusion (VQ-D) Model \cite{gu2022vector}\footnote{\url{https://github.com/microsoft/VQ-Diffusion}}. It was pre-trained on the COCO dataset. The size of the generated images is \(256 \times 256\). The pre-trained model `coco\_pretrained' served as the backbone of this method.

\textbf{LDM:} This marks the data generated by the Latent Diffusion Model \cite{rombach2022high}\footnote{\url{https://github.com/CompVis/latent-diffusion}}. It was pre-trained on the LAION-400M dataset. The size of the generated images is \(256 \times 256\). The pre-trained model, `txt2img-f8-large', is utilized.

\textbf{GLIDE:} This marks the data generated by GLIDE \cite{nichol2021glide}\footnote{\url{https://github.com/openai/glide-text2im}}. It was pre-trained on filtered CC12M. The size of the generated images is \(256 \times 256\).

\textbf{GALIP:} This marks the dataset generated by GALIP~\cite{tao2023galip}\footnote{\url{https://github.com/tobran/GALIP}}. The model was pretrained on the COCO dataset. The size of the generated images is \(256 \times 256\).

We also utilized pre-existing datasets (namely GauGAN~\cite{park2019semantic}, ProGAN~\cite{karras2017progressive}, and StyleGAN2~\cite{karras2020analyzing}) as provided by~\cite{wang2020cnn}. GauGAN was trained on the COCO~\cite{lin2014microsoft} dataset, while ProGan and StyleGan2 were trained on the LSUN~\cite{yu2015lsun} dataset. The size of images from those datasets is \(256 \times 256\).

\section{Implementation Details}
For all models, we randomly selected 1,000 images from each dataset for testing. Besides, 500 images were chosen from each dataset for training, divided into 10 sets: train1, train2, \ldots, train10. Each training set contained 50 images from a non-target dataset and 50 from a target dataset. After training for 200 epochs, we evaluated the results on a test set comprising 1,000 testing images from both the non-target and target datasets. There was no dedicated validation set in our experimental setup. By default, adversarial data augmentation was applied to half of the non-target images, with a perturbation step size, denoted as $\epsilon$, of 0.1. The Average AUC value was calculated as the mean of the AUC values obtained from the 10 trained models, each trained on a distinct training set and tested on the same test set.

\section{Comparison with Baselines}
The standard deviation of each mean of AUC is shown in Table~\ref{table: 14 baselines sup} and Table~\ref{table: 14 baselines sup2}. The standard deviation of each mean of \textbf{Accuracy} is shown in Table~\ref{table: 14 baselines sup accuracy} and Table~\ref{table: 14 baselines sup2 accuracy}.The reason why the CLIP model has a 0 standard deviation is that it is a zero-shot model, which means there are no parameters changed. Therefore, when testing on the same testing set, the result will not change.

ResNet \cite{he2016deep}, Inception \cite{szegedy2016rethinking}, DenseNet \cite{huang2017densely}, VGG~\cite{simonyan2014very}, ViT \cite{dosovitskiy2020image}, Deit \cite{touvron2021training}, Cait \cite{touvron2021going}, and Swin \cite{liu2021swin} are either milestones or leading-edge models in the field of Computer Vision. For the Image-Patch model, each image is subdivided into 2x2 patches, with each patch serving as a separate input to the ResNet-50 architecture. The ultimate prediction from the model is the average of the predictions made for each of these patches. In the case of the Feature-Patch model, which shares the same backbone, the last five layers of the standard ResNet-50 architecture are discarded, and the remaining structure is employed to extract image features. Subsequently, these features are segmented into 4x4 patches, with each patch being passed through the final linear classification layer. The model's final output is the average prediction over these patches.

\begin{figure*}[t]
    \centering
    \scriptsize
    \includegraphics[width=0.9\textwidth]{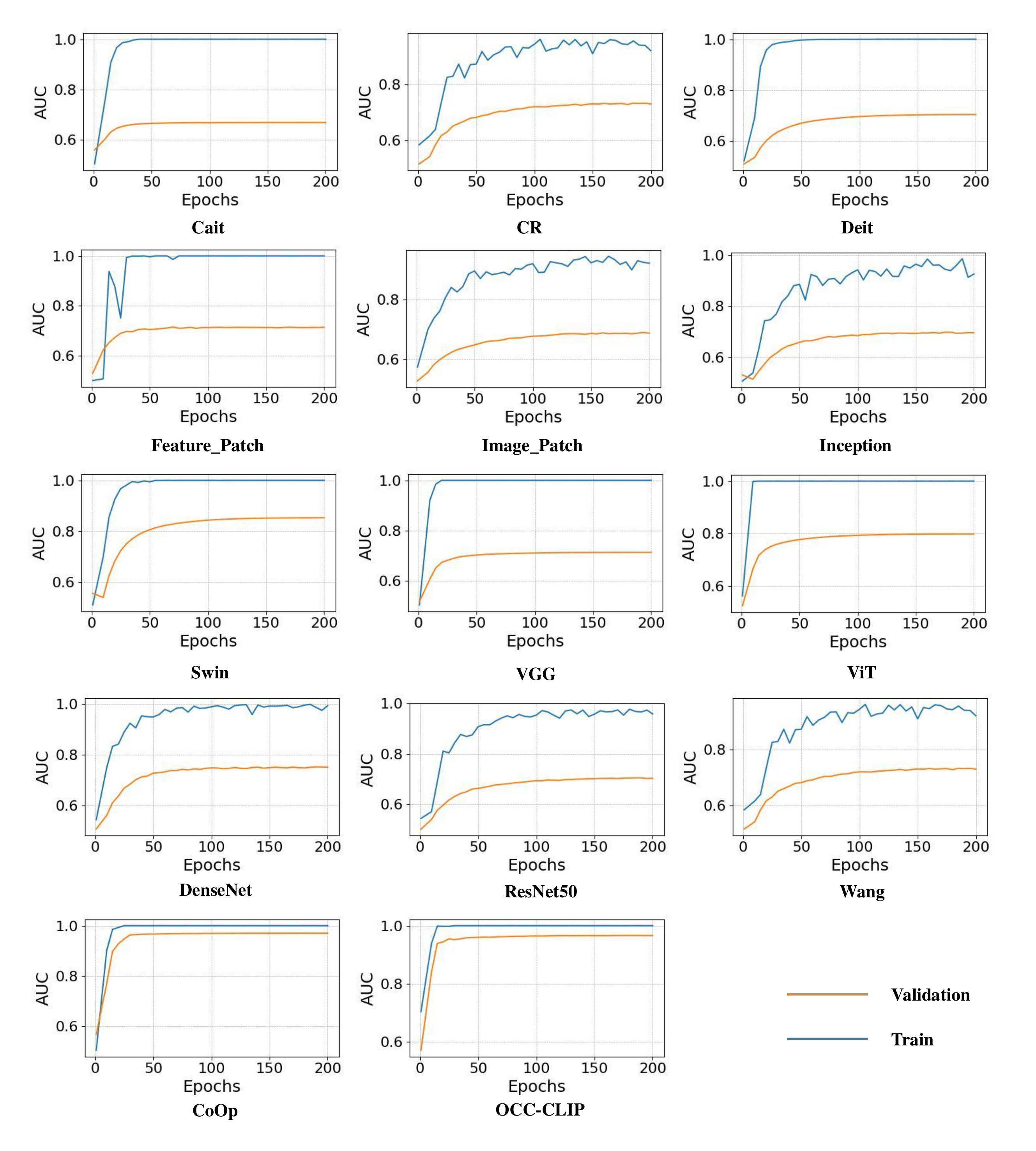}
    \vspace{-0.5cm}
    \caption{This figure illustrates the training and validation Area Under the Curve (AUC) metrics for 14 models (13 baselines + OCC-CLIP), comparing their performance in terms of both training accuracy and validation reliability.}
    \label{fig: CVPR AUC}
    \vspace{-0.3cm}
\end{figure*}

\begin{figure*}[t]
    \centering
    \scriptsize
    \includegraphics[width=0.9\textwidth]{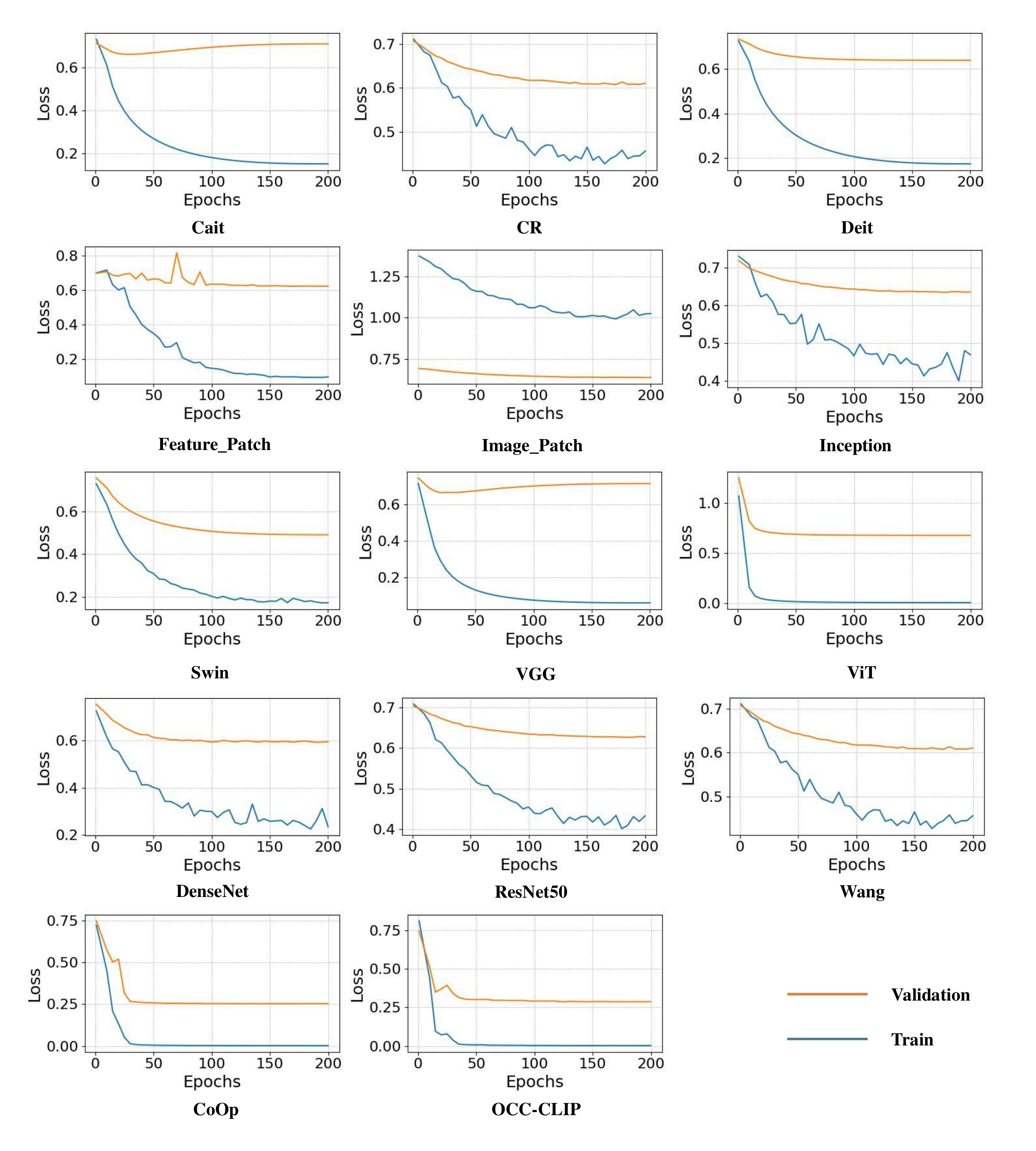}
    \vspace{-0.5cm}
    \caption{This figure illustrates the comparative analysis of training and validation loss across 14 models (13 baselines + OCC-CLIP). It provides a visual representation of how each model's loss metrics evolve over the course of training.}
    \label{fig: CVPR Loss}
    \vspace{-0.3cm}
\end{figure*}

To visualize the training loss, training AUC, validation loss, and validation AUC, apart from images in the training set and test set, I randomly select 1000 images from both non-target and target datasets to build a validation set. Figure~\ref{fig: CVPR AUC} shows the change of loss of the train set and validation set with the increasing number of epochs. Figure~\ref{fig: CVPR Loss} shows the change of AUC of the train set and validation set with the increasing number of epochs. The reason why the loss of the train set is larger than the loss of the validation set in the Image\_Patch model is that during training, each image is subdivided into 4 patches.

\section{Sensitivity to Source Models and Value of Epsilon}
To test the effect of the choice of $\epsilon$ on source models, we conducted a comprehensive evaluation using eight models with six different values of $\epsilon$. According to Table~\ref{table: Different Source Models Sup} and Table~\ref{table: Different Source Models Sup2}, it can be concluded that varying the step size $\epsilon$ has different effects on source model attribution. With the adjustment of $\epsilon$, it was observed that the use of adversarial data augmentation can have a positive effect on source model attribution tests.

\section{Sensitivity to Selection of Non-target Class and ADA Settings}
Table~\ref{table: Different Ways of Attacks Sup} and Table~\ref{table: Different Ways of Attacks Sup2} show the sensitivity evaluation of various ADA methods on different open-world real image datasets. In the default setting, the target dataset is SD, and the non-target is COCO. ADA is applied to half of the non-target images. 1) During the training phase, conditioned on applying ADA to only half of the non-target images, three additional non-target image datasets are used: ImageNet, Flickr, and CC12M. 2) When selecting non-target images exclusively from COCO, ADA is applied to half of the target images, to half of both non-target and target images, and to half of the target images, which are then treated as non-target. It is observed that the choice of non-target images affects the performance of OCC-CLIP. Additionally, applying ADA to non-target images is most effective.

\section{Sensitivity to the Proportion of Augmented Non-target Images}
Table~\ref{table: Different Proportion Sup} and Table~\ref{table: Different Proportion Sup2} present the standard deviation when applying ADA to varying proportions of non-target images. Results from seven different testing tasks and their average are also included.

\section{Sensitivity to the Number of Shots}
Table~\ref{table: Number of Shots Sup} and Table~\ref{table: Number of Shots Sup2} illustrate the mean origin attribution performance of CoOp and OCC-CLIP across seven different testing tasks with varying numbers of shots: 10, 20, 30, 40, 50, 100, and 200. In the training phase, SD is used as the target dataset, and one of COCO, CC12M, Flickr, or ImageNet is used as the non-target dataset.

\section{Sensitivity to Image Processing in Verification Stage}
Tables~\ref{Table: Defence Against Potential Attack Sup} and \ref{Table: Defence Against Potential Attack Sup2} present a comparative analysis of the performance between OCC-CLIP and baseline methods in response to potential image processing techniques: Gaussian Blur, Gaussian Noise, Grayscale, Rotation, Flip, or a combination of these attacks. The standard deviations and the testing results across seven different tasks, along with their average, are shown. The data indicate that OCC-CLIP exhibits superior robustness compared to the baseline in most cases, demonstrating enhanced resilience to various forms of image manipulation. This underscores the effectiveness of OCC-CLIP in safeguarding against image-based copyright infringements.

\section{Sensitivity to Choice of Prompts}
We explore the impact of prompt selection on the performance of OCC-CLIP. As indicated in Tables~\ref{table: Different Prompts Sup} and \ref{table: Different Prompts Sup2}, the choice of prompts influences performance across seven different testing tasks. However, OCC-CLIP consistently outperforms the baseline regardless of the prompt pair chosen.

\section{Comparison to Standard Data Augmentation}
In a comprehensive evaluation of ADA, we compare its effectiveness with common data augmentation methods: Gaussian Blur, Gaussian Noise, Grayscale, Rotation, Flip, and combinations of these. According to Tables~\ref{table: Differnt Data Augmentation sup} and \ref{table: Differnt Data Augmentation sup2}, traditional data augmentation methods generally perform poorly, often degrading performance across many of the seven testing tasks. Only ADA shows improvements, suggesting that standard data augmentation methods may not be suitable for few-shot one-class classification scenarios.

\section{Source Model Attribution with Commercial Generation API}
To assess OCC-CLIP's real-world effectiveness, we used the latest commercial digital image-generating model, DALL·E-3~\cite{betker2023improving}, to generate images. As detailed in Section~\ref{Prompts for DALL·E-3}, 103 prompts were randomly selected from the annotations of COCO's validation set~\cite{lin2014microsoft}. In addition to the images generated from these prompts, additional sample images are presented in Figure~\ref{fig: visualisation}. During the training phase, the target images are those generated by DALL·E-3, while the non-target images are from COCO. In the testing phase, the non-target images are from one of the other generated datasets. The performance of detection across various shot numbers (10, 20, 30, 40, 50) is compared. According to Tables~\ref{table: Real Application Sup} and \ref{table: Real Application Sup2}, our method consistently outperforms the baseline across all eight testing tasks and performs well even with as few as 10 shots. Due to limitations in the number of images generated by DALL·E-3, we train the model only once for each testing task.

\section{Model Attribution with Multiple Source Models}
Since this involves multi-source origin attribution, the evaluation metric used in this section is \textbf{accuracy} instead of AUC. We evaluate the model under the following conditions: 2 classes (SD, VQ-D), 4 classes (SD, VQ-D, LDM, Glide), and 6 classes (SD, VQ-D, LDM, Glide, GALIP, ProGAN).
Table~\ref{Table: OCC Multi Class} shows the accuracy of verifying multiple sources using the eight trained one-class classifiers. Table~\ref{Table: Direct Multi Class} shows the accuracy of directly training a multi-class classifier.

\section{Stronger Baselines and Harder Datasets}
We also experimented with three representative methods of fine-tuning CLIP using LoRA: replacing the linear layer in the last (LoRAl), middle (LoRAm), or all (LoRAa) residual attention blocks. The other two baseline methods are Wang \textit{et al.}'s ~\cite{wang2023alteration}, which uses inverse engineering and has model access, and Girish \textit{et al.}'s ~\cite{girish2021towards}, which uses a multistep pipeline. Table \ref{table: Additional Baselines} shows the \textbf{accuracy} of these methods. Our approach outperforms these baselines in our setting.

\section{Prompts for DALL·E-3}
\label{Prompts for DALL·E-3}
The following prompts were randomly selected from the annotations of the validation set of the COCO dataset. \\

\noindent\textit{A man holding a motion controlled video game controller  \\
A person sitting at a table filled with mexican food. \\
a work desk with a monitor and keyboard \\
A clock is standing in the middle of the grass in the middle of the afternoon. \\
some people are walking in front of a tall building \\
Three people standing before airport counters below airport signs. \\
A snow mountain being used for winter sports. \\
a man that is cutting up some kind of fruit \\
A bench sitting on a sidewalk near a line of cars. \\
A skier holds his skis as he stands in the mountains. \\
A living room filled with lots of furniture and seats. \\
A train is parked at the station loading passengers. \\
A man and woman with two Clydesdale horses.  \\
A bathroom with tub and toilet, tiled in white tiling. \\
A Amtrak train traveling on a railroad tracks. \\
A stuffed teddy bear sitting amongst pillows on a bed \\
A green train traveling down train tracks next to another train. \\
A set of three red double decker buses parked next to each other. \\
A large plate of waffles is next to plates with slices of peaches. \\
A couple of people riding skis down a snow covered slope. \\
A big building perched atop a hill with a sign in the foreground. \\
The man and woman are playing video games in the room. \\
Barbecued meat and vegetables laid out on a counter ready for dinner \\
A herd of elephants in the wild near a river. \\
A hot dog with mustard and a bun next to a ketchup cup. \\
A dog rides on the back of a sheep. \\
The woman is holding up two large hot dogs.  \\
A person is flying through the air near some mountains on his snowboard. \\
A jockey riding and jumping with a horse in an obstacle course. \\
a big bed with a lamp and bedside table and sliding glass leading to a balcony  \\
Several lambs and sheep standing on hay and eating it. \\
an image of a statue of men in a carriage being driven by horses \\
There is a hose hooked up to the fire hydrant by the building. \\
A FOUNTAIN IN A ROUNDABOUT WITH PEOPLE PASSING BY \\
A heard of sheep together in a wheat field. \\
A woman opening a suitcase on the bed \\
A traffic light over a city street with cars. \\
A photo taken from behind a fence of tennis players on the court. \\
A row of parking meters sitting in a park. \\
A woman is talking on her phone while dragging on a cigarette. \\
Subway braking on rails in front of metropolitan city \\
A young man and woman share some pastries. \\
An aerial photo of a very long train station at night. \\
a foggy day that has some lights by a road \\
A dog sitting on the floor between a person legs  \\
This dirt bike rider is smiling and raising his fist in triumph. \\
A desk top computer and a laptop sitting on a computer desk  \\
an orange brick building with a window and a big mascot  \\
A person doing a trick on a skateboard in the road \\
LOTS OF CUPCAKES ARRANGED ON A TABLE WITH NAPKINS \\
A bathroom sink underneath a medicine cabinet next to a window. \\
A red bus driving in front of a double decker bus. \\
The girl is about to kick a soccer ball. \\
A skateboarder jumping through the air and doing a trick. \\
Woman with surfboard getting kisses from dog at waters edge. \\
A skier leaning to the side on a snowy hill. \\
some motorcycles parked and the one in front is a silver three wheeler \\
A skier standing at the stop of a mountain slope. \\
Two planes that are flying in the sky. \\
A man presenting something to another man in a tent. \\
A man in a wet suit walking with another man in a wet suit with equipment. \\
a man flying a kite in a big green field \\
A man holding a dog mug pints the remote. \\
two little birds sitting on the shore by a piece of wood  \\
A giraffe eats next to a zebra among some rocks. \\
A cat thats about to take a bute out of someone's sandwich.  \\
A gray dog wearing an orange bow tie laying on a sofa. \\
A yellow swan boat ride on a body of water under a cloudless sky. \\
A old house with a clock tower in brown and white. \\
a surfer that has fallen off of his surf board \\
Someone is enjoying a small slice of pie.  \\
A coach balances a soccer ball in front of her team \\
A couple of giraffes eating out of a feeding bin in a zoo type facility. \\
A large gray teddy bear sitting next to a candle. \\
an image of a cat lying next to a stuffed animal \\
A woman smiling while she prepares a plate of food.  \\
A group of zebras are standing in a desert. \\
Cat sitting on a window sill in front of a windmill. \\
A bunch of people walking on the street with umbrellas. \\
A skate boarder doing jumps at night on city street. \\
A man with glasses holding a baby in a white outfit. \\
A man in glasses talking on a cellphone. \\
A man with glasses is smiling and clapping. \\
two people sitting on benches with trees in the background \\
A girl riding on the back of a scooter on a cobbled road.  \\
A small computer keyboard, box and matching mouse \\
a couple of people on skis stand in the snow  \\
A man sitting in a chair with an infant laying on his lap and a woman standing over the top and looking down at the infant. \\
Several alcoholic beverages and mixers on a airplane tray. \\
A very big pretty horse pulling a fancy carriage. \\
A young man has just hit a baseball with the bat \\
A woman in a green shirt stands near a table with bowls or oranges on it in a market. \\
A person watching a small plane taking off in a field \\
A bicycle being held by a bar on a train. \\
A baseball game in progress in the open air. \\
A police car parked on the side of the road. \\
A white toilet sitting next to a toilet paper roller. \\
A bright green frog on a bright green plant. \\
A man surfing a wave on his surf board. \\
A small dog reclines on a piece of furniture next to an electronic game controller. \\
A person skiing alone in the snow capped area \\
A large family group at a round table in a restaraunt. \\
A stop sign is erected next to a flower bush.
}

\clearpage

\begin{table*}[t]
    \centering
    \footnotesize
        \caption{Evaluation of OCC-CLIP, along with 14 basic methods. Training samples are sourced from SD (denoted as target) and COCO (denoted as non-target). The test data are assembled from SD as well as a different generative image dataset. The optimal outcomes for individual datasets are emphasized using \textbf{bold} formatting. The metric shown here is \textbf{AUC}.}
        \begin{tabular}{@{}l|cccccccccc@{}}
        \toprule[1pt]
        Methods & VQ-D & LDM & Glide & GALIP \\
        \midrule 
        VGG16~\cite{simonyan2014very} & $0.6458\scriptscriptstyle \pm \scriptstyle2.64e\text{-}2$
            & $0.5438\scriptscriptstyle \pm \scriptstyle2.29e\text{-}2$
            & $0.5652\scriptscriptstyle \pm \scriptstyle2.48e\text{-}2$
            & $0.7017\scriptscriptstyle \pm \scriptstyle4.15e\text{-}2$ \\
        ResNet50~\cite{he2016deep} & $0.6693\scriptscriptstyle \pm \scriptstyle5.27e\text{-}2$
            & $0.5485\scriptscriptstyle \pm \scriptstyle3.07e\text{-}2$
            & $0.6103\scriptscriptstyle \pm \scriptstyle4.03e\text{-}2$
            & $0.7307\scriptscriptstyle \pm \scriptstyle3.94e\text{-}2$ \\
        Inception-v3~\cite{szegedy2016rethinking} & $0.6378\scriptscriptstyle \pm \scriptstyle4.45e\text{-}2$
            & $0.5349\scriptscriptstyle \pm \scriptstyle2.49e\text{-}2$
            & $0.5252\scriptscriptstyle \pm \scriptstyle1.78e\text{-}2$
            & $0.6759\scriptscriptstyle \pm \scriptstyle4.26e\text{-}2$ \\
        DenseNet-121~\cite{huang2017densely} & $0.7264\scriptscriptstyle \pm \scriptstyle4.70e\text{-}2$
            & $0.5337\scriptscriptstyle \pm \scriptstyle3.40e\text{-}2$
            & $0.5730\scriptscriptstyle \pm \scriptstyle5.62e\text{-}2$
            & $0.7478\scriptscriptstyle \pm \scriptstyle4.85e\text{-}2$\\
        \midrule
        ViT-B-16~\cite{dosovitskiy2020image} & $0.7502\scriptscriptstyle \pm \scriptstyle2.70e\text{-}2$
            & $0.5387\scriptscriptstyle \pm \scriptstyle2.06e\text{-}2$
            & $0.6069\scriptscriptstyle \pm \scriptstyle4.46e\text{-}2$
            & $0.7837\scriptscriptstyle \pm \scriptstyle2.05e\text{-}2$ \\
        DeiT-B-16~\cite{touvron2021training} & $0.6825\scriptscriptstyle \pm \scriptstyle3.96e\text{-}2$
            & $0.5366\scriptscriptstyle \pm \scriptstyle2.47e\text{-}2$
            & $0.5698\scriptscriptstyle \pm \scriptstyle2.91e\text{-}2$
            & $0.6592\scriptscriptstyle \pm \scriptstyle2.76e\text{-}2$\\       
        CaiT-S-24~\cite{touvron2021going} & $0.6522\scriptscriptstyle \pm \scriptstyle4.85e\text{-}2$
            & $0.5301\scriptscriptstyle \pm \scriptstyle2.51e\text{-}2$
            & $0.5550\scriptscriptstyle \pm \scriptstyle5.18e\text{-}2$
            & $0.6466\scriptscriptstyle \pm \scriptstyle4.48e\text{-}2$ \\
        Swin-B-4~\cite{liu2021swin} & $0.8634\scriptscriptstyle \pm \scriptstyle3.69e\text{-}2$
            & $0.7180\scriptscriptstyle \pm \scriptstyle5.06e\text{-}2$
            & $0.7473\scriptscriptstyle \pm \scriptstyle3.97e\text{-}2$
            & $0.8478\scriptscriptstyle \pm \scriptstyle4.14e\text{-}2$ \\
        \midrule
        Image\_Patch~\cite{mandelli2022detecting} & $0.6638\scriptscriptstyle \pm \scriptstyle3.69e\text{-}2$
            & $0.5269\scriptscriptstyle \pm \scriptstyle1.84e\text{-}2$
            & $0.5534\scriptscriptstyle \pm \scriptstyle3.15e\text{-}2$
            & $0.6847\scriptscriptstyle \pm \scriptstyle4.41e\text{-}2$ \\
        Feature\_Patch~\cite{mandelli2022detecting} & $0.6999\scriptscriptstyle \pm \scriptstyle2.53e\text{-}2$
            & $0.6116\scriptscriptstyle \pm \scriptstyle3.27e\text{-}2$
            & $0.7385\scriptscriptstyle \pm \scriptstyle7.70e\text{-}2$
            & $0.6285\scriptscriptstyle \pm \scriptstyle5.62e\text{-}2$ \\
        \midrule
        CLIP~\cite{radford2021learning} & $0.7272\scriptscriptstyle \pm \scriptstyle0.0$
            & $0.7243\scriptscriptstyle \pm \scriptstyle0.0$
            & $0.6078\scriptscriptstyle \pm \scriptstyle0.0$
            & $0.7304\scriptscriptstyle \pm \scriptstyle0.0$ \\
        CoOp~\cite{zhou2022learning} & $0.9503\scriptscriptstyle \pm \scriptstyle1.68e\text{-}2$
            & $0.8373\scriptscriptstyle \pm \scriptstyle5.33e\text{-}2$
            & $0.9266\scriptscriptstyle \pm \scriptstyle3.19e\text{-}2$
            & $0.9660\scriptscriptstyle \pm \scriptstyle2.56e\text{-}2$ \\
        \midrule
        OCC-CLIP & $\mathbf{0.9703\scriptscriptstyle \pm \scriptstyle9.06e\text{-}3}$
            & $\mathbf{0.8801\scriptscriptstyle \pm \scriptstyle2.06e\text{-}2}$
            & $\mathbf{0.9519\scriptscriptstyle \pm \scriptstyle1.45e\text{-}2}$
            & $\mathbf{0.9798\scriptscriptstyle \pm \scriptstyle1.18e\text{-}2}$ \\
        \bottomrule[1pt]
        \end{tabular}
    
\label{table: 14 baselines sup}
\end{table*}

\begin{table*}[t]
    \centering
    \footnotesize
        \caption{Evaluation of OCC-CLIP, along with 14 basic methods. Training samples are sourced from SD (denoted as target) and COCO (denoted as non-target). The test data are assembled from SD as well as a different generative image dataset. The optimal outcomes for individual datasets are emphasized using \textbf{bold} formatting. The metric shown here is \textbf{AUC}.}
        \begin{tabular}{@{}l|cccccccccc@{}}
        \toprule[1pt]
        Methods & ProGan & StyleGan2 & GauGan & Overall \\
        \midrule 
        VGG16~\cite{simonyan2014very} 
            & $0.6609\scriptscriptstyle \pm \scriptstyle5.46e\text{-}2$
            & $0.6170\scriptscriptstyle \pm \scriptstyle7.87e\text{-}2$
            & $0.7096\scriptscriptstyle \pm \scriptstyle4.62e\text{-}2$
            & $0.6349\scriptscriptstyle \pm \scriptstyle4.61e\text{-}2$ \\
        ResNet50~\cite{he2016deep} 
            & $0.6646\scriptscriptstyle \pm \scriptstyle4.56e\text{-}2$
            & $0.6458\scriptscriptstyle \pm \scriptstyle9.33e\text{-}2$
            & $0.7475\scriptscriptstyle \pm \scriptstyle3.39e\text{-}2$
            & $0.6595\scriptscriptstyle \pm \scriptstyle5.18e\text{-}2$ \\
        Inception-v3~\cite{szegedy2016rethinking} 
            & $0.6453\scriptscriptstyle \pm \scriptstyle4.87e\text{-}2$
            & $0.5947\scriptscriptstyle \pm \scriptstyle4.88e\text{-}2$
            & $0.6982\scriptscriptstyle \pm \scriptstyle4.31e\text{-}2$
            & $0.6160\scriptscriptstyle \pm \scriptstyle4.03e\text{-}2$ \\
        DenseNet-121~\cite{huang2017densely} 
            & $0.7204\scriptscriptstyle \pm \scriptstyle3.35e\text{-}2$
            & $0.6645\scriptscriptstyle \pm \scriptstyle6.78e\text{-}2$
            & $0.8091\scriptscriptstyle \pm \scriptstyle2.94e\text{-}2$
            & $0.6821\scriptscriptstyle \pm \scriptstyle4.70e\text{-}2$ \\
        \midrule
        ViT-B-16~\cite{dosovitskiy2020image} 
            & $0.6062\scriptscriptstyle \pm \scriptstyle3.71e\text{-}2$
            & $0.6593\scriptscriptstyle \pm \scriptstyle6.11e\text{-}2$
            & $0.7259\scriptscriptstyle \pm \scriptstyle2.77e\text{-}2$
            & $0.6673\scriptscriptstyle \pm \scriptstyle3.67e\text{-}2$ \\
        DeiT-B-16~\cite{touvron2021training} 
            & $0.5721\scriptscriptstyle \pm \scriptstyle2.77e\text{-}2$
            & $0.5898\scriptscriptstyle \pm \scriptstyle4.20e\text{-}2$
            & $0.6507\scriptscriptstyle \pm \scriptstyle2.72e\text{-}2$
            & $0.6087\scriptscriptstyle \pm \scriptstyle3.17e\text{-}2$ \\       
        CaiT-S-24~\cite{touvron2021going} 
            & $0.5697\scriptscriptstyle \pm \scriptstyle2.63e\text{-}2$
            & $0.5804\scriptscriptstyle \pm \scriptstyle5.12e\text{-}2$
            & $0.6645\scriptscriptstyle \pm \scriptstyle3.42e\text{-}2$
            & $0.5998\scriptscriptstyle \pm \scriptstyle4.17e\text{-}2$ \\
        Swin-B-4~\cite{liu2021swin} 
            & $0.5879\scriptscriptstyle \pm \scriptstyle3.68e\text{-}2$
            & $0.7054\scriptscriptstyle \pm \scriptstyle6.30e\text{-}2$
            & $0.7822\scriptscriptstyle \pm \scriptstyle3.91e\text{-}2$
            & $0.7503\scriptscriptstyle \pm \scriptstyle4.48e\text{-}2$ \\
        \midrule
        Image\_Patch~\cite{mandelli2022detecting} 
            & $0.6624\scriptscriptstyle \pm \scriptstyle4.82e\text{-}2$
            & $0.6706\scriptscriptstyle \pm \scriptstyle9.62e\text{-}2$
            & $0.7674\scriptscriptstyle \pm \scriptstyle4.09e\text{-}2$
            & $0.6470\scriptscriptstyle \pm \scriptstyle5.05e\text{-}2$ \\
        Feature\_Patch~\cite{mandelli2022detecting} 
            & $0.7136\scriptscriptstyle \pm \scriptstyle2.77e\text{-}2$
            & $0.8546\scriptscriptstyle \pm \scriptstyle2.65e\text{-}2$
            & $0.8152\scriptscriptstyle \pm \scriptstyle3.14e\text{-}2$
            & $0.7231\scriptscriptstyle \pm \scriptstyle4.35e\text{-}2$ \\
        \midrule
        CLIP~\cite{radford2021learning} 
            & $0.5229\scriptscriptstyle \pm \scriptstyle0.0$
            & $0.5393\scriptscriptstyle \pm \scriptstyle0.0$
            & $0.6463\scriptscriptstyle \pm \scriptstyle0.0$
            & $0.6426\scriptscriptstyle \pm \scriptstyle0.0$ \\
        CoOp~\cite{zhou2022learning} 
            & $0.8861\scriptscriptstyle \pm \scriptstyle4.46e\text{-}2$
            & $0.9533\scriptscriptstyle \pm \scriptstyle2.30e\text{-}2$
            & $0.9643\scriptscriptstyle \pm \scriptstyle2.91e\text{-}2$
            & $0.9263\scriptscriptstyle \pm \scriptstyle3.41e\text{-}2$ \\
        \midrule
        OCC-CLIP 
            & $\mathbf{0.9452\scriptscriptstyle \pm \scriptstyle3.14e\text{-}2}$
            & $\mathbf{0.9651\scriptscriptstyle \pm \scriptstyle1.54e\text{-}2}$
            & $\mathbf{0.9910\scriptscriptstyle \pm \scriptstyle7.32e\text{-}3}$
            & $\mathbf{0.9548\scriptscriptstyle \pm \scriptstyle1.75e\text{-}2}$ \\
        \bottomrule[1pt]
        \end{tabular}
    
\label{table: 14 baselines sup2}
\end{table*}

\begin{table*}[t]
    \centering
    \footnotesize
        \caption{Evaluation of OCC-CLIP, along with 14 basic methods. Training samples are sourced from SD (denoted as target) and COCO (denoted as non-target). The test data are assembled from SD as well as a different generative image dataset. The optimal outcomes for individual datasets are emphasized using \textbf{bold} formatting. The metric shown here is \textbf{Accuracy}.}
        \begin{tabular}{@{}l|cccccccccc@{}}
        \toprule[1pt]
        Methods & VQ-D & LDM & Glide & GALIP \\
        \midrule 
        VGG16~\cite{simonyan2014very} & $0.6024\scriptscriptstyle \pm \scriptstyle2.78e\text{-}2$
 & $0.5260\scriptscriptstyle \pm \scriptstyle1.53e\text{-}2$
 & $0.5396\scriptscriptstyle \pm \scriptstyle1.63e\text{-}2$
 & $0.6501\scriptscriptstyle \pm \scriptstyle3.44e\text{-}2$ \\
        ResNet50~\cite{he2016deep} & $0.6235\scriptscriptstyle \pm \scriptstyle4.71e\text{-}2$
 & $0.5328\scriptscriptstyle \pm \scriptstyle2.49e\text{-}2$
 & $0.5764\scriptscriptstyle \pm \scriptstyle3.03e\text{-}2$
 & $0.6729\scriptscriptstyle \pm \scriptstyle3.43e\text{-}2$ \\
        Inception-v3~\cite{szegedy2016rethinking} & $0.5999\scriptscriptstyle \pm \scriptstyle3.32e\text{-}2$
 & $0.5193\scriptscriptstyle \pm \scriptstyle1.29e\text{-}2$
 & $0.5219\scriptscriptstyle \pm \scriptstyle1.31e\text{-}2$
 & $0.6297\scriptscriptstyle \pm \scriptstyle2.99e\text{-}2$ \\
        DenseNet-121~\cite{huang2017densely} & $0.6548\scriptscriptstyle \pm \scriptstyle3.33e\text{-}2$
 & $0.5198\scriptscriptstyle \pm \scriptstyle2.07e\text{-}2$
 & $0.5423\scriptscriptstyle \pm \scriptstyle4.13e\text{-}2$
 & $0.6731\scriptscriptstyle \pm \scriptstyle4.53e\text{-}2$\\
        \midrule
        ViT-B-16~\cite{dosovitskiy2020image} & $0.6849\scriptscriptstyle \pm \scriptstyle2.16e\text{-}2$
 & $0.5188\scriptscriptstyle \pm \scriptstyle1.45e\text{-}2$
 & $0.5633\scriptscriptstyle \pm \scriptstyle2.55e\text{-}2$
 & $0.7113\scriptscriptstyle \pm \scriptstyle1.58e\text{-}2$ \\
        DeiT-B-16~\cite{touvron2021training} & $0.6285\scriptscriptstyle \pm \scriptstyle3.15e\text{-}2$
 & $0.5245\scriptscriptstyle \pm \scriptstyle1.41e\text{-}2$
 & $0.5401\scriptscriptstyle \pm \scriptstyle2.25e\text{-}2$
 & $0.6119\scriptscriptstyle \pm \scriptstyle2.07e\text{-}2$\\       
        CaiT-S-24~\cite{touvron2021going} & $0.6070\scriptscriptstyle \pm \scriptstyle3.82e\text{-}2$
 & $0.5216\scriptscriptstyle \pm \scriptstyle1.54e\text{-}2$
 & $0.5394\scriptscriptstyle \pm \scriptstyle3.41e\text{-}2$
 & $0.6027\scriptscriptstyle \pm \scriptstyle3.71e\text{-}2$ \\
        Swin-B-4~\cite{liu2021swin} & $0.7796\scriptscriptstyle \pm \scriptstyle3.39e\text{-}2$
 & $0.6498\scriptscriptstyle \pm \scriptstyle3.57e\text{-}2$
 & $0.6652\scriptscriptstyle \pm \scriptstyle2.93e\text{-}2$
 & $0.7623\scriptscriptstyle \pm \scriptstyle3.86e\text{-}2$ \\
        \midrule
        Image\_Patch~\cite{mandelli2022detecting} & $0.6143\scriptscriptstyle \pm \scriptstyle2.78e\text{-}2$
 & $0.5136\scriptscriptstyle \pm \scriptstyle1.13e\text{-}2$
 & $0.5348\scriptscriptstyle \pm \scriptstyle2.06e\text{-}2$
 & $0.6341\scriptscriptstyle \pm \scriptstyle4.01e\text{-}2$ \\
        Feature\_Patch~\cite{mandelli2022detecting} & $0.6434\scriptscriptstyle \pm \scriptstyle1.97e\text{-}2$
 & $0.5767\scriptscriptstyle \pm \scriptstyle2.81e\text{-}2$
 & $0.6719\scriptscriptstyle \pm \scriptstyle5.89e\text{-}2$
 & $0.5908\scriptscriptstyle \pm \scriptstyle3.95e\text{-}2$ \\
        \midrule
        CLIP~\cite{radford2021learning} & $0.5800\scriptscriptstyle \pm \scriptstyle0.0$
            & $0.5695\scriptscriptstyle \pm \scriptstyle0.0$
            & $0.5475\scriptscriptstyle \pm \scriptstyle0.0$
            & $0.5755\scriptscriptstyle \pm \scriptstyle0.0$ \\
        CoOp~\cite{zhou2022learning} & $0.8457\scriptscriptstyle \pm \scriptstyle5.91e\text{-}2$
 & $0.6745\scriptscriptstyle \pm \scriptstyle7.54e\text{-}2$
 & $0.8041\scriptscriptstyle \pm \scriptstyle7.07e\text{-}2$
 & $0.8722\scriptscriptstyle \pm \scriptstyle5.88e\text{-}2$ \\
        \midrule
        OCC-CLIP & $\mathbf{0.9118\scriptscriptstyle \pm \scriptstyle1.73e\text{-}2}$
            & $\mathbf{0.7654\scriptscriptstyle \pm \scriptstyle3.28e\text{-}2}$
            & $\mathbf{0.8783\scriptscriptstyle \pm \scriptstyle2.20e\text{-}2}$
            & $\mathbf{0.9227\scriptscriptstyle \pm \scriptstyle1.91e\text{-}2}$ \\
        \bottomrule[1pt]
        \end{tabular}
    
\label{table: 14 baselines sup accuracy}
\end{table*}

\begin{table*}[t]
    \centering
    \footnotesize
        \caption{Evaluation of OCC-CLIP, along with 14 basic methods. Training samples are sourced from SD (denoted as target) and COCO (denoted as non-target). The test data are assembled from SD as well as a different generative image dataset. The optimal outcomes for individual datasets are emphasized using \textbf{bold} formatting. The metric shown here is \textbf{Accuracy}.}
        \begin{tabular}{@{}l|cccccccccc@{}}
        \toprule[1pt]
        Methods & ProGan & StyleGan2 & GauGan & Overall \\
        \midrule 
        VGG16~\cite{simonyan2014very} 
            & $0.6218\scriptscriptstyle \pm \scriptstyle4.14e\text{-}2$
 & $0.5939\scriptscriptstyle \pm \scriptstyle5.57e\text{-}2$
 & $0.6597\scriptscriptstyle \pm \scriptstyle3.82e\text{-}2$

& $0.5991\scriptscriptstyle \pm \scriptstyle3.53e\text{-}2$ \\
        ResNet50~\cite{he2016deep} 
            & $0.6157\scriptscriptstyle \pm \scriptstyle3.92e\text{-}2$
 & $0.5910\scriptscriptstyle \pm \scriptstyle6.70e\text{-}2$
 & $0.6858\scriptscriptstyle \pm \scriptstyle2.48e\text{-}2$

& $0.6140\scriptscriptstyle \pm \scriptstyle4.07e\text{-}2$ \\
        Inception-v3~\cite{szegedy2016rethinking} 
            & $0.6045\scriptscriptstyle \pm \scriptstyle3.65e\text{-}2$
 & $0.5711\scriptscriptstyle \pm \scriptstyle3.59e\text{-}2$
 & $0.6455\scriptscriptstyle \pm \scriptstyle2.96e\text{-}2$

& $0.5846\scriptscriptstyle \pm \scriptstyle2.88e\text{-}2$ \\
        DenseNet-121~\cite{huang2017densely} 
            & $0.6556\scriptscriptstyle \pm \scriptstyle3.34e\text{-}2$
 & $0.5967\scriptscriptstyle \pm \scriptstyle5.24e\text{-}2$
 & $0.7276\scriptscriptstyle \pm \scriptstyle2.66e\text{-}2$

& $0.6243\scriptscriptstyle \pm \scriptstyle3.76e\text{-}2$ \\
        \midrule
        ViT-B-16~\cite{dosovitskiy2020image} 
            & $0.5649\scriptscriptstyle \pm \scriptstyle2.79e\text{-}2$
 & $0.6148\scriptscriptstyle \pm \scriptstyle4.20e\text{-}2$
 & $0.6606\scriptscriptstyle \pm \scriptstyle2.62e\text{-}2$

& $0.6169\scriptscriptstyle \pm \scriptstyle2.62e\text{-}2$ \\
        DeiT-B-16~\cite{touvron2021training} 
            & $0.5451\scriptscriptstyle \pm \scriptstyle2.04e\text{-}2$
 & $0.5601\scriptscriptstyle \pm \scriptstyle3.47e\text{-}2$
 & $0.6051\scriptscriptstyle \pm \scriptstyle2.08e\text{-}2$

& $0.5736\scriptscriptstyle \pm \scriptstyle2.44e\text{-}2$ \\       
        CaiT-S-24~\cite{touvron2021going} 
            & $0.5437\scriptscriptstyle \pm \scriptstyle2.33e\text{-}2$
 & $0.5468\scriptscriptstyle \pm \scriptstyle3.63e\text{-}2$
 & $0.6166\scriptscriptstyle \pm \scriptstyle2.68e\text{-}2$

& $0.5683\scriptscriptstyle \pm \scriptstyle3.12e\text{-}2$ \\
        Swin-B-4~\cite{liu2021swin} 
             & $0.5378\scriptscriptstyle \pm \scriptstyle1.82e\text{-}2$
 & $0.6057\scriptscriptstyle \pm \scriptstyle3.40e\text{-}2$
 & $0.6929\scriptscriptstyle \pm \scriptstyle3.75e\text{-}2$
 & $0.6705\scriptscriptstyle \pm \scriptstyle3.31e\text{-}2$ \\
        \midrule
        Image\_Patch~\cite{mandelli2022detecting} 
            & $0.6128\scriptscriptstyle \pm \scriptstyle4.33e\text{-}2$
 & $0.6230\scriptscriptstyle \pm \scriptstyle7.13e\text{-}2$
 & $0.7010\scriptscriptstyle \pm \scriptstyle3.19e\text{-}2$

& $0.6048\scriptscriptstyle \pm \scriptstyle3.95e\text{-}2$ \\
        Feature\_Patch~\cite{mandelli2022detecting} 
            & $0.6582\scriptscriptstyle \pm \scriptstyle2.29e\text{-}2$
 & $0.7678\scriptscriptstyle \pm \scriptstyle1.65e\text{-}2$
 & $0.7333\scriptscriptstyle \pm \scriptstyle2.30e\text{-}2$

& $0.6632\scriptscriptstyle \pm \scriptstyle3.28e\text{-}2$ \\
        \midrule
        CLIP~\cite{radford2021learning} 
            & $0.5415\scriptscriptstyle \pm \scriptstyle0.0$
            & $0.5860\scriptscriptstyle \pm \scriptstyle0.0$
            & $0.5170\scriptscriptstyle \pm \scriptstyle0.0$
            & $0.5645\scriptscriptstyle \pm \scriptstyle0.0$ \\
        CoOp~\cite{zhou2022learning} 
            & $0.7053\scriptscriptstyle \pm \scriptstyle8.57e\text{-}2$
 & $0.8540\scriptscriptstyle \pm \scriptstyle5.87e\text{-}2$
 & $0.8854\scriptscriptstyle \pm \scriptstyle3.84e\text{-}2$

& $0.8059\scriptscriptstyle \pm \scriptstyle6.54e\text{-}2$ \\
        \midrule
        OCC-CLIP 
            & $\mathbf{0.8635\scriptscriptstyle \pm \scriptstyle5.26e\text{-}2}$
            & $\mathbf{0.8939\scriptscriptstyle \pm \scriptstyle2.69e\text{-}2}$
            & $\mathbf{0.9409\scriptscriptstyle \pm \scriptstyle1.79e\text{-}2}$
            & $\mathbf{0.8824\scriptscriptstyle \pm \scriptstyle2.94e\text{-}2}$ \\
        \bottomrule[1pt]
        \end{tabular}
    
\label{table: 14 baselines sup2 accuracy}
\end{table*}

\begin{table*}[ht]
    \centering
    \scriptsize
    
        \caption{Evaluation sensitivity of OCC-CLIP to Source Models and choice of Epsilon. The leftmost column represents target datasets. In training phase, the non-target images are from COCO. In testing phase, the non-target images are from a different generative image dataset shown in the first row. Different epsilon is chosen: 0.0, 0.03125, 0.0625, 0.1, 0.2, andm 0.5. The optimal outcomes for individual datasets are emphasized using \textbf{bold} formatting.}
        \begin{tabular}{@{}l|l|ccccccccccc@{}}
        \toprule[1pt]
        Target & Epsilon & SD\textsubscript{LAION-5B} & VQ-D\textsubscript{coco} & LDM\textsubscript{LAION-400M} & Glide\textsubscript{filtered-cc} \\
        \midrule
        \multirow{6}{*}{SD\textsubscript{LAION-5B}} 
        & 0.0
        & --
        & $0.9503\scriptscriptstyle \pm \scriptstyle1.68e\text{-}2$
        & $0.8373\scriptscriptstyle \pm \scriptstyle5.33e\text{-}2$
        & $0.9266\scriptscriptstyle \pm \scriptstyle3.19e\text{-}2$ \\
        & 0.03125
        & --
        & $0.9627\scriptscriptstyle \pm \scriptstyle1.74e\text{-}2$
        & $0.8782\scriptscriptstyle \pm \scriptstyle4.48e\text{-}2$
        & $0.9248\scriptscriptstyle \pm \scriptstyle3.58e\text{-}2$ \\
        & 0.0625
        & --
        & $0.9644\scriptscriptstyle \pm \scriptstyle1.73e\text{-}2$
        & $0.8720\scriptscriptstyle \pm \scriptstyle3.74e\text{-}2$
        & $0.9332\scriptscriptstyle \pm \scriptstyle3.10e\text{-}2$ \\
        & 0.1
        & --
        & $\mathbf{0.9703\scriptscriptstyle \pm \scriptstyle9.06e\text{-}3}$
        & $\mathbf{0.8801\scriptscriptstyle \pm \scriptstyle2.06e\text{-}2}$
        & $\mathbf{0.9519\scriptscriptstyle \pm \scriptstyle1.45e\text{-}2}$ \\       
        & 0.2
        & --
        & $0.9595\scriptscriptstyle \pm \scriptstyle1.49e\text{-}2$
        & $0.8456\scriptscriptstyle \pm \scriptstyle4.05e\text{-}2$
        & $0.9388\scriptscriptstyle \pm \scriptstyle2.54e\text{-}2$ \\
        & 0.5
        & --
        & $0.9601\scriptscriptstyle \pm \scriptstyle1.76e\text{-}2$
        & $0.8475\scriptscriptstyle \pm \scriptstyle4.87e\text{-}2$
        & $0.9384\scriptscriptstyle \pm \scriptstyle3.18e\text{-}2$ \\
        \midrule
        \multirow{6}{*}{VQ-D\textsubscript{coco}}
        & 0.0
        & $0.9988\scriptscriptstyle \pm \scriptstyle9.79e\text{-}4$
        & --
        & $\mathbf{0.7337\scriptscriptstyle \pm \scriptstyle4.63e\text{-}2}$
        & $0.7435\scriptscriptstyle \pm \scriptstyle4.39e\text{-}2$ \\
        & 0.03125
        & $0.9982\scriptscriptstyle \pm \scriptstyle5.66e\text{-}4$
        & --
        & $0.7077\scriptscriptstyle \pm \scriptstyle4.06e\text{-}2$
        & $0.6988\scriptscriptstyle \pm \scriptstyle4.61e\text{-}2$ \\
        & 0.0625
        & $0.9986\scriptscriptstyle \pm \scriptstyle1.08e\text{-}3$
        & --
        & $0.6783\scriptscriptstyle \pm \scriptstyle5.50e\text{-}2$
        & $0.7112\scriptscriptstyle \pm \scriptstyle5.12e\text{-}2$ \\
        & 0.1
        & $0.9992\scriptscriptstyle \pm \scriptstyle6.54e\text{-}4$
        & --
        & $0.6924\scriptscriptstyle \pm \scriptstyle6.56e\text{-}2$
        & $0.7264\scriptscriptstyle \pm \scriptstyle5.62e\text{-}2$ \\
        & 0.2
        & $0.9992\scriptscriptstyle \pm \scriptstyle3.85e\text{-}4$
        & --
        & $0.6998\scriptscriptstyle \pm \scriptstyle3.20e\text{-}2$
        & $\mathbf{0.7478\scriptscriptstyle \pm \scriptstyle4.84e\text{-}2}$ \\
        & 0.5
        & $\mathbf{0.9994\scriptscriptstyle \pm \scriptstyle3.74e\text{-}4}$
        & --
        & $0.6887\scriptscriptstyle \pm \scriptstyle5.88e\text{-}2$
        & $0.7468\scriptscriptstyle \pm \scriptstyle4.32e\text{-}2$ \\
        \midrule
        \multirow{6}{*}{LDM\textsubscript{LAION-400M}}
        & 0.0
        & $0.9925\scriptscriptstyle \pm \scriptstyle7.39e\text{-}3$
        & $0.6793\scriptscriptstyle \pm \scriptstyle6.63e\text{-}2$
        & --
        & $0.6468\scriptscriptstyle \pm \scriptstyle6.15e\text{-}2$ \\
        & 0.03125
        & $0.9941\scriptscriptstyle \pm \scriptstyle4.49e\text{-}3$
        & $0.7295\scriptscriptstyle \pm \scriptstyle3.77e\text{-}2$
        & --
        & $0.6502\scriptscriptstyle \pm \scriptstyle6.56e\text{-}2$ \\
        & 0.0625
        & $0.9945\scriptscriptstyle \pm \scriptstyle5.66e\text{-}3$
        & $0.7473\scriptscriptstyle \pm \scriptstyle4.92e\text{-}2$
        & --
        & $0.6761\scriptscriptstyle \pm \scriptstyle5.30e\text{-}2$ \\
        & 0.1
        & $0.9957\scriptscriptstyle \pm \scriptstyle3.97e\text{-}3$
        & $\mathbf{0.7507\scriptscriptstyle \pm \scriptstyle5.51e\text{-}2}$
        & --
        & $0.6847\scriptscriptstyle \pm \scriptstyle5.32e\text{-}2$ \\
        & 0.2
        & $\mathbf{0.9958\scriptscriptstyle \pm \scriptstyle2.73e\text{-}3}$
        & $0.7461\scriptscriptstyle \pm \scriptstyle5.98e\text{-}2$
        & --
        & $0.6980\scriptscriptstyle \pm \scriptstyle6.25e\text{-}2$ \\
        & 0.5
        & $0.9951\scriptscriptstyle \pm \scriptstyle3.13e\text{-}3$
        & $0.7404\scriptscriptstyle \pm \scriptstyle5.12e\text{-}2$
        & --
        & $\mathbf{0.7142\scriptscriptstyle \pm \scriptstyle3.77e\text{-}2}$ \\
        \midrule
        \multirow{6}{*}{Glide\textsubscript{filtered-cc}}
        & 0.0
        & $0.9985\scriptscriptstyle \pm \scriptstyle1.86e\text{-}3$
        & $0.8573\scriptscriptstyle \pm \scriptstyle4.27e\text{-}2$
        & $0.8314\scriptscriptstyle \pm \scriptstyle3.08e\text{-}2$
        & -- \\
        & 0.03125
        & $0.9997\scriptscriptstyle \pm \scriptstyle3.16e\text{-}4$
        & $0.8970\scriptscriptstyle \pm \scriptstyle1.53e\text{-}2$
        & $0.8553\scriptscriptstyle \pm \scriptstyle2.96e\text{-}2$
        & -- \\
        & 0.0625
        & $0.9998\scriptscriptstyle \pm \scriptstyle1.90e\text{-}4$
        & $0.8919\scriptscriptstyle \pm \scriptstyle1.76e\text{-}2$
        & $0.8559\scriptscriptstyle \pm \scriptstyle2.49e\text{-}2$
        & -- \\
        & 0.1
        & $0.9998\scriptscriptstyle \pm \scriptstyle2.47e\text{-}4$
        & $0.8958\scriptscriptstyle \pm \scriptstyle2.18e\text{-}2$
        & $0.8585\scriptscriptstyle \pm \scriptstyle2.50e\text{-}2$
        & -- \\
        & 0.2
        & $0.9999\scriptscriptstyle \pm \scriptstyle1.18e\text{-}4$
        & $\mathbf{0.9073\scriptscriptstyle \pm \scriptstyle1.04e\text{-}2}$
        & $\mathbf{0.8731\scriptscriptstyle \pm \scriptstyle2.42e\text{-}2}$
        & -- \\
        & 0.5
        & $\mathbf{0.9999\scriptscriptstyle \pm \scriptstyle1.08e\text{-}4}$
        & $0.8898\scriptscriptstyle \pm \scriptstyle2.37e\text{-}2$
        & $0.8724\scriptscriptstyle \pm \scriptstyle3.14e\text{-}2$
        & -- \\
        \midrule
        \multirow{6}{*}{GALIP\textsubscript{coco}}
        & 0.0
        & $0.9999\scriptscriptstyle \pm \scriptstyle2.29e\text{-}4$
        & $0.9036\scriptscriptstyle \pm \scriptstyle2.75e\text{-}2$
        & $0.8441\scriptscriptstyle \pm \scriptstyle4.62e\text{-}2$
        & $0.7802\scriptscriptstyle \pm \scriptstyle4.74e\text{-}2$ \\
        & 0.03125
        & $0.9996\scriptscriptstyle \pm \scriptstyle1.85e\text{-}4$
        & $0.8970\scriptscriptstyle \pm \scriptstyle4.26e\text{-}2$
        & $0.8256\scriptscriptstyle \pm \scriptstyle3.54e\text{-}2$
        & $0.7044\scriptscriptstyle \pm \scriptstyle3.36e\text{-}2$ \\
        & 0.0625
        & $0.9998\scriptscriptstyle \pm \scriptstyle1.83e\text{-}4$
        & $0.9205\scriptscriptstyle \pm \scriptstyle2.30e\text{-}2$
        & $0.8165\scriptscriptstyle \pm \scriptstyle3.54e\text{-}2$
        & $0.7545\scriptscriptstyle \pm \scriptstyle4.19e\text{-}2$ \\
        & 0.1
        & $0.9999\scriptscriptstyle \pm \scriptstyle8.94e\text{-}5$
        & $0.9345\scriptscriptstyle \pm \scriptstyle1.99e\text{-}2$
        & $0.8626\scriptscriptstyle \pm \scriptstyle1.99e\text{-}2$
        & $0.7779\scriptscriptstyle \pm \scriptstyle2.88e\text{-}2$ \\
        & 0.2
        & $0.9999\scriptscriptstyle \pm \scriptstyle6.40e\text{-}5$
        & $\mathbf{0.9380\scriptscriptstyle \pm \scriptstyle1.81e\text{-}2}$
        & $0.8776\scriptscriptstyle \pm \scriptstyle2.96e\text{-}2$
        & $0.8238\scriptscriptstyle \pm \scriptstyle2.09e\text{-}2$ \\
        & 0.5
        & $\mathbf{1.0000\scriptscriptstyle \pm \scriptstyle4.00e\text{-}5}$
        & $0.9332\scriptscriptstyle \pm \scriptstyle1.57e\text{-}2$
        & $\mathbf{0.8907\scriptscriptstyle \pm \scriptstyle3.24e\text{-}2}$
        & $\mathbf{0.8361\scriptscriptstyle \pm \scriptstyle1.73e\text{-}2}$ \\
        \midrule
        \multirow{6}{*}{ProGan\textsubscript{lsun}}
        & 0.0
        & $0.9972\scriptscriptstyle \pm \scriptstyle1.91e\text{-}3$
        & $0.9475\scriptscriptstyle \pm \scriptstyle2.01e\text{-}2$
        & $0.9544\scriptscriptstyle \pm \scriptstyle1.86e\text{-}2$
        & $0.9453\scriptscriptstyle \pm \scriptstyle2.00e\text{-}2$ \\
        & 0.03125
        & $0.9904\scriptscriptstyle \pm \scriptstyle6.10e\text{-}3$
        & $0.9429\scriptscriptstyle \pm \scriptstyle2.49e\text{-}2$
        & $0.9646\scriptscriptstyle \pm \scriptstyle2.12e\text{-}2$
        & $0.8852\scriptscriptstyle \pm \scriptstyle5.46e\text{-}2$ \\
        & 0.0625
        & $0.9934\scriptscriptstyle \pm \scriptstyle4.67e\text{-}3$
        & $0.9313\scriptscriptstyle \pm \scriptstyle3.26e\text{-}2$
        & $0.9517\scriptscriptstyle \pm \scriptstyle2.18e\text{-}2$
        & $0.8868\scriptscriptstyle \pm \scriptstyle4.72e\text{-}2$ \\
        & 0.1
        & $0.9961\scriptscriptstyle \pm \scriptstyle3.27e\text{-}3$
        & $0.9477\scriptscriptstyle \pm \scriptstyle2.46e\text{-}2$
        & $0.9585\scriptscriptstyle \pm \scriptstyle2.58e\text{-}2$
        & $0.9320\scriptscriptstyle \pm \scriptstyle3.59e\text{-}2$ \\
        & 0.2
        & $0.9970\scriptscriptstyle \pm \scriptstyle1.73e\text{-}3$
        & $0.9689\scriptscriptstyle \pm \scriptstyle7.48e\text{-}3$
        & $0.9762\scriptscriptstyle \pm \scriptstyle7.41e\text{-}3$
        & $0.9639\scriptscriptstyle \pm \scriptstyle1.40e\text{-}2$ \\
        & 0.5
        & $\mathbf{0.9980\scriptscriptstyle \pm \scriptstyle1.50e\text{-}3}$
        & $\mathbf{0.9749\scriptscriptstyle \pm \scriptstyle1.13e\text{-}2}$
        & $\mathbf{0.9779\scriptscriptstyle \pm \scriptstyle8.49e\text{-}3}$
        & $\mathbf{0.9750\scriptscriptstyle \pm \scriptstyle1.16e\text{-}2}$ \\
        \midrule
        \multirow{6}{*}{StyleGan2\textsubscript{lsun}}
        & 0.0
        & $0.9992\scriptscriptstyle \pm \scriptstyle8.01e\text{-}4$
        & $0.9856\scriptscriptstyle \pm \scriptstyle8.77e\text{-}3$
        & $0.9850\scriptscriptstyle \pm \scriptstyle1.12e\text{-}2$
        & $0.9584\scriptscriptstyle \pm \scriptstyle3.33e\text{-}2$ \\
        & 0.03125
        & $0.9989\scriptscriptstyle \pm \scriptstyle8.72e\text{-}4$
        & $0.9921\scriptscriptstyle \pm \scriptstyle6.86e\text{-}3$
        & $0.9894\scriptscriptstyle \pm \scriptstyle1.14e\text{-}2$
        & $0.9352\scriptscriptstyle \pm \scriptstyle3.12e\text{-}2$ \\
        & 0.0625
        & $0.9995\scriptscriptstyle \pm \scriptstyle3.75e\text{-}4$
        & $0.9911\scriptscriptstyle \pm \scriptstyle9.81e\text{-}3$
        & $0.9877\scriptscriptstyle \pm \scriptstyle1.32e\text{-}2$
        & $0.9511\scriptscriptstyle \pm \scriptstyle2.88e\text{-}2$ \\
        & 0.1
        & $0.9996\scriptscriptstyle \pm \scriptstyle4.63e\text{-}4$
        & $0.9937\scriptscriptstyle \pm \scriptstyle5.33e\text{-}3$
        & $0.9851\scriptscriptstyle \pm \scriptstyle2.31e\text{-}2$
        & $0.9652\scriptscriptstyle \pm \scriptstyle2.42e\text{-}2$ \\
        & 0.2
        & $0.9997\scriptscriptstyle \pm \scriptstyle2.32e\text{-}4$
        & $\mathbf{0.9941\scriptscriptstyle \pm \scriptstyle2.36e\text{-}3}$
        & $0.9900\scriptscriptstyle \pm \scriptstyle4.43e\text{-}3$
        & $0.9677\scriptscriptstyle \pm \scriptstyle1.23e\text{-}2$ \\
        & 0.5
        & $\mathbf{0.9997\scriptscriptstyle \pm \scriptstyle1.66e\text{-}4}$
        & $0.9943\scriptscriptstyle \pm \scriptstyle2.17e\text{-}3$
        & $\mathbf{0.9917\scriptscriptstyle \pm \scriptstyle3.25e\text{-}3}$
        & $\mathbf{0.9688\scriptscriptstyle \pm \scriptstyle9.89e\text{-}3}$ \\
        \midrule
        \multirow{6}{*}{GauGan\textsubscript{coco}}
        & 0.0
        & $0.9991\scriptscriptstyle \pm \scriptstyle6.97e\text{-}4$
        & $0.9388\scriptscriptstyle \pm \scriptstyle2.52e\text{-}2$
        & $0.9862\scriptscriptstyle \pm \scriptstyle6.91e\text{-}3$
        & $0.9757\scriptscriptstyle \pm \scriptstyle1.02e\text{-}2$ \\
        & 0.03125
        & $0.9960\scriptscriptstyle \pm \scriptstyle2.16e\text{-}3$
        & $0.8420\scriptscriptstyle \pm \scriptstyle6.06e\text{-}2$
        & $0.9616\scriptscriptstyle \pm \scriptstyle1.45e\text{-}2$
        & $0.8890\scriptscriptstyle \pm \scriptstyle4.38e\text{-}2$ \\
        & 0.0625
        & $0.9976\scriptscriptstyle \pm \scriptstyle1.32e\text{-}3$
        & $0.8593\scriptscriptstyle \pm \scriptstyle3.97e\text{-}2$
        & $0.9678\scriptscriptstyle \pm \scriptstyle8.16e\text{-}3$
        & $0.9149\scriptscriptstyle \pm \scriptstyle2.89e\text{-}2$ \\
        & 0.1
        & $0.9982\scriptscriptstyle \pm \scriptstyle1.33e\text{-}3$
        & $0.9132\scriptscriptstyle \pm \scriptstyle2.93e\text{-}2$
        & $0.9745\scriptscriptstyle \pm \scriptstyle9.90e\text{-}3$
        & $0.9593\scriptscriptstyle \pm \scriptstyle2.54e\text{-}2$ \\
        & 0.2
        & $0.9992\scriptscriptstyle \pm \scriptstyle1.34e\text{-}3$
        & $0.9456\scriptscriptstyle \pm \scriptstyle2.44e\text{-}2$
        & $0.9881\scriptscriptstyle \pm \scriptstyle5.31e\text{-}3$
        & $0.9904\scriptscriptstyle \pm \scriptstyle3.47e\text{-}3$ \\
        & 0.5
        & $\mathbf{0.9997\scriptscriptstyle \pm \scriptstyle2.94e\text{-}4}$
        & $\mathbf{0.9518\scriptscriptstyle \pm \scriptstyle2.79e\text{-}2}$
        & $\mathbf{0.9908\scriptscriptstyle \pm \scriptstyle5.50e\text{-}3}$
        & $\mathbf{0.9914\scriptscriptstyle \pm \scriptstyle4.66e\text{-}3}$ \\
        \bottomrule[1pt]
        \end{tabular}
    
\label{table: Different Source Models Sup}
\end{table*}

\begin{table*}[ht]
    \centering
    \tiny
    
        \caption{Evaluation sensitivity of OCC-CLIP to Source Models and choice of Epsilon. The leftmost column represents target datasets. In training phase, the non-target images are from COCO. In testing phase, the non-target images are from a different generative image dataset shown in the first row. Different epsilon is chosen: 0.0, 0.03125, 0.0625, 0.1, 0.2, andm 0.5. The optimal outcomes for individual datasets are emphasized using \textbf{bold} formatting.}
        \begin{tabular}{@{}l|l|ccccc@{}}
        \toprule[1pt]
        Target & Epsilon & GALIP\textsubscript{coco} & ProGan\textsubscript{lsun} & StyleGan2\textsubscript{lsun} & GauGan\textsubscript{coco} & Overall \\
        \midrule
        \multirow{6}{*}{SD\textsubscript{LAION-5B}} 
        & 0.0
        & $0.9660\scriptscriptstyle \pm \scriptstyle2.56e\text{-}2$
        & $0.8861\scriptscriptstyle \pm \scriptstyle4.46e\text{-}2$
        & $0.9533\scriptscriptstyle \pm \scriptstyle2.30e\text{-}2$
        & $0.9643\scriptscriptstyle \pm \scriptstyle2.91e\text{-}2$
        & $0.9263\scriptscriptstyle \pm \scriptstyle3.41e\text{-}2$ \\
        & 0.03125
        & $0.9764\scriptscriptstyle \pm \scriptstyle1.83e\text{-}2$
        & $\mathbf{0.9564\scriptscriptstyle \pm \scriptstyle1.84e\text{-}2}$
        & $\mathbf{0.9695\scriptscriptstyle \pm \scriptstyle2.23e\text{-}2}$
        & $\mathbf{0.9966\scriptscriptstyle \pm \scriptstyle2.79e\text{-}3}$
        & $0.9521\scriptscriptstyle \pm \scriptstyle2.61e\text{-}2$ \\
        & 0.0625
        & $0.9762\scriptscriptstyle \pm \scriptstyle1.77e\text{-}2$
        & $0.9470\scriptscriptstyle \pm \scriptstyle3.82e\text{-}2$
        & $0.9684\scriptscriptstyle \pm \scriptstyle1.67e\text{-}2$
        & $0.9939\scriptscriptstyle \pm \scriptstyle7.33e\text{-}3$
        & $0.9507\scriptscriptstyle \pm \scriptstyle2.61e\text{-}2$ \\
        & 0.1
        & $\mathbf{0.9798\scriptscriptstyle \pm \scriptstyle1.18e\text{-}2}$
        & $0.9452\scriptscriptstyle \pm \scriptstyle3.14e\text{-}2$
        & $0.9651\scriptscriptstyle \pm \scriptstyle1.54e\text{-}2$
        & $0.9910\scriptscriptstyle \pm \scriptstyle7.32e\text{-}3$
        & $\mathbf{0.9548\scriptscriptstyle \pm \scriptstyle1.75e\text{-}2}$ \\       
        & 0.2
        & $0.9722\scriptscriptstyle \pm \scriptstyle1.38e\text{-}2$
        & $0.9196\scriptscriptstyle \pm \scriptstyle2.95e\text{-}2$
        & $0.9578\scriptscriptstyle \pm \scriptstyle2.46e\text{-}2$
        & $0.9821\scriptscriptstyle \pm \scriptstyle1.04e\text{-}2$
        & $0.9394\scriptscriptstyle \pm \scriptstyle2.47e\text{-}2$ \\
        & 0.5
        & $0.9670\scriptscriptstyle \pm \scriptstyle2.01e\text{-}2$
        & $0.9002\scriptscriptstyle \pm \scriptstyle3.15e\text{-}2$
        & $0.9627\scriptscriptstyle \pm \scriptstyle1.99e\text{-}2$
        & $0.9816\scriptscriptstyle \pm \scriptstyle1.03e\text{-}2$
        & $0.9368\scriptscriptstyle \pm \scriptstyle2.83e\text{-}2$ \\
        \midrule
        \multirow{6}{*}{VQ-D\textsubscript{coco}}
        & 0.0
        & $\mathbf{0.7558\scriptscriptstyle \pm \scriptstyle3.82e\text{-}2}$
        & $0.9290\scriptscriptstyle \pm \scriptstyle2.80e\text{-}2$
        & $0.9924\scriptscriptstyle \pm \scriptstyle3.87e\text{-}3$
        & $0.9352\scriptscriptstyle \pm \scriptstyle1.95e\text{-}2$
        & $0.8698\scriptscriptstyle \pm \scriptstyle3.09e\text{-}2$ \\
        & 0.03125
        & $0.7303\scriptscriptstyle \pm \scriptstyle5.59e\text{-}2$
        & $0.9920\scriptscriptstyle \pm \scriptstyle4.31e\text{-}3$
        & $\mathbf{0.9971\scriptscriptstyle \pm \scriptstyle2.20e\text{-}3}$
        & $0.9946\scriptscriptstyle \pm \scriptstyle2.51e\text{-}3$
        & $0.8741\scriptscriptstyle \pm \scriptstyle3.15e\text{-}2$ \\
        & 0.0625
        & $0.7245\scriptscriptstyle \pm \scriptstyle6.14e\text{-}2$
        & $0.9923\scriptscriptstyle \pm \scriptstyle5.29e\text{-}3$
        & $0.9938\scriptscriptstyle \pm \scriptstyle5.97e\text{-}3$
        & $\mathbf{0.9949\scriptscriptstyle \pm \scriptstyle3.19e\text{-}3}$
        & $0.8705\scriptscriptstyle \pm \scriptstyle3.68e\text{-}2$ \\
        & 0.1
        & $0.7327\scriptscriptstyle \pm \scriptstyle5.82e\text{-}2$
        & $0.9931\scriptscriptstyle \pm \scriptstyle3.58e\text{-}3$
        & $0.9936\scriptscriptstyle \pm \scriptstyle5.51e\text{-}3$
        & $0.9936\scriptscriptstyle \pm \scriptstyle2.50e\text{-}3$
        & $0.8758\scriptscriptstyle \pm \scriptstyle3.95e\text{-}2$ \\
        & 0.2
        & $0.7120\scriptscriptstyle \pm \scriptstyle4.83e\text{-}2$
        & $\mathbf{0.9938\scriptscriptstyle \pm \scriptstyle2.40e\text{-}3}$
        & $0.9958\scriptscriptstyle \pm \scriptstyle1.95e\text{-}3$
        & $0.9931\scriptscriptstyle \pm \scriptstyle2.39e\text{-}3$
        & $\mathbf{0.8774\scriptscriptstyle \pm \scriptstyle2.86e\text{-}2}$ \\
        & 0.5
        & $0.6876\scriptscriptstyle \pm \scriptstyle5.09e\text{-}2$
        & $0.9899\scriptscriptstyle \pm \scriptstyle4.30e\text{-}3$
        & $0.9960\scriptscriptstyle \pm \scriptstyle2.28e\text{-}3$
        & $0.9909\scriptscriptstyle \pm \scriptstyle3.81e\text{-}3$
        & $0.8713\scriptscriptstyle \pm \scriptstyle3.37e\text{-}2$ \\
        \midrule
        \multirow{6}{*}{LDM\textsubscript{LAION-400M}}
        & 0.0
        & $0.6263\scriptscriptstyle \pm \scriptstyle7.51e\text{-}2$
        & $0.9565\scriptscriptstyle \pm \scriptstyle4.27e\text{-}2$
        & $0.9896\scriptscriptstyle \pm \scriptstyle8.87e\text{-}3$
        & $0.9758\scriptscriptstyle \pm \scriptstyle2.86e\text{-}2$
        & $0.8381\scriptscriptstyle \pm \scriptstyle4.87e\text{-}2$ \\
        & 0.03125
        & $0.6276\scriptscriptstyle \pm \scriptstyle6.30e\text{-}2$
        & $0.9970\scriptscriptstyle \pm \scriptstyle2.09e\text{-}3$
        & $0.9957\scriptscriptstyle \pm \scriptstyle3.08e\text{-}3$
        & $\mathbf{0.9997\scriptscriptstyle \pm \scriptstyle3.55e\text{-}4}$
        & $0.8563\scriptscriptstyle \pm \scriptstyle3.73e\text{-}2$ \\
        & 0.0625
        & $0.6389\scriptscriptstyle \pm \scriptstyle4.84e\text{-}2$
        & $\mathbf{0.9971\scriptscriptstyle \pm \scriptstyle1.94e\text{-}3}$
        & $\mathbf{0.9957\scriptscriptstyle \pm \scriptstyle3.73e\text{-}3}$
        & $0.9997\scriptscriptstyle \pm \scriptstyle4.27e\text{-}4$
        & $0.8642\scriptscriptstyle \pm \scriptstyle3.30e\text{-}2$ \\
        & 0.1
        & $\mathbf{0.6530\scriptscriptstyle \pm \scriptstyle4.87e\text{-}2}$
        & $0.9956\scriptscriptstyle \pm \scriptstyle3.43e\text{-}3$
        & $0.9940\scriptscriptstyle \pm \scriptstyle3.47e\text{-}3$
        & $0.9992\scriptscriptstyle \pm \scriptstyle7.47e\text{-}4$
        & $0.8676\scriptscriptstyle \pm \scriptstyle3.44e\text{-}2$ \\
        & 0.2
        & $0.6515\scriptscriptstyle \pm \scriptstyle5.97e\text{-}2$
        & $0.9919\scriptscriptstyle \pm \scriptstyle3.47e\text{-}3$
        & $0.9939\scriptscriptstyle \pm \scriptstyle2.97e\text{-}3$
        & $0.9977\scriptscriptstyle \pm \scriptstyle1.57e\text{-}3$
        & $\mathbf{0.8679\scriptscriptstyle \pm \scriptstyle3.98e\text{-}2}$ \\
        & 0.5
        & $0.6388\scriptscriptstyle \pm \scriptstyle3.67e\text{-}2$
        & $0.9905\scriptscriptstyle \pm \scriptstyle4.88e\text{-}3$
        & $0.9952\scriptscriptstyle \pm \scriptstyle2.17e\text{-}3$
        & $0.9973\scriptscriptstyle \pm \scriptstyle1.68e\text{-}3$
        & $0.8674\scriptscriptstyle \pm \scriptstyle2.79e\text{-}2$ \\
        \midrule
        \multirow{6}{*}{Glide\textsubscript{filtered-cc}}
        & 0.0
        & $0.6687\scriptscriptstyle \pm \scriptstyle8.82e\text{-}2$
        & $0.9629\scriptscriptstyle \pm \scriptstyle4.83e\text{-}2$
        & $0.9916\scriptscriptstyle \pm \scriptstyle6.77e\text{-}3$
        & $0.9814\scriptscriptstyle \pm \scriptstyle3.92e\text{-}2$
        & $0.8988\scriptscriptstyle \pm \scriptstyle4.55e\text{-}2$ \\
        & 0.03125
        & $0.6910\scriptscriptstyle \pm \scriptstyle4.59e\text{-}2$
        & $0.9974\scriptscriptstyle \pm \scriptstyle2.15e\text{-}3$
        & $\mathbf{0.9980\scriptscriptstyle \pm \scriptstyle1.48e\text{-}3}$
        & $0.9998\scriptscriptstyle \pm \scriptstyle4.35e\text{-}4$
        & $\mathbf{0.9197\scriptscriptstyle \pm \scriptstyle2.15e\text{-}2}$ \\
        & 0.0625
        & $\mathbf{0.6942\scriptscriptstyle \pm \scriptstyle7.30e\text{-}2}$
        & $0.9973\scriptscriptstyle \pm \scriptstyle2.94e\text{-}3$
        & $0.9951\scriptscriptstyle \pm \scriptstyle3.15e\text{-}3$
        & $0.9996\scriptscriptstyle \pm \scriptstyle7.15e\text{-}4$
        & $0.9191\scriptscriptstyle \pm \scriptstyle2.99e\text{-}2$ \\
        & 0.1
        & $0.6834\scriptscriptstyle \pm \scriptstyle6.43e\text{-}2$
        & $\mathbf{0.9974\scriptscriptstyle \pm \scriptstyle1.66e\text{-}3}$
        & $0.9949\scriptscriptstyle \pm \scriptstyle2.73e\text{-}3$
        & $0.9997\scriptscriptstyle \pm \scriptstyle4.78e\text{-}4$
        & $0.9185\scriptscriptstyle \pm \scriptstyle2.74e\text{-}2$ \\
        & 0.2
        & $0.6406\scriptscriptstyle \pm \scriptstyle4.52e\text{-}2$
        & $0.9965\scriptscriptstyle \pm \scriptstyle1.45e\text{-}3$
        & $0.9976\scriptscriptstyle \pm \scriptstyle1.20e\text{-}3$
        & $\mathbf{0.9998\scriptscriptstyle \pm \scriptstyle2.07e\text{-}4}$
        & $0.9164\scriptscriptstyle \pm \scriptstyle1.98e\text{-}2$ \\
        & 0.5
        & $0.6535\scriptscriptstyle \pm \scriptstyle6.54e\text{-}2$
        & $0.9926\scriptscriptstyle \pm \scriptstyle6.24e\text{-}3$
        & $0.9961\scriptscriptstyle \pm \scriptstyle2.96e\text{-}3$
        & $0.9989\scriptscriptstyle \pm \scriptstyle1.91e\text{-}3$
        & $0.9147\scriptscriptstyle \pm \scriptstyle2.90e\text{-}2$ \\
        \midrule
        \multirow{6}{*}{GALIP\textsubscript{coco}}
        & 0.0
        & --
        & $0.9982\scriptscriptstyle \pm \scriptstyle1.16e\text{-}3$
        & $0.9992\scriptscriptstyle \pm \scriptstyle7.43e\text{-}4$
        & $0.9996\scriptscriptstyle \pm \scriptstyle5.71e\text{-}4$
        & $0.9321\scriptscriptstyle \pm \scriptstyle2.71e\text{-}2$ \\
        & 0.03125
        & --
        & $0.9999\scriptscriptstyle \pm \scriptstyle1.10e\text{-}4$
        & $0.9994\scriptscriptstyle \pm \scriptstyle5.35e\text{-}4$
        & $1.0000\scriptscriptstyle \pm \scriptstyle0.0$
        & $0.9180\scriptscriptstyle \pm \scriptstyle2.45e\text{-}2$ \\
        & 0.0625
        & --
        & $\mathbf{0.9999\scriptscriptstyle \pm \scriptstyle7.00e\text{-}5}$
        & $0.9994\scriptscriptstyle \pm \scriptstyle5.14e\text{-}4$
        & $1.0000\scriptscriptstyle \pm \scriptstyle0.0$
        & $0.9272\scriptscriptstyle \pm \scriptstyle2.25e\text{-}2$ \\
        & 0.1
        & --
        & $0.9999\scriptscriptstyle \pm \scriptstyle1.35e\text{-}4$
        & $0.9993\scriptscriptstyle \pm \scriptstyle5.58e\text{-}4$
        & $1.0000\scriptscriptstyle \pm \scriptstyle0.0$
        & $0.9392\scriptscriptstyle \pm \scriptstyle1.52e\text{-}2$ \\
        & 0.2
        & --
        & $0.9999\scriptscriptstyle \pm \scriptstyle8.06e\text{-}5$
        & $0.9995\scriptscriptstyle \pm \scriptstyle4.41e\text{-}4$
        & $\mathbf{1.0000\scriptscriptstyle \pm \scriptstyle0.0}$
        & $0.9484\scriptscriptstyle \pm \scriptstyle1.53e\text{-}2$ \\
        & 0.5
        & --
        & $0.9997\scriptscriptstyle \pm \scriptstyle2.84e\text{-}4$
        & $\mathbf{0.9997\scriptscriptstyle \pm \scriptstyle2.68e\text{-}4}$
        & $1.0000\scriptscriptstyle \pm \scriptstyle3.00e\text{-}5$
        & $\mathbf{0.9513\scriptscriptstyle \pm \scriptstyle1.51e\text{-}2}$ \\
        \midrule
        \multirow{6}{*}{ProGan\textsubscript{lsun}}
        & 0.0
        & $0.9818\scriptscriptstyle \pm \scriptstyle1.24e\text{-}2$
        & --
        & $\mathbf{0.9278\scriptscriptstyle \pm \scriptstyle1.49e\text{-}2}$
        & $0.7993\scriptscriptstyle \pm \scriptstyle2.00e\text{-}2$
        & $0.9362\scriptscriptstyle \pm \scriptstyle1.66e\text{-}2$ \\
        & 0.03125
        & $0.9856\scriptscriptstyle \pm \scriptstyle9.97e\text{-}3$
        & --
        & $0.9065\scriptscriptstyle \pm \scriptstyle3.16e\text{-}2$
        & $\mathbf{0.9249\scriptscriptstyle \pm \scriptstyle1.62e\text{-}2}$
        & $0.9429\scriptscriptstyle \pm \scriptstyle2.79e\text{-}2$ \\
        & 0.0625
        & $0.9830\scriptscriptstyle \pm \scriptstyle1.18e\text{-}2$
        & --
        & $0.8478\scriptscriptstyle \pm \scriptstyle3.35e\text{-}2$
        & $0.9146\scriptscriptstyle \pm \scriptstyle1.04e\text{-}2$
        & $0.9298\scriptscriptstyle \pm \scriptstyle2.71e\text{-}2$ \\
        & 0.1
        & $0.9885\scriptscriptstyle \pm \scriptstyle7.41e\text{-}3$
        & --
        & $0.8471\scriptscriptstyle \pm \scriptstyle4.24e\text{-}2$
        & $0.8885\scriptscriptstyle \pm \scriptstyle1.16e\text{-}2$
        & $0.9369\scriptscriptstyle \pm \scriptstyle2.55e\text{-}2$ \\
        & 0.2
        & $\mathbf{0.9937\scriptscriptstyle \pm \scriptstyle3.71e\text{-}3}$
        & --
        & $0.8719\scriptscriptstyle \pm \scriptstyle3.09e\text{-}2$
        & $0.8456\scriptscriptstyle \pm \scriptstyle2.75e\text{-}2$
        & $0.9453\scriptscriptstyle \pm \scriptstyle1.71e\text{-}2$ \\
        & 0.5
        & $0.9936\scriptscriptstyle \pm \scriptstyle4.05e\text{-}3$
        & --
        & $0.9109\scriptscriptstyle \pm \scriptstyle3.60e\text{-}2$
        & $0.8698\scriptscriptstyle \pm \scriptstyle2.34e\text{-}2$
        & $\mathbf{0.9572\scriptscriptstyle \pm \scriptstyle1.77e\text{-}2}$ \\
        \midrule
        \multirow{6}{*}{StyleGan2\textsubscript{lsun}}
        & 0.0
        & $0.9849\scriptscriptstyle \pm \scriptstyle1.35e\text{-}2$
        & $0.8687\scriptscriptstyle \pm \scriptstyle3.78e\text{-}2$
        & --
        & $0.9543\scriptscriptstyle \pm \scriptstyle1.79e\text{-}2$
        & $0.9623\scriptscriptstyle \pm \scriptstyle2.15e\text{-}2$ \\
        & 0.03125
        & $0.9925\scriptscriptstyle \pm \scriptstyle4.85e\text{-}3$
        & $0.9585\scriptscriptstyle \pm \scriptstyle2.45e\text{-}2$
        & --
        & $0.9952\scriptscriptstyle \pm \scriptstyle2.71e\text{-}3$
        & $0.9803\scriptscriptstyle \pm \scriptstyle1.60e\text{-}2$ \\
        & 0.0625
        & $0.9917\scriptscriptstyle \pm \scriptstyle5.44e\text{-}3$
        & $\mathbf{0.9685\scriptscriptstyle \pm \scriptstyle1.35e\text{-}2}$
        & --
        & $\mathbf{0.9953\scriptscriptstyle \pm \scriptstyle2.49e\text{-}3}$
        & $0.9836\scriptscriptstyle \pm \scriptstyle1.37e\text{-}2$ \\
        & 0.1
        & $\mathbf{0.9926\scriptscriptstyle \pm \scriptstyle6.85e\text{-}3}$
        & $0.9649\scriptscriptstyle \pm \scriptstyle1.50e\text{-}2$
        & --
        & $0.9946\scriptscriptstyle \pm \scriptstyle1.80e\text{-}3$
        & $\mathbf{0.9851\scriptscriptstyle \pm \scriptstyle1.43e\text{-}2}$ \\
        & 0.2
        & $0.9924\scriptscriptstyle \pm \scriptstyle3.55e\text{-}3$
        & $0.9500\scriptscriptstyle \pm \scriptstyle1.69e\text{-}2$
        & --
        & $0.9889\scriptscriptstyle \pm \scriptstyle4.87e\text{-}3$
        & $0.9833\scriptscriptstyle \pm \scriptstyle8.43e\text{-}3$ \\
        & 0.5
        & $0.9912\scriptscriptstyle \pm \scriptstyle4.58e\text{-}3$
        & $0.9407\scriptscriptstyle \pm \scriptstyle2.38e\text{-}2$
        & --
        & $0.9885\scriptscriptstyle \pm \scriptstyle4.97e\text{-}3$
        & $0.9821\scriptscriptstyle \pm \scriptstyle1.02e\text{-}2$ \\
        \midrule
        \multirow{6}{*}{GauGan\textsubscript{coco}}
        & 0.0
        & $0.9901\scriptscriptstyle \pm \scriptstyle6.44e\text{-}3$
        & $0.6812\scriptscriptstyle \pm \scriptstyle6.34e\text{-}2$
        & $\mathbf{0.9676\scriptscriptstyle \pm \scriptstyle1.29e\text{-}2}$
        & --
        & $0.9368\scriptscriptstyle \pm \scriptstyle2.68e\text{-}2$ \\
        & 0.03125
        & $0.9862\scriptscriptstyle \pm \scriptstyle1.03e\text{-}2$
        & $0.6965\scriptscriptstyle \pm \scriptstyle5.18e\text{-}2$
        & $0.9225\scriptscriptstyle \pm \scriptstyle3.06e\text{-}2$
        & --
        & $0.8991\scriptscriptstyle \pm \scriptstyle3.69e\text{-}2$ \\
        & 0.0625
        & $0.9908\scriptscriptstyle \pm \scriptstyle4.86e\text{-}3$
        & $0.7132\scriptscriptstyle \pm \scriptstyle6.78e\text{-}2$
        & $0.8919\scriptscriptstyle \pm \scriptstyle3.83e\text{-}2$
        & --
        & $0.9051\scriptscriptstyle \pm \scriptstyle3.50e\text{-}2$ \\
        & 0.1
        & $0.9958\scriptscriptstyle \pm \scriptstyle3.90e\text{-}3$
        & $0.7752\scriptscriptstyle \pm \scriptstyle6.78e\text{-}2$
        & $0.9425\scriptscriptstyle \pm \scriptstyle3.78e\text{-}2$
        & --
        & $0.9370\scriptscriptstyle \pm \scriptstyle3.31e\text{-}2$ \\
        & 0.2
        & $\mathbf{0.9980\scriptscriptstyle \pm \scriptstyle1.97e\text{-}3}$
        & $\mathbf{0.8039\scriptscriptstyle \pm \scriptstyle4.45e\text{-}2}$
        & $0.9492\scriptscriptstyle \pm \scriptstyle1.26e\text{-}2$
        & --
        & $0.9535\scriptscriptstyle \pm \scriptstyle1.99e\text{-}2$ \\
        & 0.5
        & $0.9976\scriptscriptstyle \pm \scriptstyle2.05e\text{-}3$
        & $0.7709\scriptscriptstyle \pm \scriptstyle4.78e\text{-}2$
        & $0.9583\scriptscriptstyle \pm \scriptstyle1.30e\text{-}2$
        & --
        & $\mathbf{0.9515\scriptscriptstyle \pm \scriptstyle2.17e\text{-}2}$ \\
        \bottomrule[1pt]
        \end{tabular}
    
\label{table: Different Source Models Sup2}
\end{table*}

\begin{table*}[ht]
    \centering
    \footnotesize
        \caption{Evaluation sensitivity of different ways of doing ADA on different open world real image datasets. 1)In training phase, conditioned on doing ADA only on half of non-target image set, one of COCO, ImageNet, Flickr, or CC12M is used as the non-target dataset. 2) Conditioned on using non-target images selected from COCO, we do ADA on half of target image set, on half of the both non-target images and target images, and on half of the target images which are then treated as non-target ones.}
        \begin{tabular}{@{}l|cccccccccc@{}}
        \toprule[1pt]
        Methods & VQ-D & LDM & Glide & GALIP \\
        \midrule
        COCO+Neg 
            & $0.9703\scriptscriptstyle \pm \scriptstyle9.06e\text{-}3$
            & $0.8801\scriptscriptstyle \pm \scriptstyle2.06e\text{-}2$
            & $0.9519\scriptscriptstyle \pm \scriptstyle1.45e\text{-}2$
            & $\mathbf{0.9798\scriptscriptstyle \pm \scriptstyle1.18e\text{-}2}$ \\
        Flickr+Neg & $0.9586\scriptscriptstyle \pm \scriptstyle2.17e\text{-}2$
            & $0.8775\scriptscriptstyle \pm \scriptstyle2.86e\text{-}2$
            & $0.8805\scriptscriptstyle \pm \scriptstyle4.33e\text{-}2$
            & $0.9546\scriptscriptstyle \pm \scriptstyle1.89e\text{-}2$ \\
        CC12M+Neg & $0.8995\scriptscriptstyle \pm \scriptstyle2.68e\text{-}2$
            & $0.8523\scriptscriptstyle \pm \scriptstyle3.96e\text{-}2$
            & $0.9364\scriptscriptstyle \pm \scriptstyle1.74e\text{-}2$
            & $0.9444\scriptscriptstyle \pm \scriptstyle2.73e\text{-}2$ \\
        ImageNet+Neg & $\mathbf{0.9773\scriptscriptstyle \pm \scriptstyle7.67e\text{-}3}$
            & $\mathbf{0.9136\scriptscriptstyle \pm \scriptstyle2.51e\text{-}2}$
            & $\mathbf{0.9664\scriptscriptstyle \pm \scriptstyle1.39e\text{-}2}$
            & $0.9710\scriptscriptstyle \pm \scriptstyle2.34e\text{-}2$ \\
        \midrule
        COCO+None & $0.9503\scriptscriptstyle \pm \scriptstyle1.68e\text{-}2$
            & $0.8373\scriptscriptstyle \pm \scriptstyle5.33e\text{-}2$
            & $0.9266\scriptscriptstyle \pm \scriptstyle3.19e\text{-}2$
            & $0.9660\scriptscriptstyle \pm \scriptstyle2.56e\text{-}2$ \\
        \midrule
        COCO+Neg 
            & $\mathbf{0.9703\scriptscriptstyle \pm \scriptstyle9.06e\text{-}3}$
            & $0.8801\scriptscriptstyle \pm \scriptstyle2.06e\text{-}2$
            & $0.9519\scriptscriptstyle \pm \scriptstyle1.45e\text{-}2$
            & $0.9798\scriptscriptstyle \pm \scriptstyle1.18e\text{-}2$ \\
        COCO+Both & $0.9346\scriptscriptstyle \pm \scriptstyle2.32e\text{-}2$
            & $0.8229\scriptscriptstyle \pm \scriptstyle5.47e\text{-}2$
            & $0.9070\scriptscriptstyle \pm \scriptstyle2.81e\text{-}2$
            & $0.9655\scriptscriptstyle \pm \scriptstyle1.86e\text{-}2$ \\
        COCO+Target & $0.8779\scriptscriptstyle \pm \scriptstyle6.93e\text{-}2$
            & $0.7618\scriptscriptstyle \pm \scriptstyle8.77e\text{-}2$
            & $0.8728\scriptscriptstyle \pm \scriptstyle4.90e\text{-}2$
            & $0.9205\scriptscriptstyle \pm \scriptstyle3.83e\text{-}2$ \\
        COCO+T-NT & $0.9684\scriptscriptstyle \pm \scriptstyle2.45e\text{-}2$
            & $\mathbf{0.8818\scriptscriptstyle \pm \scriptstyle4.91e\text{-}2}$
            & $\mathbf{0.9522\scriptscriptstyle \pm \scriptstyle2.98e\text{-}2}$
            & $\mathbf{0.9803\scriptscriptstyle \pm \scriptstyle1.26e\text{-}2}$ \\
        \bottomrule[1pt]
        \end{tabular}
    
\label{table: Different Ways of Attacks Sup}
\end{table*}

\begin{table*}[ht]
    \centering
    \footnotesize
        \caption{Evaluation sensitivity of different ways of doing ADA on different open world real image datasets. 1)In training phase, conditioned on doing ADA only on half of non-target image set, one of COCO, ImageNet, Flickr, or CC12M is used as the non-target dataset. 2) Conditioned on using non-target images selected from COCO, we do ADA on half of target image set, on half of the both non-target images and target images, and on half of the target images which are then treated as non-target ones.}
\begin{tabular}{@{}l|cccccccccc@{}}
        \toprule[1pt]
        Methods & ProGan & StyleGan2 & GauGan & Overall \\
        \midrule
        COCO+Neg 
            & $0.9452\scriptscriptstyle \pm \scriptstyle3.14e\text{-}2$
            & $0.9651\scriptscriptstyle \pm \scriptstyle1.54e\text{-}2$
            & $0.9910\scriptscriptstyle \pm \scriptstyle7.32e\text{-}3$
            & $0.9548\scriptscriptstyle \pm \scriptstyle1.75e\text{-}2$ \\
        Flickr+Neg
            & $0.9494\scriptscriptstyle \pm \scriptstyle2.17e\text{-}2$
            & $0.9368\scriptscriptstyle \pm \scriptstyle3.01e\text{-}2$
            & $0.9908\scriptscriptstyle \pm \scriptstyle9.65e\text{-}3$
            & $0.9355\scriptscriptstyle \pm \scriptstyle2.67e\text{-}2$ \\
        CC12M+Neg
            & $0.9671\scriptscriptstyle \pm \scriptstyle1.32e\text{-}2$
            & $0.9422\scriptscriptstyle \pm \scriptstyle2.45e\text{-}2$
            & $0.9822\scriptscriptstyle \pm \scriptstyle6.75e\text{-}3$
            & $0.9320\scriptscriptstyle \pm \scriptstyle2.43e\text{-}2$ \\
        ImageNet+Neg
            & $\mathbf{0.9862\scriptscriptstyle \pm \scriptstyle4.11e\text{-}3}$
            & $\mathbf{0.9861\scriptscriptstyle \pm \scriptstyle4.22e\text{-}3}$
            & $\mathbf{0.9966\scriptscriptstyle \pm \scriptstyle1.16e\text{-}3}$
            & $\mathbf{0.9710\scriptscriptstyle \pm \scriptstyle1.44e\text{-}2}$ \\
        \midrule
        COCO+None 
            & $0.8861\scriptscriptstyle \pm \scriptstyle4.46e\text{-}2$
            & $0.9533\scriptscriptstyle \pm \scriptstyle2.30e\text{-}2$
            & $0.9643\scriptscriptstyle \pm \scriptstyle2.91e\text{-}2$
            & $0.9263\scriptscriptstyle \pm \scriptstyle3.41e\text{-}2$ \\
\end{tabular}
    
\label{table: Different Ways of Attacks Sup2}
\end{table*}

\begin{table*}[t]
    \centering
    \footnotesize
    
        \caption{Evaluation of OCC-CLIP Relative to the Proportion of Augmented Non-Target Images. Five proportions are investigated: \(0\%\), \(25\%\), \(50\%\), \(75\%\), and \(100\%\).}
        \begin{tabular}{@{}l|cccccccccc@{}}
        \toprule[1pt]
        Methods & VQ-D & LDM & Glide & GALIP \\
        \midrule
        OCC-CLIP 0\% & $0.9503\scriptscriptstyle \pm \scriptstyle1.68e\text{-}2$
            & $0.8373\scriptscriptstyle \pm \scriptstyle5.33e\text{-}2$
            & $0.9266\scriptscriptstyle \pm \scriptstyle3.19e\text{-}2$
            & $0.9660\scriptscriptstyle \pm \scriptstyle2.56e\text{-}2$ \\

        OCC-CLIP 25\% & $0.9558\scriptscriptstyle \pm \scriptstyle2.19e\text{-}2$
            & $0.8519\scriptscriptstyle \pm \scriptstyle4.65e\text{-}2$
            & $0.9356\scriptscriptstyle \pm \scriptstyle3.23e\text{-}2$
            & $0.9754\scriptscriptstyle \pm \scriptstyle1.55e\text{-}2$ \\

        OCC-CLIP 50\% & $\mathbf{0.9703\scriptscriptstyle \pm \scriptstyle9.06e\text{-}3}$
            & $0.8801\scriptscriptstyle \pm \scriptstyle2.06e\text{-}2$
            & $0.9519\scriptscriptstyle \pm \scriptstyle1.45e\text{-}2$
            & $0.9798\scriptscriptstyle \pm \scriptstyle1.18e\text{-}2$ \\

        OCC-CLIP 75\% & $0.9683\scriptscriptstyle \pm \scriptstyle1.25e\text{-}2$
            & $\mathbf{0.8828\scriptscriptstyle \pm \scriptstyle3.34e\text{-}2}$
            & $\mathbf{0.9531\scriptscriptstyle \pm \scriptstyle2.17e\text{-}2}$
            & $\mathbf{0.9815\scriptscriptstyle \pm \scriptstyle9.53e\text{-}3}$ \\

        OCC-CLIP 100\% & $0.9252\scriptscriptstyle \pm \scriptstyle1.72e\text{-}2$
            & $0.8414\scriptscriptstyle \pm \scriptstyle2.19e\text{-}2$
            & $0.8996\scriptscriptstyle \pm \scriptstyle2.84e\text{-}2$
            & $0.9614\scriptscriptstyle \pm \scriptstyle8.53e\text{-}3$ \\
        \bottomrule[1pt]
        \end{tabular}
    
\label{table: Different Proportion Sup}
\end{table*}

\begin{table*}[t]
    \centering
    \footnotesize
    
        \caption{Evaluation of OCC-CLIP Relative to the Proportion of Augmented Non-Target Images. Five proportions are investigated: \(0\%\), \(25\%\), \(50\%\), \(75\%\), and \(100\%\).}
        \begin{tabular}{@{}l|cccccccccc@{}}
        \toprule[1pt]
        Methods & ProGan & StyleGan2 & GauGan & Overall \\
        \midrule
        OCC-CLIP 0\% 
            & $0.8861\scriptscriptstyle \pm \scriptstyle4.46e\text{-}2$
            & $0.9533\scriptscriptstyle \pm \scriptstyle2.30e\text{-}2$
            & $0.9643\scriptscriptstyle \pm \scriptstyle2.91e\text{-}2$
            & $0.9263\scriptscriptstyle \pm \scriptstyle3.41e\text{-}2$ \\

        OCC-CLIP 25\% 
            & $0.9209\scriptscriptstyle \pm \scriptstyle5.46e\text{-}2$
            & $0.9561\scriptscriptstyle \pm \scriptstyle3.20e\text{-}2$
            & $0.9878\scriptscriptstyle \pm \scriptstyle1.15e\text{-}2$
            & $0.9405\scriptscriptstyle \pm \scriptstyle3.39e\text{-}2$ \\

        OCC-CLIP 50\% 
            & $0.9452\scriptscriptstyle \pm \scriptstyle3.14e\text{-}2$
            & $0.9651\scriptscriptstyle \pm \scriptstyle1.54e\text{-}2$
            & $0.9910\scriptscriptstyle \pm \scriptstyle7.32e\text{-}3$
            & $0.9548\scriptscriptstyle \pm \scriptstyle1.75e\text{-}2$ \\

        OCC-CLIP 75\% 
            & $\mathbf{0.9466\scriptscriptstyle \pm \scriptstyle2.23e\text{-}2}$
            & $\mathbf{0.9675\scriptscriptstyle \pm \scriptstyle1.58e\text{-}2}$
            & $\mathbf{0.9911\scriptscriptstyle \pm \scriptstyle5.11e\text{-}3}$
            & $\mathbf{0.9558\scriptscriptstyle \pm \scriptstyle1.93e\text{-}2}$ \\

        OCC-CLIP 100\% 
            & $0.9393\scriptscriptstyle \pm \scriptstyle1.86e\text{-}2$
            & $0.8674\scriptscriptstyle \pm \scriptstyle2.64e\text{-}2$
            & $0.9861\scriptscriptstyle \pm \scriptstyle5.64e\text{-}3$
            & $0.9172\scriptscriptstyle \pm \scriptstyle1.98e\text{-}2$ \\
        \bottomrule[1pt]
        \end{tabular}
    
\label{table: Different Proportion Sup2}
\end{table*}

\begin{table*}[t]
    \centering
    \footnotesize
        \caption{Evaluation of OCC-CLIP on Different Numbers of Shots. This table shows the mean origin attribution performance of CoOp and OCC-CLIP on 7 different testing tasks with a variable number of shots: 10, 20, 30, 40, 50, 100, and 200. The target dataset is SD, while the non-target dataset is from one of COCO, CC12M, Flickr, or ImageNet.}
        \begin{tabular}{@{}l|c|ccccccc@{}}
        \toprule
        Non-Target & Methods & 10 & 20 & 30 & 40 \\
        \midrule
        \multirow{2}{*}{COCO} 
            & CoOp 
            & $0.7771\scriptscriptstyle \pm \scriptstyle1.02e\text{-}1$
            & $0.8500\scriptscriptstyle \pm \scriptstyle8.36e\text{-}2$
            & $0.8836\scriptscriptstyle \pm \scriptstyle4.45e\text{-}2$
            & $0.9183\scriptscriptstyle \pm \scriptstyle4.57e\text{-}2$ \\
            & OCC-CLIP
            & $\mathbf{0.8842\scriptscriptstyle \pm \scriptstyle7.22e\text{-}2}$
            & $\mathbf{0.9293\scriptscriptstyle \pm \scriptstyle4.45e\text{-}2}$
            & $\mathbf{0.9435\scriptscriptstyle \pm \scriptstyle2.23e\text{-}2}$
            & $\mathbf{0.9494\scriptscriptstyle \pm \scriptstyle2.62e\text{-}2}$ \\
        \midrule
        \multirow{2}{*}{Flickr} 
            & CoOp 
            & $0.7905\scriptscriptstyle \pm \scriptstyle7.82e\text{-}2$
            & $0.8649\scriptscriptstyle \pm \scriptstyle7.54e\text{-}2$
            & $0.8784\scriptscriptstyle \pm \scriptstyle4.39e\text{-}2$
            & $0.8734\scriptscriptstyle \pm \scriptstyle8.64e\text{-}2$ \\
            & OCC-CLIP
            & $\mathbf{0.8921\scriptscriptstyle \pm \scriptstyle8.26e\text{-}2}$
            & $\mathbf{0.9231\scriptscriptstyle \pm \scriptstyle3.86e\text{-}2}$
            & $\mathbf{0.9360\scriptscriptstyle \pm \scriptstyle2.32e\text{-}2}$
            & $\mathbf{0.9358\scriptscriptstyle \pm \scriptstyle3.88e\text{-}2}$\\
        \midrule
        \multirow{2}{*}{CC12M} 
            & CoOp 
            & $0.7737\scriptscriptstyle \pm \scriptstyle8.18e\text{-}2$
            & $0.8302\scriptscriptstyle \pm \scriptstyle7.79e\text{-}2$
            & $0.8640\scriptscriptstyle \pm \scriptstyle5.67e\text{-}2$
            & $0.8582\scriptscriptstyle \pm \scriptstyle4.77e\text{-}2$ \\
\bottomrule
        \end{tabular}
    
    \label{table: Number of Shots Sup}
\end{table*}

\begin{table*}[t]
    \centering
    \footnotesize
        \caption{Evaluation of OCC-CLIP on Different Numbers of Shots. This table shows the mean origin attribution performance of CoOp and OCC-CLIP on 7 different testing tasks with a variable number of shots: 10, 20, 30, 40, 50, 100, and 200. The target dataset is SD, while the non-target dataset is from one of COCO, CC12M, Flickr, or ImageNet.}
        \begin{tabular}{@{}l|c|ccccccc@{}}
        \toprule
        Non-Target & Methods & 50 & 100 & 200 \\
        \midrule
        \multirow{2}{*}{COCO} 
            & CoOp 
            & $0.9263\scriptscriptstyle \pm \scriptstyle3.41e\text{-}2$
            & $0.9541\scriptscriptstyle \pm \scriptstyle2.21e\text{-}2$
            & $0.9601\scriptscriptstyle \pm \scriptstyle2.00e\text{-}2$ \\
            & OCC-CLIP
            & $\mathbf{0.9548\scriptscriptstyle \pm \scriptstyle1.75e\text{-}2}$
            & $\mathbf{0.9684\scriptscriptstyle \pm \scriptstyle1.56e\text{-}2}$
            & $\mathbf{0.9746\scriptscriptstyle \pm \scriptstyle1.08e\text{-}2}$ \\
        \midrule
        \multirow{2}{*}{Flickr} 
            & CoOp 
            & $0.8788\scriptscriptstyle \pm \scriptstyle7.17e\text{-}2$
            & $0.9071\scriptscriptstyle \pm \scriptstyle3.68e\text{-}2$
            & $0.9177\scriptscriptstyle \pm \scriptstyle3.46e\text{-}2$ \\
            & OCC-CLIP
            & $\mathbf{0.9355\scriptscriptstyle \pm \scriptstyle2.67e\text{-}2}$ 
            & $\mathbf{0.9531\scriptscriptstyle \pm \scriptstyle2.88e\text{-}2}$
            & $\mathbf{0.9588\scriptscriptstyle \pm \scriptstyle3.35e\text{-}2}$ \\
        \midrule
        \multirow{2}{*}{CC12M} 
            & CoOp 
            & $0.8689\scriptscriptstyle \pm \scriptstyle4.27e\text{-}2$
            & $0.8973\scriptscriptstyle \pm \scriptstyle3.62e\text{-}2$
            & $0.8922\scriptscriptstyle \pm \scriptstyle3.70e\text{-}2$ \\
            & OCC-CLIP
            & $\mathbf{0.9320\scriptscriptstyle \pm \scriptstyle2.43e\text{-}2}$
            & $\mathbf{0.9342\scriptscriptstyle \pm \scriptstyle3.18e\text{-}2}$
            & $\mathbf{0.9331\scriptscriptstyle \pm \scriptstyle3.55e\text{-}2}$ \\
\bottomrule
        \end{tabular}
    
    \label{table: Number of Shots Sup2}
\end{table*}

\begin{table*}[t]
    \centering
    \scriptsize
        \caption{Evaluation of OCC-CLIP under Image Processing. Six image processing methods are executed: Gaussian Blur, Gaussian Noise, Grayscale, Rotation, Flip, and a mixture of all the these data augmentation methods.}
        \begin{tabular}{l|c|cccccc}
        \toprule[1pt]
        Process & Method
        & VQ-D
        & LDM
        & Glide  
        & GALIP \\
        \midrule
        \multirow{2}{*}{None} 
            & CoOp 
            & $0.9503\scriptscriptstyle \pm \scriptstyle1.68e\text{-}2$
            & $0.8373\scriptscriptstyle \pm \scriptstyle5.33e\text{-}2$
            & $0.9266\scriptscriptstyle \pm \scriptstyle3.19e\text{-}2$
            & $0.9660\scriptscriptstyle \pm \scriptstyle2.56e\text{-}2$ \\
            & OCC-CLIP 
            & $\mathbf{0.9703\scriptscriptstyle \pm \scriptstyle9.06e\text{-}3}$
            & $\mathbf{0.8801\scriptscriptstyle \pm \scriptstyle2.06e\text{-}2}$
            & $\mathbf{0.9519\scriptscriptstyle \pm \scriptstyle1.45e\text{-}2}$
            & $\mathbf{0.9798\scriptscriptstyle \pm \scriptstyle1.18e\text{-}2}$ \\
        \midrule 
        \multirow{2}{*}{Gaussian Blur} 
            & CoOp 
            & $0.8918\scriptscriptstyle \pm \scriptstyle3.15e\text{-}2$
            & $0.7067\scriptscriptstyle \pm \scriptstyle6.27e\text{-}2$
            & $0.8555\scriptscriptstyle \pm \scriptstyle3.86e\text{-}2$
            & $0.9240\scriptscriptstyle \pm \scriptstyle4.39e\text{-}2$ \\
            & OCC-CLIP 
            & $\mathbf{0.9286\scriptscriptstyle \pm \scriptstyle2.54e\text{-}2}$
            & $\mathbf{0.7700\scriptscriptstyle \pm \scriptstyle4.71e\text{-}2}$
            & $\mathbf{0.8914\scriptscriptstyle \pm \scriptstyle3.59e\text{-}2}$
            & $\mathbf{0.9502\scriptscriptstyle \pm \scriptstyle2.53e\text{-}2}$ \\
        \midrule 
        \multirow{2}{*}{Gaussian Noise} 
            & CoOp 
            & $0.9159\scriptscriptstyle \pm \scriptstyle3.99e\text{-}2$
            & $0.7646\scriptscriptstyle \pm \scriptstyle8.89e\text{-}2$
            & $0.8847\scriptscriptstyle \pm \scriptstyle5.81e\text{-}2$
            & $0.9374\scriptscriptstyle \pm \scriptstyle4.72e\text{-}2$ \\
\end{tabular}
    
\label{Table: Defence Against Potential Attack Sup}
\end{table*}

\begin{table*}[t]
    \centering
    \scriptsize
        \caption{Evaluation of OCC-CLIP under Image Processing. Six image processing methods are executed: Gaussian Blur, Gaussian Noise, Grayscale, Rotation, Flip, and a mixture of all the these data augmentation methods.}
        \begin{tabular}{l|l|ccccccccccc}
        \toprule[1pt]
        Process & Method
        & ProGan
        & StyleGan2
        & GauGan
        & Overall \\
        \midrule
        \multirow{2}{*}{None} 
            & CoOp 
            & $0.8861\scriptscriptstyle \pm \scriptstyle4.46e\text{-}2$
            & $0.9533\scriptscriptstyle \pm \scriptstyle2.30e\text{-}2$
            & $0.9643\scriptscriptstyle \pm \scriptstyle2.91e\text{-}2$
            & $0.9263\scriptscriptstyle \pm \scriptstyle3.41e\text{-}2$ \\
            & OCC-CLIP 
            & $\mathbf{0.9452\scriptscriptstyle \pm \scriptstyle3.14e\text{-}2}$
            & $\mathbf{0.9651\scriptscriptstyle \pm \scriptstyle1.54e\text{-}2}$
            & $\mathbf{0.9910\scriptscriptstyle \pm \scriptstyle7.32e\text{-}3}$
            & $\mathbf{0.9548\scriptscriptstyle \pm \scriptstyle1.75e\text{-}2}$ \\
        \midrule 
        \multirow{2}{*}{Gaussian Blur} 
            & CoOp 
            & $0.7772\scriptscriptstyle \pm \scriptstyle8.53e\text{-}2$
            & $0.9042\scriptscriptstyle \pm \scriptstyle3.74e\text{-}2$
            & $0.9181\scriptscriptstyle \pm \scriptstyle6.47e\text{-}2$
            & $0.8539\scriptscriptstyle \pm \scriptstyle5.50e\text{-}2$ \\
            & OCC-CLIP 
            & $\mathbf{0.8796\scriptscriptstyle \pm \scriptstyle4.99e\text{-}2}$
            & $\mathbf{0.9072\scriptscriptstyle \pm \scriptstyle4.48e\text{-}2}$
            & $\mathbf{0.9743\scriptscriptstyle \pm \scriptstyle1.54e\text{-}2}$
            & $\mathbf{0.9002\scriptscriptstyle \pm \scriptstyle3.69e\text{-}2}$ \\
        \midrule 
        \multirow{2}{*}{Gaussian Noise} 
            & CoOp 
            & $0.8298\scriptscriptstyle \pm \scriptstyle6.86e\text{-}2$
            & $0.9269\scriptscriptstyle \pm \scriptstyle3.56e\text{-}2$
            & $0.9418\scriptscriptstyle \pm \scriptstyle5.39e\text{-}2$
            & $0.8859\scriptscriptstyle \pm \scriptstyle5.85e\text{-}2$ \\
\end{tabular}
    
\label{Table: Defence Against Potential Attack Sup2}
\end{table*}

\begin{table*}[t]
    \centering
    \scriptsize
        \caption{Evaluation sensitivity to choice of prompts. This table shows the performance of OCC-CLIP and CoOp under different choices of label pairs.}
        \begin{tabular}{@{}l|c|ccccccc@{}}
        \toprule
        Names of Classes & Methods
        & VQ-D
        & LDM
        & Glide
        & GALIP \\
        \midrule
        \multirow{2}{*}{fake-real}
        & CoOp
        & $0.9432\scriptscriptstyle \pm \scriptstyle2.04e\text{-}2$
        & $0.8224\scriptscriptstyle \pm \scriptstyle5.69e\text{-}2$
        & $0.9128\scriptscriptstyle \pm \scriptstyle3.87e\text{-}2$
        & $0.9631\scriptscriptstyle \pm \scriptstyle1.99e\text{-}2$ \\
        & OCC-CLIP
        & $\mathbf{0.9469\scriptscriptstyle \pm \scriptstyle2.19e\text{-}2}$
        & $\mathbf{0.8324\scriptscriptstyle \pm \scriptstyle4.15e\text{-}2}$
        & $\mathbf{0.9215\scriptscriptstyle \pm \scriptstyle2.33e\text{-}2}$
        & $\mathbf{0.9693\scriptscriptstyle \pm \scriptstyle1.58e\text{-}2}$ \\
        \midrule
        \multirow{2}{*}{negative-positive}
        & CoOp
        & $0.9521\scriptscriptstyle \pm \scriptstyle2.18e\text{-}2$
        & $0.8578\scriptscriptstyle \pm \scriptstyle4.72e\text{-}2$
        & $0.9180\scriptscriptstyle \pm \scriptstyle4.11e\text{-}2$
        & $0.9610\scriptscriptstyle \pm \scriptstyle2.61e\text{-}2$ \\
        & OCC-CLIP
        & $\mathbf{0.9552\scriptscriptstyle \pm \scriptstyle2.46e\text{-}2}$
        & $\mathbf{0.8492\scriptscriptstyle \pm \scriptstyle4.55e\text{-}2}$
        & $\mathbf{0.9163\scriptscriptstyle \pm \scriptstyle3.34e\text{-}2}$
        & $\mathbf{0.9597\scriptscriptstyle \pm \scriptstyle2.22e\text{-}2}$ \\
        \midrule
        \multirow{2}{*}{other-this}
        & CoOp
        & $0.9543\scriptscriptstyle \pm \scriptstyle2.99e\text{-}2$
        & $0.8510\scriptscriptstyle \pm \scriptstyle6.57e\text{-}2$
        & $0.9316\scriptscriptstyle \pm \scriptstyle3.63e\text{-}2$
        & $0.9727\scriptscriptstyle \pm \scriptstyle2.51e\text{-}2$ \\
\bottomrule
        \end{tabular}
    
    \label{table: Different Prompts Sup}
\end{table*}

\begin{table*}[t]
    \centering
    \scriptsize
        \caption{Evaluation sensitivity to choice of prompts. This table shows the performance of OCC-CLIP and CoOp under different choices of label pairs.}
        \begin{tabular}{@{}l|c|ccccccc@{}}
        \toprule
        Names of Classes & Methods
        & ProGan
        & StyleGan2
        & GauGan
        & Overall \\
        \midrule
        \multirow{2}{*}{fake-real}
        & CoOp
        & $0.8220\scriptscriptstyle \pm \scriptstyle5.78e\text{-}2$
        & $0.9431\scriptscriptstyle \pm \scriptstyle2.01e\text{-}2$
        & $0.9477\scriptscriptstyle \pm \scriptstyle2.54e\text{-}2$
        & $0.9078\scriptscriptstyle \pm \scriptstyle3.77e\text{-}2$ \\
        & OCC-CLIP
        & $\mathbf{0.8847\scriptscriptstyle \pm \scriptstyle3.59e\text{-}2}$
        & $\mathbf{0.9473\scriptscriptstyle \pm \scriptstyle1.90e\text{-}2}$
        & $\mathbf{0.9734\scriptscriptstyle \pm \scriptstyle8.03e\text{-}3}$
        & $\mathbf{0.9251\scriptscriptstyle \pm \scriptstyle2.60e\text{-}2}$ \\
        \midrule
        \multirow{2}{*}{negative-positive}
        & CoOp
        & $0.8228\scriptscriptstyle \pm \scriptstyle7.85e\text{-}2$
        & $0.9422\scriptscriptstyle \pm \scriptstyle3.87e\text{-}2$
        & $0.9549\scriptscriptstyle \pm \scriptstyle2.33e\text{-}2$
        & $0.9155\scriptscriptstyle \pm \scriptstyle4.36e\text{-}2$ \\
        & OCC-CLIP
        & $\mathbf{0.9120\scriptscriptstyle \pm \scriptstyle2.99e\text{-}2}$
        & $\mathbf{0.9441\scriptscriptstyle \pm \scriptstyle2.86e\text{-}2}$
        & $\mathbf{0.9857\scriptscriptstyle \pm \scriptstyle6.60e\text{-}3}$
        & $\mathbf{0.9317\scriptscriptstyle \pm \scriptstyle2.94e\text{-}2}$ \\
        \midrule
        \multirow{2}{*}{other-this}
        & CoOp
        & $0.8596\scriptscriptstyle \pm \scriptstyle6.79e\text{-}2$
        & $0.9355\scriptscriptstyle \pm \scriptstyle4.04e\text{-}2$
        & $0.9542\scriptscriptstyle \pm \scriptstyle2.79e\text{-}2$
        & $0.9227\scriptscriptstyle \pm \scriptstyle4.50e\text{-}2$ \\
\bottomrule
        \end{tabular}
    
    \label{table: Different Prompts Sup2}
\end{table*}

\begin{table*}[t]
    \centering
    \footnotesize
    
        \caption{ADA vs. other data augmentation methods. Different data augmentation methods are used during the training phase.}
        \begin{tabular}{@{}l|cccccccc@{}}
        \toprule
        Methods 
        & VQ-D
        & LDM
        & Glide
        & GALIP \\
        \midrule
        None & $0.9503\scriptscriptstyle \pm \scriptstyle1.68e\text{-}2$
            & $0.8373\scriptscriptstyle \pm \scriptstyle5.33e\text{-}2$
            & $0.9266\scriptscriptstyle \pm \scriptstyle3.19e\text{-}2$
            & $0.9660\scriptscriptstyle \pm \scriptstyle2.56e\text{-}2$ \\
        Gaussian Blur & $0.7780\scriptscriptstyle \pm \scriptstyle3.10e\text{-}2$
            & $0.6417\scriptscriptstyle \pm \scriptstyle5.40e\text{-}2$
            & $0.8460\scriptscriptstyle \pm \scriptstyle3.84e\text{-}2$
            & $0.8034\scriptscriptstyle \pm \scriptstyle4.79e\text{-}2$ \\
        Gaussian Noise & $0.7673\scriptscriptstyle \pm \scriptstyle3.51e\text{-}2$
            & $0.6078\scriptscriptstyle \pm \scriptstyle4.78e\text{-}2$
            & $0.7735\scriptscriptstyle \pm \scriptstyle4.63e\text{-}2$
            & $0.7004\scriptscriptstyle \pm \scriptstyle4.42e\text{-}2$ \\
        Grayscale & $0.7490\scriptscriptstyle \pm \scriptstyle4.32e\text{-}2$
            & $0.6111\scriptscriptstyle \pm \scriptstyle4.53e\text{-}2$
            & $0.7794\scriptscriptstyle \pm \scriptstyle3.65e\text{-}2$
            & $0.6877\scriptscriptstyle \pm \scriptstyle4.09e\text{-}2$ \\
        Rotate & $0.7653\scriptscriptstyle \pm \scriptstyle4.36e\text{-}2$
            & $0.6174\scriptscriptstyle \pm \scriptstyle4.94e\text{-}2$
            & $0.8029\scriptscriptstyle \pm \scriptstyle4.12e\text{-}2$
            & $0.7496\scriptscriptstyle \pm \scriptstyle5.03e\text{-}2$ \\
        Flip & $0.7614\scriptscriptstyle \pm \scriptstyle4.44e\text{-}2$
            & $0.6182\scriptscriptstyle \pm \scriptstyle4.77e\text{-}2$
            & $0.7880\scriptscriptstyle \pm \scriptstyle4.67e\text{-}2$
            & $0.7370\scriptscriptstyle \pm \scriptstyle3.71e\text{-}2$ \\
\bottomrule
        \end{tabular}
    
\label{table: Differnt Data Augmentation sup}
\end{table*}

\begin{table*}[t]
    \centering
    \footnotesize
    
        \caption{ADA vs. other data augmentation methods. Different data augmentation methods are used during training phase.}
        \begin{tabular}{@{}l|cccccccc@{}}
        \toprule
        Methods 
        & ProGan
        & StyleGan2
        & GauGan
        & Overall \\
        \midrule
        None & $0.8861\scriptscriptstyle \pm \scriptstyle4.46e\text{-}2$ 
            & $0.9533\scriptscriptstyle \pm \scriptstyle2.30e\text{-}2$
            & $0.9643\scriptscriptstyle \pm \scriptstyle2.91e\text{-}2$
            & $0.9263\scriptscriptstyle \pm \scriptstyle3.41e\text{-}2$ \\
        Gaussian Blur & $0.9856\scriptscriptstyle \pm \scriptstyle1.19e\text{-}2$
            & $0.9978\scriptscriptstyle \pm \scriptstyle1.87e\text{-}3$
            & $0.9949\scriptscriptstyle \pm \scriptstyle5.02e\text{-}3$
            & $0.8639\scriptscriptstyle \pm \scriptstyle3.34e\text{-}2$ \\
        Gaussian Noise & $0.9926\scriptscriptstyle \pm \scriptstyle5.84e\text{-}3$
            & $0.9974\scriptscriptstyle \pm \scriptstyle2.32e\text{-}3$
            & $0.9991\scriptscriptstyle \pm \scriptstyle1.21e\text{-}3$
            & $0.8340\scriptscriptstyle \pm \scriptstyle3.31e\text{-}2$ \\
        Grayscale & $0.9916\scriptscriptstyle \pm \scriptstyle4.96e\text{-}3$
            & $0.9978\scriptscriptstyle \pm \scriptstyle1.27e\text{-}3$
            & $0.9988\scriptscriptstyle \pm \scriptstyle9.65e\text{-}4$
            & $0.8308\scriptscriptstyle \pm \scriptstyle3.15e\text{-}2$ \\
        Rotate & $0.9930\scriptscriptstyle \pm \scriptstyle2.77e\text{-}3$
            & $0.9985\scriptscriptstyle \pm \scriptstyle9.18e\text{-}4$
            & $0.9985\scriptscriptstyle \pm \scriptstyle9.73e\text{-}4$
            & $0.8465\scriptscriptstyle \pm \scriptstyle3.50e\text{-}2$ \\
        Flip & $0.9936\scriptscriptstyle \pm \scriptstyle3.06e\text{-}3$
            & $0.9983\scriptscriptstyle \pm \scriptstyle1.45e\text{-}3$
            & $0.9985\scriptscriptstyle \pm \scriptstyle1.02e\text{-}3$
            & $0.8422\scriptscriptstyle \pm \scriptstyle3.34e\text{-}2$ \\
\bottomrule
        \end{tabular}

\label{table: Differnt Data Augmentation sup2}
\end{table*}

\begin{table*}[!htbp]
    \centering
    \footnotesize
        \caption{Evaluation of OCC-CLIP with commercial generation API. During the training phase, the target images are generated by DALL·E-3, and the non-target images are from COCO.}
        \begin{tabular}{@{}c|c|ccccccccc@{}}
        \toprule
        No. of Shots & Methods 
        & SD
        & VQ-D
        & LDM
        & Glide  
        & GALIP  \\
        \midrule
        \multirow{2}{*}{10}
        & CoOp 
        & $0.9146\scriptscriptstyle \pm \scriptstyle0.0$
        & $0.9465\scriptscriptstyle \pm \scriptstyle0.0$
        & $0.9213\scriptscriptstyle \pm \scriptstyle0.0$
        & $0.9235\scriptscriptstyle \pm \scriptstyle0.0$
        & $0.9636\scriptscriptstyle \pm \scriptstyle0.0$ \\
        & OCC-CLIP 
        & $\mathbf{0.9244\scriptscriptstyle \pm \scriptstyle0.0}$
        & $\mathbf{0.9922\scriptscriptstyle \pm \scriptstyle0.0}$
        & $\mathbf{0.9852\scriptscriptstyle \pm \scriptstyle0.0}$
        & $\mathbf{0.9874\scriptscriptstyle \pm \scriptstyle0.0}$
        & $\mathbf{0.9969\scriptscriptstyle \pm \scriptstyle0.0}$ \\
        \midrule
        \multirow{2}{*}{20}
        & CoOp 
        & $0.9692\scriptscriptstyle \pm \scriptstyle0.0$
        & $0.9881\scriptscriptstyle \pm \scriptstyle0.0$
        & $0.9779\scriptscriptstyle \pm \scriptstyle0.0$
        & $0.9713\scriptscriptstyle \pm \scriptstyle0.0$
        & $0.9924\scriptscriptstyle \pm \scriptstyle0.0$ \\
        & OCC-CLIP 
        & $\mathbf{0.9604\scriptscriptstyle \pm \scriptstyle0.0}$
        & $\mathbf{0.9967\scriptscriptstyle \pm \scriptstyle0.0}$
        & $\mathbf{0.9867\scriptscriptstyle \pm \scriptstyle0.0}$
        & $\mathbf{0.9888\scriptscriptstyle \pm \scriptstyle0.0}$
        & $\mathbf{0.9984\scriptscriptstyle \pm \scriptstyle0.0}$ \\
        \midrule
        \multirow{2}{*}{30}
        & CoOp 
        & $0.9338\scriptscriptstyle \pm \scriptstyle0.0$
        & $0.9817\scriptscriptstyle \pm \scriptstyle0.0$
        & $0.9824\scriptscriptstyle \pm \scriptstyle0.0$
        & $0.9748\scriptscriptstyle \pm \scriptstyle0.0$
        & $0.9970\scriptscriptstyle \pm \scriptstyle0.0$ \\
\bottomrule
        \end{tabular}
    
\label{table: Real Application Sup}
\end{table*}

\begin{table}[h!]
    \centering
    \scriptsize 
    \caption{Additional Baselines (tested on VQ-D, LDM, Glide, GALIP, ProGAN, StyleGAN2, GauGAN). The table shows the overall average \textbf{accuracy} tested on 7 different testing tasks.}
        \setlength{\tabcolsep}{5pt}
        \begin{tabular}{@{}l|cccccccc@{}}
        \toprule[1pt]
        Methods & ViT-L & ViT-Res & CLIP-LoRAm & CLIP-LoRAl \\ 
        \midrule 
        Average ACC & $0.5966$
        & $0.6269$
        & $0.5646$
        & $0.5507$ \\
        \bottomrule[1pt]
        \toprule[1pt]
        Methods & CLIP-LoRAa & Girish \textit{et al.} & Wang \textit{et al.} & OCC-CLIP\\ 
        \midrule 
        Average ACC & $0.5951$
        & $0.6079$
        & $0.7600$
        & $\mathbf{0.8824}$ \\
        \bottomrule[1pt]
        \end{tabular}
    \label{table: Additional Baselines}
\end{table}

\begin{table*}[!htbp]
    \centering
    \footnotesize
        \caption{Evaluation of OCC-CLIP with commercial generation API. During the training phase, the target images are generated by DALL·E-3, and the non-target images are from COCO.}
        \begin{tabular}{@{}c|c|ccccccccc@{}}
        \toprule
        No. of Shots & Methods 
        & ProGan
        & StyleGan2
        & GauGan 
        & Overall \\
        \midrule
        \multirow{2}{*}{10}
        & CoOp 
        & $0.8609\scriptscriptstyle \pm \scriptstyle0.0$
        & $0.9700\scriptscriptstyle \pm \scriptstyle0.0$
        & $0.8723\scriptscriptstyle \pm \scriptstyle0.0$
        & $0.9216\scriptscriptstyle \pm \scriptstyle0.0$ \\
        & OCC-CLIP 
        & $\mathbf{0.9620\scriptscriptstyle \pm \scriptstyle0.0}$
        & $\mathbf{0.9795\scriptscriptstyle \pm \scriptstyle0.0}$
        & $\mathbf{0.9756\scriptscriptstyle \pm \scriptstyle0.0}$
        & $\mathbf{0.9754\scriptscriptstyle \pm \scriptstyle0.0}$ \\
        \midrule
        \multirow{2}{*}{20}
        & CoOp 
        & $0.9108\scriptscriptstyle \pm \scriptstyle0.0$
        & $\mathbf{0.9940\scriptscriptstyle \pm \scriptstyle0.0}$
        & $0.9460\scriptscriptstyle \pm \scriptstyle0.0$
        & $0.9687\scriptscriptstyle \pm \scriptstyle0.0$ \\
        & OCC-CLIP 
        & $\mathbf{0.9786\scriptscriptstyle \pm \scriptstyle0.0}$
        & $0.9807\scriptscriptstyle \pm \scriptstyle0.0$
        & $\mathbf{0.9879\scriptscriptstyle \pm \scriptstyle0.0}$
        & $\mathbf{0.9848\scriptscriptstyle \pm \scriptstyle0.0}$ \\
        \midrule
        \multirow{2}{*}{30}
        & CoOp 
        & $0.8908\scriptscriptstyle \pm \scriptstyle0.0$
        & $0.9936\scriptscriptstyle \pm \scriptstyle0.0$
        & $0.8959\scriptscriptstyle \pm \scriptstyle0.0$
        & $0.9563\scriptscriptstyle \pm \scriptstyle0.0$ \\
        & OCC-CLIP 
        & $\mathbf{0.9935\scriptscriptstyle \pm \scriptstyle0.0}$
        & $\mathbf{0.9968\scriptscriptstyle \pm \scriptstyle0.0}$
        & $\mathbf{0.9992\scriptscriptstyle \pm \scriptstyle0.0}$
        & $\mathbf{0.9936\scriptscriptstyle \pm \scriptstyle0.0}$ \\
\bottomrule
        \end{tabular}
    
\label{table: Real Application Sup2}
\end{table*}

\begin{table*}[t]
    \centering
    \scriptsize
        \caption{Employing an ensemble of one-class classifiers for multi-class classification tasks. The following scenarios are considered: 2 classes, 4 classes, and 6 classes. The standard deviations of all the values are less than 0.04.}
        \vspace{-0.1cm}
        \begin{tabular}{l|c|cccccccccc}
        \toprule[1pt]
        Num of Class & Method 
        & LDM
        & Glide
        & GALIP
        & ProGAN
        & StyleGAN2
        & GauGAN
        & Overall \\
        \midrule
        \multirow{2}{*}{2 classes} 
            & CoOp
            & $0.6366$
             & $0.6448$
             & $0.6378$
             & $0.6928$
             & $0.8706$
             & $0.7259$
            & $0.7014$ \\ 
            & OCC-CLIP 
            & $\mathbf{0.6449}$
 & $\mathbf{0.6499}$
 & $\mathbf{0.6457}$
 & $\mathbf{0.8742}$
 & $\mathbf{0.8938}$
 & $\mathbf{0.9255}$
            & $\mathbf{0.7723}$ \\
        \midrule 
        \multirow{2}{*}{4 classes} 
            & CoOp
            & -- 
            & -- 
            & $0.5863$
 & $0.6124$
 & $0.7004$
 & $0.6376$
            & $0.6342$ \\
\end{tabular}
        \label{Table: OCC Multi Class}
\end{table*}

\begin{table*}[h]
    \centering
    \scriptsize
        \caption{Directly training a multi-class classifier based on the OCC-CLIP or CoOp. The following scenarios are considered: 2 classes, 4 classes, and 6 classes. The standard deviations of all the values are less than 0.04.}
        \begin{tabular}{l|c|cccccccccc}
        \toprule[1pt]
        Num of Class & Method 
        & LDM
        & Glide
        & GALIP
        & ProGan
        & StyleGan2
        & GauGan
        & Overall \\
        \midrule
        \multirow{2}{*}{2 classes} 
            & CoOp
            & $0.6312$
            & $\mathbf{0.6397}$
            & $0.6336$
            & $0.7150$
            & $\mathbf{0.8653}$
            & $0.8228$
            & $0.7179$ \\ 
            & OCC-CLIP 
            & $\mathbf{0.6416}$
            & $0.6375$
            & $\mathbf{0.6382}$
            & $\mathbf{0.7921}$
            & $0.8491$
            & $\mathbf{0.8991}$
            & $\mathbf{0.7429}$ \\ 
        \midrule 
        \multirow{2}{*}{4 classes} 
            & CoOp
            & -- 
            & -- 
            & $0.6858$
            & $0.7242$
            & $0.8076$
            & $0.7771$
            & $0.7487$ \\
\end{tabular}
        
        \label{Table: Direct Multi Class}
\end{table*}

\end{document}